%% file: main.tex
\documentclass[pdflatex,sn-mathphys-num,iicol]{sn-jnl}

\usepackage{graphicx}%
\usepackage{anyfontsize}
\usepackage{multirow}%
\usepackage{amsmath,amssymb,amsfonts}%
\usepackage{amsthm}%
\usepackage{mathrsfs}%
\usepackage[title]{appendix}%
\usepackage{xcolor}%
\usepackage{textcomp}%
\usepackage{manyfoot}%
\usepackage{booktabs}%
\usepackage{algorithm}%
\usepackage{algorithmicx}%
\usepackage{algpseudocode}%
\usepackage{listings}%
\usepackage{cleveref}
\usepackage{array}
\usepackage{mathrsfs}
\usepackage{bookmark}
\usepackage{caption} 
\hypersetup{bookmarksdepth=3}
\algnewcommand\Input{\item[\textbf{Input:}]}
\algnewcommand\Output{\item[\textbf{Output:}]}

\definecolor{codegreen}{rgb}{0,0.6,0}
\definecolor{codegray}{rgb}{0.5,0.5,0.5}
\definecolor{codepurple}{rgb}{0.58,0,0.82}
\definecolor{backcolour}{rgb}{0.95,0.95,0.92}

\definecolor{brickred}{rgb}{0.8, 0.25, 0.33}

\input{Notations.tex}

\theoremstyle{thmstyleone}%
\theoremstyle{thmstyletwo}%
\newtheorem{remark}{Remark}%

\theoremstyle{thmstylethree}%

\newenvironment{sketchofproof}{\begin{proof}[Sketch of proof]}{\end{proof}}

\begin{document}
\title[Exact Evaluation of the Accuracy of Diffusion Models for Inverse Problems with Gaussian Data Distributions]{Exact Evaluation of the Accuracy of Diffusion Models for Inverse Problems with Gaussian Data Distributions}


\author[1]{\fnm{Emile} \sur{Pierret}}\email{pierret@math.cnrs.fr}

\author*[1,2]{\fnm{Bruno} \sur{Galerne}}\email{bruno.galerne@univ-orleans.fr}

\affil[1]{\orgdiv{Université d'Orléans, Université de Tours, CNRS, IDP, UMR 7013, Orléans, France}}

\affil[2]{\orgdiv{Institut universitaire de France (IUF)}}


\abstract{Used as priors for Bayesian inverse problems, diffusion models have recently attracted considerable attention in the literature. Their flexibility and high variance enable them to generate multiple solutions for a given task, such as inpainting, super-resolution, and deblurring. 
However, there is still a lack of understanding about how accurately these conditional diffusion algorithms perform conditional sampling.
In this article, we investigate the errors induced by these models when applied to a Gaussian data distribution for which the score function is exactly known.
Within this constrained context, we are able to precisely analyze the discrepancy between the theoretical resolution of inverse problems via conditional sampling and the practical distributions generated by conditional diffusion models.
This is done by characterizing all the involved iterative Gaussian processes and by computing the exact Wasserstein distance between the distributions of the diffusion model samplers and the ideal conditional distribution associated with the inverse problem.
Our findings allow for the comparison of two major algorithms from the literature, Deep Posterior Sampling (DPS) and Pseudo-inverse Guided Diffusion Models ($\Pi$GDM), and the introduction of the new paradigm Conditional Gaussian Diffusion Models (CGDM) that is shown to be more accurate for Gaussian data distributions. 
}




\maketitle

\section{Introduction}

Inverse problems are ubiquitous in scientific imaging, where the goal is to reconstruct a clean image from partial or degraded observations. Such problems arise in a wide range of applications, including microscopy, medical imaging, computational photography, and satellite observation. Common tasks such as deblurring, super-resolution, and inpainting are typical examples. These problems are inherently ill-posed: multiple solutions are consistent with the observed data, making a single reconstruction often unreliable or unrepresentative of the underlying ambiguity.

The Bayesian framework offers a principled approach for handling this uncertainty. In this setting, observations are modeled as degraded realizations from a prior distribution, and the objective becomes to characterize the posterior distribution of the clean image conditioned on these observations. This posterior encodes the full set of plausible solutions. The central challenge is thus to sample from this distribution in a faithful and efficient manner.

Generative models—particularly those trained on large datasets of natural images—have recently demonstrated remarkable capabilities in producing realistic samples. These include variational autoencoders (VAEs) \cite{Kingma_VAE}, generative adversarial networks (GANs) \cite{Goodfellow_GAN_2014_Neurips}, normalizing flows \cite{Kingma_Dhariwal_GLOW_NEURIPS2018}, and, more recently, diffusion models \cite{Ho_DDPM_2020_Neurips,Song_score-based_through_SDE_2023,Song_Ermon_generative_modeling_by_estimating_gradients_of_the_data_distribution_NEURIPS2019} and flow matching \cite{Albergo_etal_stochastic_interpolants_JMLR2025,Lipman_etal_flow_matching_for_generative_modeling_ICLR2023,Liu_Gong_Liu_Flow_Straight_and_Fast_ICLR2023}. 
Among these, diffusion models stand out for their training stability, their theoretically grounded formulation based on stochastic processes, and their ability to generate perceptually high-quality samples \cite{Dhariwal_diffusion_beats_gan_2021_neurips}.
Many algorithms  builds on the generative capabilities of generative models to solve inverse problem~\cite{Prost_HVAE,Karras_styleGan,Lugmayr_SRFlow_2020_ECCV,Daras_diffusion_survey_inverse_problems_2024,Chung_DPS_ICLR_2023,Chung_MCG_2022_neurips,Song_pseudo_inverse_guided_diffusion_2023_ICLR,Kawar_DDRM_2022_ICLR,FlowDPS_Kim_2025,
Zhang_2025_CVPR, 
Generation_phases_FM_Gagneux_2026,PnP_flow_martin_2025}.
Notably, diffusion models have been successfully employed to produce visually convincing reconstructions that capture the diversity of plausible solutions~\cite{Chung_DPS_ICLR_2023,Chung_MCG_2022_neurips,Song_pseudo_inverse_guided_diffusion_2023_ICLR,Kawar_DDRM_2022_ICLR, Daras_diffusion_survey_inverse_problems_2024, Zhang_2025_CVPR}, making these approaches the current state of the art. 

However, despite their empirical success, a crucial question often remains overlooked: \textbf{To what extent do the samples generated by these models faithfully reflect the true posterior distribution?} This issue, already studied in the literature \cite{moroy2025evaluatingposteriorsamplingability,Louppe_EM_diffusion,thong2024bayesianimagingmethodsreport}, is especially pressing in sensitive contexts, such as biomedical imaging or remote sensing, where biased or under-representative uncertainty estimates may have significant consequences. Common evaluation metrics, such as the Fréchet Inception Distance (FID) \cite{Heusel_etal_GANs_local_nash_equilibrium_2017_neurips}, are not suited for assessing statistical fidelity to the target posterior distribution since FID summarizes all image information to the low dimensional (2048) feature space of the inception network. In this work, to sidestep these limitations, we directly compare image distributions.

In prior work \cite{pierret_diffusion_models_gaussian_distributions_2024}, we studied diffusion models in their continuous formulation \cite{Song_score-based_through_SDE_2023}, focusing on Gaussian data distributions. While such a setting lacks direct practical relevance for real-world inverse problems, it provides a controlled and analytically tractable framework for evaluating the accuracy of diffusion-based posterior sampling. This Gaussian setting is also leveraged in recent theoretical studies to establish convergence and approximation guarantees for diffusion models \cite{Strasman_analysis_noise_schedule_SGM_2025_TMLR,hurault2025scorematchingdiffusionfinegrained}.

Building on these foundations, the present work focuses on the application of various diffusion-based algorithms from the literature to linear inverse problems involving images drawn from a Gaussian distribution. Under these assumptions, we are able to perform computations on low-dimensional toy examples and investigate the deblurring of Gaussian microtextures \cite{Galerne_Gousseau_Morel_random_phase_textures_2011_IEEE} at larger scales.
Rather than relying on perceptual or empirical metrics, we propose a more rigorous analysis based on exact computation of Wasserstein distances {directly between image distributions}. This approach enables an exact quantitative assessment of the discrepancy between the generated distribution and the ground-truth posterior in a Gaussian framework where both quantities are explicitly accessible.

The remainder of the paper is organized as follows. In \Cref{sec:reminder_diffusion_models_for_solving_inverse_problems}, we begin by reviewing the discrete DDPM model \cite{Ho_DDPM_2020_Neurips}, which serves as the basis for our analysis and then  we introduce, within a unified framework, two posterior sampling algorithms from the literature: DPS \cite{Chung_DPS_ICLR_2023} and $\Pi$GDM \cite{Song_pseudo_inverse_guided_diffusion_2023_ICLR}. 
Next, in \Cref{sec:study_under_Gaussian_assumption}, under the assumption of Gaussian data, we present the Conditional Gaussian Diffusion Model (CGDM), an algorithm inspired by closed-form expressions available in this regime and we describe an efficient procedure for comparing these algorithms using the 2-Wasserstein distance, which we apply to toy examples and several deblurring scenarios involving Gaussian microtextures in \Cref{sec:study_case_deblurring}. We conclude with a discussion on the challenges of extending this methodology to broader classes of inverse problems in \Cref{sec:discussion}.

\section{Preliminaries on diffusion models for solving inverse problems}

\label{sec:reminder_diffusion_models_for_solving_inverse_problems}

\subsection{Diffusion models for image generation}

The objective of generative modeling is to sample from a data distribution \(p_0\) defined over images.
In this work, we consider the Discrete Denoising Diffusion Probabilistic Model (DDPM) introduced in \cite{Ho_DDPM_2020_Neurips}. 
This framework consists of two successive phases. First, a discrete forward diffusion process gradually corrupts the data by adding noise. Second, a discrete reverse process aims at inverting the diffusion dynamics, thereby constructing trajectories from noise samples to data samples.

The forward process is defined as
 \begin{equation}
 \label{eq:forward_DDPM}
 \begin{aligned}
     \x_{t} & = \sqrt{1-\beta_t}\x_{t-1}+\sqrt{\beta_t}\z_t, \\
    & \quad 1 \leq t \leq T, \quad \z_t \sim \N, \quad  \x_0 \sim p_0,
 \end{aligned}
 \end{equation}
 where $\x_t$ lives in $\R^d$, $\N$ designates the standard normal distribution, $T = 1000$ is the number of steps and $\left(\beta_t\right)_{1 \leq t \leq T}$ is an increasing noise schedule. Ho \emph{et al.} \cite{Ho_DDPM_2020_Neurips} propose a linear schedule from $\beta_1 = 10^{-4}$ to $\beta_T = 0.02$.  All the transitions $p(\x_t\mid \x_{t-1})$ are Gaussian and by denoting $p_t$ the probability density of $\x_t$, $\alpha_t = 1-\beta_t$ and $\alphabar_t = \prod_{s=1}^t \alpha_s$, for $1 \leq t \leq T$,
  \begin{equation}
  \begin{aligned}
 \label{eq:forward_xt_x0}
 	\x_{t} & = \sqrt{\alphabar_t}\x_{0}+\sqrt{1-\alphabar_t}\bxi_t, \\
    & \quad 1 \leq t \leq T, \quad\bxi_t \sim \N, \quad  \x_0 \sim p_0.
     \end{aligned}
 \end{equation}
Consequently, by assuming that $p_0$ admits an expectation $\bmu$ and a covariance matrix $\bSigma$, it follows that
\begin{align}
	\E\left[\x_t\right] & = \sqrt{\alphabar_t}\bmu, \\
	\Cov(\x_t) & = \alphabar_t\bSigma + (1-\alphabar_t)\I.
\end{align}
Observe that the sequence \((\bar{\alpha}_t)_{t=0}^T\) is decreasing and satisfies \(\bar{\alpha}_T \approx 0\). Consequently, as $t$ grows, the marginal distribution \(p_t\) of \(\x_t\) is close to the standard Gaussian distribution \(\N\). 

To construct an approximate sampling procedure for the data distribution \(p_0\), one introduces the reverse diffusion process, referred to as the \emph{backward process}. Following \emph{Ho et al.}~\cite{Ho_DDPM_2020_Neurips}, this process is defined through the iterative scheme
  \begin{equation}
 \label{eq:backward_DDPM}
 \begin{aligned}
     \y_T & \sim \N \\
     \y_{t-1} & = \frac{1}{\sqrt{\alpha_t}}\left(\y_t + \beta_t \nabla \log p_t(\y _t)\right) + \sigma_t \z_t, \\
     &\z_t\sim \N,  \quad 1 \leq t \leq T,
 \end{aligned}
 \end{equation}
 where $\nabla \log p_t$ is called the score function. 
Note that we use the variable $\y$ rather than $\x$ to distinguish the backward process from the forward process. 
 { Diffusion models are particularly used in the literature because the score function can be well estimated by  a neural network (generally a U-Net model or a transformer) by score matching \cite{Ho_DDPM_2020_Neurips,sohl-dickstein_deep_learning_nonequilibrium_thermodynamics_2015_ICLR}.} 

 \begin{remark}[Backward variance schedule]
 	  The choice that $p(\y_{t-1}\mid \y_t)$ has a diagonal covariance $\sigma_t^2\I$ is optimal in sense of the KL divergence for $\sigma_t^2 = \tilde{\beta}_t = \frac{1-\alphabar_{t-1}}{1-\alphabar_{t}}\beta_t$ \cite{sohl-dickstein_deep_learning_nonequilibrium_thermodynamics_2015_ICLR}. However, in \cite{Ho_DDPM_2020_Neurips}, experimental results are similar with $\sigma^2_t = \beta_t$. Another approach is to learn the noise schedule $(\sigma_t)_{1 \leq t \leq T}$ in the form $\exp(v\log \beta_t + (1-v)\log \tilde{\beta}_t)$ \cite{Dhariwal_diffusion_beats_gan_2021_neurips}. In the following, for the sake of simplicity we assume that $\sigma_t = \beta_t$ in our experiments {but our results can easily be extended to other variance schedules.}
 \end{remark}

 \subsection{DDPM for solving inverse problems}
 \label{sec:conditional_DDPM}
 
Let us recall some key aspects of diffusion models in the context of image restoration. We focus on solving linear inverse problems
 \begin{equation}
 	\label{eq:inverse_problem}
    \begin{aligned}
 	\V & = \A\x_0 + \sigma \n \in \R^m, \\
    & \quad  \x_0 \sim p_0,\quad \sigma > 0,\quad \n \sim \N,
    \end{aligned}
 \end{equation}
that is, the data \(\x_0\) is degraded through a linear operator \(\A\in\R^{m\times d}\) and corrupted with additive measurement noise \(\sigma \n\).
The objective is therefore to sample from the posterior distribution \({p_0(\,\cdot \mid \V)}\).
Such a sampling procedure makes it possible to generate a collection of plausible reconstructions consistent with the corrupted observation \(\V\).
As discussed previously, DDPMs provide a natural framework for sampling from a prescribed probability distribution.
Accordingly, the DDPM considered in this work is designed to sample from
 \begin{equation}
 \label{eq:conditional_forward_DDPM}
 \begin{aligned}
 	\tx_{t} & = \sqrt{1-\beta_t}\tx_{t-1}+\sqrt{\beta_t}\tz_t, \\
    & \quad 1 \leq t \leq T, \quad \tz_t \sim \N, \quad  \tx_0 \sim p_0(\cdot \mid \V),
\end{aligned}
 \end{equation}
and the distribution of \(\tx_t\) is denoted by \(\tp_t\) for all \(0 \leq t \leq T\).
The corresponding reverse diffusion process is given by
 \begin{equation}
 \label{eq:conditional_backward_DDPM}
 \begin{aligned}
  \ty_T & \sim \N \\
     \ty_{t-1} & = \frac{1}{\sqrt{\alpha_t}}\left(\ty_t + \beta_t \nabla \log \tp_t(\y_t) \right) + \sqrt{\beta_t} \tz_t, \\
     & \quad \tz_t\sim \N,\quad  1 \leq t \leq T.
 \end{aligned}
 \end{equation}
A straightforward approach would consist in training a DDPM directly on pairs of clean and degraded data so as to approximate the conditional score \(\nabla \log \tp_t\). 
However, such a strategy is computationally expensive, since each inverse problem requires training a dedicated model.
To circumvent this limitation, a key observation is that Bayes' rule yields
 \begin{equation}
 \label{eq:Bayes_rule}
 	\nabla_{\x} \log \tp_t(\x_t) = \nabla_{\x} \log p_t(\x_t) + \nabla_{\x} \log p_t(\V \mid \x_t), 
 \end{equation}
where \(p_t\) denotes the marginal distribution associated with the unconditional forward process~\eqref{eq:forward_DDPM}.
In the following, the distribution \(p_t(\V \mid \x_t)\) is referred to as the \emph{noisy likelihood}, while \(p_t(\x_t \mid \V)\) is referred to as the \emph{noisy posterior}.

Assuming that the score function \(\nabla_{\x} \log p_t(\x)\) is available through a diffusion model trained for image generation, relation~\eqref{eq:Bayes_rule} provides an explicit decomposition of the conditional score \(\nabla_{\x} \log \tp_t(\x_t)\).
The remaining difficulty lies in estimating the term \(\nabla_{\x} \log p_t(\V \mid \x_t)\).
In practice, the exact noisy likelihood \(p_t(\V \mid \x_t)\) is generally intractable due to the complexity of the underlying prior distribution.
To overcome this issue, several works~\cite{Chung_DPS_ICLR_2023,Song_pseudo_inverse_guided_diffusion_2023_ICLR} approximate \(p_t(\V \mid \x_t)\) by a Gaussian distribution, which is therefore entirely determined by its mean and covariance matrix.

In particular, the conditional expectation associated with the noisy likelihood \(p_t(\V \mid \x_t)\) satisfies
\begin{equation}
\label{eq:conditional_mean_V_xt}
\E(\V \mid \x_t)
=
\E(\A \x_0 + \sigma \n \mid \x_t)
=
\A \E(\x_0 \mid \x_t),
\end{equation}
since the measurement noise is centered and independent of \(\x_t\).
Consequently, the characterization of \(\E(\V \mid \x_t)\) reduces to the estimation of the conditional expectation \(\E(\x_0 \mid \x_t)\), which corresponds to the minimum mean square error (MMSE) estimator of \(\x_0\) given \(\x_t\).
Tweedie's formula establishes that this estimator can be expressed directly in terms of the score function associated with the unconditional diffusion process, providing
\begin{equation} 
\begin{aligned} 
\widehat{\x}_0(\x_t) & 
:= \E\left[\x_0 \mid \x_t\right] \\ 
& = \frac{1}{\sqrt{\alphabar_t}}\left(\x_t + (1-\alphabar_t) \nabla_{\x}\log p_t(\x_t)\right) .
\end{aligned} 
\label{eq:Tweedie_formula} 
\end{equation}
Substituting~\eqref{eq:Tweedie_formula} into~\eqref{eq:conditional_mean_V_xt} yields
\begin{equation}
\begin{aligned}
\E\left[\V \mid \x_t\right]
&=
\A \widehat{\x}_0(\x_t) \\
&=
\frac{1}{\sqrt{\alphabar_t}}
\A
\left(
\x_t
+
(1-\alphabar_t)\nabla_{\x}\log p_t(\x_t)
\right).
\end{aligned}
\end{equation}
It remains to specify a covariance matrix \(\bC_{\V \mid t}\) in order to approximate the conditional covariance \(\Cov(\V \mid \x_t)\).
Under the Gaussian approximation of the noisy likelihood, this leads to the expression
\begin{equation}
\nabla_{\x} \log p_t(\V \mid \x_t)
=
-\frac{1}{2}
\nabla_{\x}
\left\|
\V - \A \widehat{\x}_0(\x_t)
\right\|_{\bC_{\V \mid t}^{-1}}^2,
\label{eq:grad_noisy_likelihood_approx_covariance_generic}
\end{equation}
where, for any symmetric positive definite matrix \(\A\), the notation
$
\|\x\|_{\A}^2 := \x^\top \A \x
$
is employed.
Combining this approximation with the reverse diffusion dynamics yields the following conditional DDPM iteration
\begin{equation}
\begin{aligned}
\ty_{t-1}
=
&\frac{1}{\sqrt{\alpha_t}}
\Bigg(
\ty_t
+
\beta_t \nabla_{\ty} \log p_t(\ty_t)
\\
&
-
\frac{\beta_t}{2}
\nabla_{\ty_t}
\left\|
\V-\A\widehat{\tx}_0(\ty_t)
\right\|_{\bC_{\V\mid t}^{-1}}^2
\Bigg)
\\
&
+ \sigma_t \z_t,
\quad
\z_t \sim \N,
\quad
1 \leq t \leq T.
\end{aligned}
\end{equation}
This scheme corresponds to the conditional reverse diffusion process described in \Cref{algo:backward_conditional_DDPM}.
 
    \begin{algorithm*}
      \caption{Generic algorithm to DPS, $\Pi$GDM, CGDM}\label{algo:backward_conditional_DDPM} 
      \begin{algorithmic}[1]
      \Input The noisy observation $\V$, the approximate likelihood covariances $(\bC^{\mathrm{algo}}_{\V \mid t})_{0 \leq t \leq T}$ (see~\Cref{tab:comparison_expression_grad_cond}).
\State $\y_T \sim \N$
        \For{$t=T$ \textbf{to} 1}
        \State $\widehat{\y}_0(\y_t) \leftarrow \frac{1}{\sqrt{\alphabar_t}}\left(\y_t + (1-\alphabar_t) \nabla \log p_t(y_t)\right)$
    \State $\y_{t-1} \leftarrow\frac{1}{\sqrt{\alpha_t}}\left[\y_t + \beta_t \left(\nabla\log p_t(\y_t) -\frac{1}{2}\nabla_{\y} \left\|\V-\A\widehat{\x}_0(\y_t) \right\|_{\left(\bC^{\mathrm{algo}}_{\V\mid t}\right)^{-1}}^2\right)\right] + \sqrt{\beta_t} \z_t$, $\quad \z_t\sim \N$
    \EndFor   
    \Output The reconstructed image $\y_0$.   
      \end{algorithmic}
    \end{algorithm*}

\renewcommand{\arraystretch}{2} 

\begin{table*}[t]
\centering
\begin{tabular}{| >{\raggedright\arraybackslash}m{2.5cm}| >{\raggedright\arraybackslash}m{6.5cm} | >{\raggedright\arraybackslash}m{5cm} |}
\hline
Method 
&
Approximation of $p(\x_0\mid \x_t)$ 
&
Approximate likelihood covariance $\bC^{\mathrm{algo}}_{\V\mid t}$  \\
\hline
DPS \cite{Chung_DPS_ICLR_2023} 
& $p(\x_0\mid \x_t) \approx \delta_{\widehat{\x_0}(\x_t)}$
& $\frac{\sigma^2}{\alphaDPS} \I$ \\
\hline
$\Pi$GDM \cite{Song_pseudo_inverse_guided_diffusion_2023_ICLR}$\;$ 
& $p(\x_0\mid \x_t) \approx \mathcal{N}(\widehat{\x}_0(\x_t) , r_t^2\I)$
&  $(1-\alphabar_t)\A  \A^T+\sigma^2\I$ \\
\hline
CGDM 
& $p(\x_0 \mid \x_t)  \approx \mathcal{N}
		\left(
		\widehat{\x}_0(\x_t), (1-\alphabar_t)\bSigma \bSigma_t^{-1}
		\right)$, $\bSigma_t = \alphabar_t \bSigma + (1-\alphabar_t)\I$
		$\quad$ $\quad$ $\quad$ $\quad$ $\quad$ $\quad${\footnotesize  (exact if $p_0 = \mathcal{N}(\bmu,\bSigma)$)}
& $(1-\alphabar_t)\A \bSigma \bSigma_t^{-1} \A^T+\sigma^2\I$
$\quad$ $\quad$ $\quad${\footnotesize (exact if $p_0 = \mathcal{N}(\bmu,\bSigma)$)}
 \\
\hline
\end{tabular}
\caption[
Comparison of the covariance matrices associated with the noisy likelihood model \(p(\V \mid \x_t)\) for DPS, \(\Pi\)GDM, and CGDM.
]{
\label{tab:comparison_expression_grad_cond}
\textbf{Comparison of the covariance matrices used to model the noisy likelihood distribution \(p(\V \mid \x_t)\) in CGDM, DPS, and \(\Pi\)GDM.}
The matrix \(\bC_{\V \mid t}\) defines the approximate noisy likelihood $\nabla_{\x_t}\log p(\V \mid \x_t)$ (see Equation~\eqref{eq:grad_noisy_likelihood_approx_covariance_generic}).
The CGDM expression corresponds to the exact covariance obtained under the Gaussian assumption, whereas DPS and \(\Pi\)GDM rely on approximate parameterizations.
}
\end{table*}
\renewcommand{\arraystretch}{1}

\noindent 
In the remainder of this work, attention is restricted to two approaches proposed in the literature, namely DPS~\cite{Chung_DPS_ICLR_2023} and \(\Pi\)GDM~\cite{Song_pseudo_inverse_guided_diffusion_2023_ICLR}.
Both methods can be interpreted as distinct parameterizations of the covariance matrix \(\bC_{\V \mid t}\) used to approximate the noisy likelihood.
Their respective formulations are presented below and summarized in Table~\ref{tab:comparison_expression_grad_cond}.
A broader overview of diffusion-based methods for inverse problems can be found in the survey of Daras et al. \cite{Daras_diffusion_survey_inverse_problems_2024}.

\paragraph{Diffusion Posterior Sampling (DPS)} 
DPS is designed to address a broad class of inverse problems, including linear tasks such as inpainting, deblurring, and super-resolution, as well as nonlinear problems such as phase retrieval and non-uniform deblurring  \cite{Chung_DPS_ICLR_2023}.
The central approximation of DPS consists in replacing the noisy likelihood by
\begin{equation}
\log p_t(\V \mid \x_t)
\approx
\log p(\V \mid \x_0 = \widehat{\x}_0(\x_t)).
\end{equation}
Recalling from Equation~\eqref{eq:inverse_problem} that
\begin{equation}
p(\V \mid \x_0)
=
\mathcal{N}\left(\A\x_0,\sigma^2\I\right),
\end{equation}
the previous approximation amounts to modeling the noisy likelihood with covariance matrix
\begin{equation}
\bC^{\tiny \mathrm{DPS}}_{\V \mid t}
=
\sigma^2 \I.
\end{equation}
The corresponding likelihood gradient is therefore given by
\begin{equation} 
\begin{aligned} 
\nabla_{\x_t}\log p(\V\mid & \x_0 = \widehat{\x}_0(\x_t)) 
\\ 
& = -\frac{1}{2\sigma^2}\nabla_{\x_t}\left\|\V-\A\widehat{\x}_0(\x_t)\right\|^2. 
\end{aligned} 
\end{equation}
In practice, this approximation often leads to numerical instabilities.
Indeed, it implicitly assumes that
\begin{equation}
p(\x_0 \mid \x_t)
\approx
\delta_{\widehat{\x}_0(\x_t)},
\end{equation}
where \(\delta_{\widehat{\x}_0(\x_t)}\) denotes the Dirac distribution concentrated at \(\widehat{\x}_0(\x_t)\).
Hence, the conditional variance of \(\x_0\) given \(\x_t\) is neglected, which may result in an underestimation of the covariance matrix \(\bC_{\V \mid t}\).
As a consequence, the associated inverse covariance weighting can become excessively large and destabilize the reverse diffusion iterations.
To alleviate this issue, Chung et al. introduce a parameter \(\alpha_{\mathrm{DPS}} > 0\), leading to the modified gradient
\begin{equation} 
\begin{aligned} 
\nabla_{\x_t}\log p(\V\mid & \x_0 = \widehat{\x}_0(\x_t)) \\ 
& = -\frac{\alphaDPS}{2\sigma^2}\nabla_{\x_t}\left\|\V-\A\widehat{\x}_0(\x_t)\right\|^2. 
\end{aligned} 
\end{equation}
The parameter \(\alpha_{\mathrm{DPS}}\) depends both on the considered dataset and on the inverse problem under study (see \cite[Appendix~D.1]{Chung_DPS_ICLR_2023}).
Accordingly, the effective covariance matrix associated with DPS is taken as
\begin{equation} 
\bC^{\tiny \mathrm{DPS}}_{\V \mid t}
=
\frac{\sigma^2}{\alpha_{\mathrm{DPS}}}\I.
\end{equation}
 
 \begin{remark}[Gap between theory and practical implementation of the DPS algorithm]
 \label{rem:DPS_implementation}
 	In practice, a second approximation is made \cite{Chung_DPS_ICLR_2023}, introducing the adaptative reweighting
 \begin{equation}
 \begin{aligned}
 	 	\frac{\beta_t}{2\sqrt{\alpha_t}} &\nabla_{\x_t}\log p(\V\mid  \x_0 = \widehat{\x}_0(\x_t))
 	 	\\
        & = -\frac{\alpha_{\text{\tiny DPS}}}{\|\A\widehat{\x}_0(\x_t)-\V\|} \nabla_{\x_t}\left\|\V-\A\widehat{\x}_0(\x_t)\right\|^2 \\
 	 	& = -\alpha_{\text{\tiny DPS}} \nabla_{\x_t}\left\|\V-\A\widehat{\x}_0(\x_t)\right\|.
  \end{aligned}
 \end{equation}
 This new formulation changes considerably the initial model and it amounts to setting
 \begin{equation}
 \begin{aligned}
 	\nabla_{\x_t} \log p(\V\mid & \x_0 = \widehat{\x}_0(\x_t)) \\
 	& \propto -\frac{2\sqrt{\alpha_t} \alpha_{\text{\tiny DPS}}}{\beta_t}\nabla_{\x_t}\left\|\V-\A\widehat{\x}_0(\x_t)\right\|.  
\end{aligned}
 \end{equation}
 It can be interpreted as modeling the distribution $p(\V\mid \x_0 = \widehat{\x}_0(\x_t))$ not as a Gaussian distribution but a modified Multivariate Generalized Gaussian Distribution (MGGD) \cite{Manzano_matrix_variate_generalization_power_exponential_family_2002,Pascal_Parameter_estimation_generalized_Gaussian_IEEE_2013}.
 Another heuristic setting which is used in the official implementation of this method\footnote{\url{https://github.com/DPS2022/diffusion-posterior-sampling}} and that guarantees its stability is the clamping of the estimated denoised image $\widehat{\x}_0(\x_t)$ between $-1$ and $1$. To stay in the Gaussian realm, we do not consider these heuristic corrections in what follows.
 \end{remark}
 
\paragraph{Pseudo-inverse Guided Diffusion Models ($\Pi$GDM)}

The \(\Pi\)GDM algorithm \cite{Song_pseudo_inverse_guided_diffusion_2023_ICLR} is designed for inverse problems such as inpainting, JPEG artifact removal, and deblurring.
The key approximation proposed by Song \emph{et al.} consists in modeling the conditional distribution \(p(\x_0 \mid \x_t)\) as
\begin{equation}
\label{eq:first_approx_Song}
p(\x_0 \mid \x_t)
\approx
\mathcal{N}
\left(
\widehat{\x}_0(\x_t),
r_t^2 \I
\right)
\end{equation}
for some choice of $r_t$ discussed below. Combining this approximation with the observation model~\eqref{eq:inverse_problem} yields
\begin{equation}
p(\V \mid \x_t)
\approx
\mathcal{N}
\left(
\A\widehat{\x}_0(\x_t),
r_t^2 \A\A^\top + \sigma^2 \I
\right).
\end{equation}
Accordingly, the covariance matrix associated with the noisy likelihood is chosen as
\begin{equation}
\bC^{\tiny \Pi\mathrm{GDM}}_{\V \mid t}
=
r_t^2 \A\A^\top + \sigma^2 \I,
\end{equation}
which explicitly depends on both the degradation operator \(\A\) and the diffusion time \(t\).
Such a dependence is natural, since the uncertainty associated with the reconstruction should reflect both the forward degradation mechanism and the noise level induced by the diffusion process.
The parameter \(r_t\) is estimated by considering the particular case where the data distribution \(p_0\) is a standard Gaussian distribution.
For DDPMs, this heuristic leads to the choice
$r_t^2 = 1 - \alphabar_t.$

 \begin{remark}[$\Pi$GDM algorithm for DDPM]
 \label{rem:PiGDM_implementation}
 	$\Pi$GDM \cite{Song_pseudo_inverse_guided_diffusion_2023_ICLR} was first described for the DDIM algorithm \cite{song_DDIM}. However, the approximation of $p_t(\V\mid \x_t)$ can be extended to the DDPM one. 
 Note that $r_t^2\A\A^T + \sigma^2\I$ is invertible since $\A\A^T$ is positive semi-definite. In the noiseless setting $\sigma = 0$ (not considered here), a pseudo-inverse of $\A$ is applied, which is the reason this method is referred to as \emph{Pseudoinverse}-Guided Diffusion Models.
  \end{remark}

 \section{Proposed approach: Study under Gaussian assumption}
\label{sec:study_under_Gaussian_assumption}

Diffusion-based algorithms for inverse problems are generally assessed through empirical performance metrics computed on large-scale datasets.
However, the intractability of the score function and of its conditional counterparts constitutes a major obstacle to the development of a rigorous theoretical analysis of these methods.
Furthermore, as discussed previously, the diffusion framework associated with linear inverse problems naturally involves Gaussian structures.
In order to enable a tractable theoretical comparison of the different algorithms, we propose an analysis restricted to the setting where the prior distribution \(p_0\) is Gaussian.
More precisely, the following assumption is adopted throughout this work.
\begin{assumption}[Gaussian assumption]
\label{assumption:Gaussian}
The prior distribution \(p_0\) is assumed to be Gaussian:
\[
p_0 = \mathcal{N}(\bmu,\bSigma)
\qquad \text{in } \mathbb{R}^d.
\]
\end{assumption}
In this case, as developed below, we can derive all the closed-forms expressions of the distributions and precisely compare the different algorithms. 
\begin{remark}
The covariance matrix \(\bSigma\) is not assumed to be full rank.
Consequently, the Gaussian distribution \(p_0 = \mathcal{N}(\bmu,\bSigma)\) may be degenerate and supported on a strict affine subspace of \(\mathbb{R}^d\).
This setting notably encompasses distributions concentrated on low-dimensional manifolds embedded in the ambient space.
Such a situation is particularly relevant in high-dimensional imaging problems, where data are often assumed to lie close to a lower-dimensional structure.
Allowing \(\bSigma\) to be singular therefore makes it possible to capture intrinsic low-dimensional geometries while retaining a Gaussian framework amenable to theoretical analysis.
\end{remark}

\subsection{Exact conditional Gaussian distribution}
\label{subsec:exact_Gaussian_formulas}
First, in the Gaussian setting, the use of diffusion models for solving the inverse problem is, in principle, unnecessary.
Indeed, the posterior distribution associated with the observation model can be derived explicitly.
More precisely, the conditional distribution of \(\x_0\) given the observation \(\V\) is Gaussian and is characterized by
\begin{equation}
\begin{aligned}
\label{eq:conditional_Gaussian_init}
p(\x_0 \mid \V)
     & = \mathcal{N}\left({\bmu}_{0\mid \V}, {\bC}_{0\mid \V}\right) \\
     \text{with } &{\bmu}_{0\mid \V} = \bmu+\bSigma \A^T \M^{-1}\left(\V-\A\bmu\right), \\
     &{\bC}_{0\mid \V} = \bSigma - \bSigma \A^T \M^{-1} \A\bSigma, \\
       & \M = \A\bSigma\A^T + \sigma^2 \I.
\end{aligned}
\end{equation}
For the sake of completness, a detailed derivation of this expression is provided in Appendix~\ref{appendix:x_t_mid_v}.
Consequently, exact posterior sampling can be performed directly, without resorting to an approximate diffusion-based procedure.

\subsection{Conditional Gaussian Diffusion Models}

In this context, the unconditional score $\nabla_{\x} \log p_t(\x_t)$ is explicit and is given by
\begin{equation}
\label{eq:Gaussian_score_DDPM}
	\nabla_{\x} \log p_t(\x)  = -\bSigma_t^{-1}\left(\x-\sqrt{\alphabar_t}\bmu\right),
\end{equation}
with
\begin{equation}
\bSigma_t = \alphabar_t \bSigma + (1-\alphabar_t)\I
\end{equation}
and where $\bSigma_t$ is invertible for $t> 0$, as described for the continuous case in \cite{pierret_diffusion_models_gaussian_distributions_2024}.
Using these explicit expressions, the conditional forward DDPM~\eqref{eq:conditional_forward_DDPM} associated with the initialization \(\tx_0 \sim p(\,\cdot \mid \V)\) can be characterized exactly.
More precisely, for every \(t \in \{0,\dots,T\}\),
\begin{equation}
\begin{aligned}
\label{eq:exact_cond_back_density}
& p_t(\x_t \mid \V)
     = \mathcal{N}\left({\bmu}_{t\mid \V}, {\bC}_{t\mid \V}\right) \\
    & \text{ with } {\bmu}_{t\mid \V} = \sqrt{\alphabar_t}\bmu+\sqrt{\alphabar_t}\bSigma \A^T \M^{-1}\left(\V-\A\bmu\right), \\
    & \text{ and } {\bC}_{t\mid \V} = \bSigma_t - \alphabar_t\bSigma \A^T \M^{-1} \A\bSigma.
\end{aligned}
\end{equation}
Within the Gaussian framework, the diffusion-based algorithms under consideration can therefore be compared to this exact conditional forward process, which they are intended to approximate through their reverse dynamics.

Another fundamental quantity is the noisy likelihood \(p_t(\V \mid \x_t)\), which is, as discussed above, approximated by a Gaussian distribution in both DPS and \(\Pi\)GDM.
In the present Gaussian setting, however, no additional approximation is required: the distribution \(p_t(\V \mid \x_t)\) is itself Gaussian and admits the explicit expression
\begin{equation}
\scriptsize
p_t(\V \mid \x_t)
		= \mathcal{N}
		\left(
		\A \widehat{\x}_0(\x_t), (1-\alphabar_t)\A \bSigma \bSigma_t^{-1} \A^T
		+\sigma^2\I\right),  
\label{eq:exact_cov}
\end{equation}
where $\widehat{\x}_0(\x_t)$ follows the Tweedie's formula~\eqref{eq:Tweedie_formula} that thanks to Equation~\eqref{eq:Gaussian_score_DDPM}
simplifies to
\begin{equation}
\widehat{\x}_0(\x_t)
	 = \bmu + \sqrt{\alphabar_t}\bSigma \bSigma_t^{-1}(\x_t - \sqrt{\alphabar}_t \bmu).
	\label{eq:Gaussian_x_0_chap}
\end{equation}
The proofs are provided in Appendix~\ref{appendix:x_zero_chap} and \ref{appendix:v_mid_x_t}. 

Thanks to this exact expression of the noisy likelihood, we are now able to introduce a new algorithm, referred to as the \emph{Conditional Gaussian Diffusion Model} (CGDM), for which the covariance matrix of the noisy likelihood is chosen to be the exact covariance obtained under the Gaussian assumption:
\begin{equation}
\bC_{ \V \mid t}^{\mathrm{CGDM}}
=
(1-\alphabar_t)\A \bSigma \bSigma_t^{-1} \A^\top
+
\sigma^2\I.
\end{equation}
To summarize, diffusion models-based inverse problems solvers DPS and $\Pi$GDM introduce an approximate likelihood covariance $\bC_{\V\mid t}$. Under the Gaussian assumption, there is no need for such an approximation, the exact covariance is given by \eqref{eq:exact_cov} and is directly used by our proposed CGDM algorithm (see also last line of \Cref{tab:comparison_expression_grad_cond}).

Now that the CGDM algorithm has been introduced, we can turn to a theoretical and numerical comparison of the three algorithms DPS, $\Pi$GDM and CGDM under \Cref{assumption:Gaussian}.

\subsection{Comparison of the algorithms under Gaussian assumption}
\label{sec:comp_algo_under_Gaussian_assumption}

In the construction of the three previous algorithms, the noisy likelihood \(p(\V \mid \x_t)\) is approximated by a Gaussian distribution with covariance matrix \(\bC_{\V \mid t}\).
The objective of this section is to investigate the implications of this modeling choice on the corresponding posterior distribution.

\paragraph{Comparison of the covariance matrices associated with the noisy likelihood model \(p(\V \mid \x_t)\) for DPS, \(\Pi\)GDM, and CGDM.}

It is instructive to compare the behavior of the covariance matrices \(\bC^{\mathrm{algo}}_{\V \mid t}\) associated with the different algorithms.
For simplicity, let us consider the case \(\alpha_{\mathrm{DPS}} = 1\).
When \(t\) is close to \(T\), the covariance matrix \(\bSigma_t\) satisfies \(\bSigma_t \approx \I\), which yields the approximations
\begin{equation}
\begin{aligned}
\bC^{\tiny \mathrm{DPS}}_{\V \mid t}
&=
\sigma^2 \I,
\\
\bC^{\tiny \Pi\mathrm{GDM}}_{\V \mid t}
&\approx
(1-\alphabar_t)\A\A^\top + \sigma^2 \I,
\\
\bC^{\tiny \mathrm{CGDM}}_{\V \mid t}
&\approx
(1-\alphabar_t)\A\bSigma\A^\top + \sigma^2 \I.
\end{aligned}
\end{equation}
Consequently, the covariance matrix used in \(\Pi\)GDM provides a closer approximation to the exact Gaussian expression \(\bC^{\tiny \mathrm{CGDM}}_{\V \mid t}\), although it does not incorporate the covariance structure of the prior distribution through \(\bSigma\).
By contrast, the DPS covariance matrix substantially underestimates the uncertainty associated with the noisy likelihood, especially for diffusion times \(t\) close to \(T\), since \(1-\alphabar_t \approx 1\) in this regime.
This observation provides a theoretical explanation for the numerical instabilities discussed in \Cref{sec:conditional_DDPM}.

When \(t\) is close to \(0\), 
\(\bSigma_t \approx \bSigma\), leading to
\begin{equation}
\begin{aligned}
\bC^{\tiny \mathrm{DPS}}_{\V \mid t}
&=
\sigma^2 \I,
\\
\bC^{\tiny \Pi\mathrm{GDM}}_{\V \mid t}
&\approx
(1-\alphabar_t)\A\A^\top + \sigma^2 \I,
\\
\bC^{\tiny \mathrm{CGDM}}_{\V \mid t}
&\approx
(1-\alphabar_t)\A\A^\top + \sigma^2 \I.
\end{aligned}
\end{equation}
Hence, for small diffusion times, the covariance matrix used in \(\Pi\)GDM becomes asymptotically equivalent to the exact Gaussian covariance \(\bC^{\tiny \mathrm{CGDM}}_{\V \mid t}\).
In particular, both DPS and \(\Pi\)GDM recover the exact covariance at time \(t=0\), since \(\alphabar_0 = 1\), yielding
\begin{equation}
\bC_{\V \mid 0}
=
\sigma^2 \I.
\end{equation}
This observation is of particular importance in practice, as it shows that the different algorithms become exact at the final stages of the reverse diffusion process, where the generated samples are expected to recover the fine details of the target distribution.

\paragraph{Emulate the derivation of the noisy posterior \(p_t(\x_t \mid \V)\) for each algorithm.}
For each algorithm, we substitute the closed-form expression of \(\nabla_{\x} \log p_t(\x)\), which is exactly known in the Gaussian setting, together with the corresponding model adopted by the algorithm for \(\log p_t^\mathrm{algo}(\V \mid \x_t)\) (see Table~\ref{tab:comparison_expression_grad_cond}).
Using Bayes' rule (\Cref{eq:Bayes_rule}) in reverse, we define a distribution \(\tilde{p}_t^{\mathrm{algo}}(\x_t \mid \V)\) such that
$
\log \tilde{p}_t^{\mathrm{algo}}(\x_t \mid \V)
=
\log p_t(\x_t)
+
\log p_t^{\mathrm{algo}}(\V \mid \x_t),
$
up to an additive normalization constant independent of \(\x_t\).
\textbf{Note that this noisy posterior generally does not coincide with the distribution induced by the backward dynamics of the corresponding algorithm, but it still provides a useful interpretation of the underlying modeling assumptions.}
Denoting
$
\tilde{p}_t^{\mathrm{algo}}(\x_t \mid \V)
=
\mathcal{N}\!\left(
\tilde{\bmu}^{\mathrm{algo}}_{t \mid \V},
\tilde{\bC}^{\mathrm{algo}}_{t \mid \V}
\right),
$
the resulting computations yield
{\small
\begin{align}
	 & \tilde{\bC}^{\mathrm{DPS}}_{t \mid \V} \notag
      = \bSigma_t \notag\\
     & -\alphabar_t\bSigma\A^T\left(\sigma^2\I+\alphabar_t \A\bSigma^2\bSigma_t^{-1}\A^T\right)^{-1}\A\bSigma \\
	 & \tilde{\bC}^{{\Pi\mathrm{GDM}}}_{t \mid \V} \notag 
	 = \bSigma_t \notag\\
	&-\alphabar_t\bSigma\A^T\left(\sigma^2\I+\A\A^T+\alphabar_t \A(\bSigma^2\bSigma_t^{-1}\A^T-\I)\right)^{-1}\A\bSigma \\
	 &\tilde{\bC}^{\mathrm{CGDM}}_{t \mid \V}\notag 
 = \bSigma_t\notag\\
	 & -\alphabar_t\bSigma\A^T\left(\sigma^2\I+\A\bSigma\A^T\right)^{-1}\A\bSigma.
\end{align}%
}%
All derivations, including the explicit expressions of \(\tilde{\bmu}^{\mathrm{algo}}_{t \mid \V}\), are provided in Appendix~\ref{appendix:proof_expression_C_v_t}. 
Let us focus our discussion on the covariance matrices \(\tilde{\bC}^{\mathrm{algo}}_{t \mid \V}\), even though analogous conclusions can also be drawn for the posterior means \(\tilde{\bmu}^{\mathrm{algo}}_{t \mid \V}\).
First, observe that CGDM exactly recovers the forward conditional distributions (see \Cref{eq:exact_cond_back_density}). Moreover, for \(t = 0\), assuming that \(\bSigma\) is invertible and that \(\alphabar_0 = 1\), we obtain
 \begin{align}
	\tilde{\bC}^{\mathrm{DPS}}_{0 \mid \V}
	& = \bSigma-\bSigma\A^T\left(\sigma^2\I+\A\bSigma\A^T\right)^{-1}\A\bSigma \\
	\tilde{\bC}^{{\Pi\mathrm{GDM}}}_{0 \mid \V}
	& = \bSigma-\bSigma\A^T\left(\sigma^2\I+\A\bSigma\A^T\right)^{-1}\A\bSigma.
\end{align}
These expressions coincide with the exact covariance matrix of the posterior distribution \(p(\x_0 \mid \V)\).

\paragraph{Algorithms studied in forward time evolution}
A natural question is whether the distributions \(p_t^{\mathrm{algo}}(\x_t \mid \V)\) can be interpreted as the marginals of a forward DDPM process. Establishing such a correspondence would provide further insight into the effective target distributions implicitly modeled by these algorithms.
For a family of Gaussian distributions to correspond to the marginals of a forward diffusion process, the covariance matrices \(\tilde{\bC}_{t \mid \V}\) are expected to satisfy the structural form
\begin{equation}
	\tilde{\bC}_{t\mid \V} = \alphabar_t \tilde{\bC}_{0\mid \V} + (1-\alphabar_t)\I.
\end{equation}
The covariance structure associated with \(\tilde{p}_t^{\mathrm{DPS}}(\x_t \mid \V)\) does not, in general, correspond to that of a forward DDPM process. Indeed, the effective covariance term involved in the posterior expression depends explicitly on \(t\). The only situation in which this dependence disappears is the degenerate case \(\A = \zero\).
A similar observation holds for the \(\Pi\)GDM algorithm. In general, the covariance matrix associated with \(\tilde{p}_t^{\Pi\mathrm{GDM}}(\x_t \mid \V)\) also involves terms that vary with \(t\), preventing an interpretation in terms of a standard forward DDPM process. An important exception arises in the particular case \(\bSigma = \I\), for which the covariance matrices of \(\Pi\)GDM and CGDM coincide. This is consistent with the construction of \(r_t^2\) in \Cref{sec:conditional_DDPM}, which was designed to be exact when \(p_0\) is Gaussian with identity covariance.
By contrast, the covariance matrix associated with CGDM exactly matches the covariance structure obtained by applying the forward DDPM process (\Cref{eq:forward_DDPM}) to the conditional distribution \(p_{0 \mid \V}\) defined in \Cref{eq:conditional_Gaussian_init}. Consequently, the family of distributions \(\bigl(\tilde{p}_t^{\mathrm{CGDM}}(\x_t \mid \V)\bigr)_{0 \leq t \leq T}\) can be interpreted precisely as the marginals of the corresponding forward diffusion process.

\subsection{Recursive computation of the backward distributions}

\label{sec:recursive_computation_backward_distributions}

We provide here the building blocks required to efficiently compute the distributions generated by the different algorithms, enabling comparison with the target distribution.

\begin{prop}
\label{prop:compute_recursively_cov_mus_algos}
	Under the Gaussian assumption, for each $\algo \in \{\mathrm{DPS},\Pi\mathrm{GDM},\mathrm{CGDM}\}$, there exists $\A_t^\algo \in \R^{d\times d}$ and $\bb_t^\algo \in \R^{d}$ at each iteration $t=1,\ldots,1000$ such that
	\begin{equation}
\label{eq:algo_A_t_b_t}
	\begin{aligned}
		\y_T & \sim \N, \\
		\y_{t-1} & = \A_t^{\mathrm{algo}}\y_{t} + \bb_t^{\mathrm{algo}}+\sqrt{\beta_t} \z_t,\\
		&\quad 1 \leq t \leq T, \quad \z_t \sim \N.
	\end{aligned}
\end{equation}
Consequently, the distribution of the $t$-th iteration (generated in \textbf{backward time}) of each algorithm is 
Gaussian and by denoting them $p_t^{\mathrm{algo}}(\y_t\mid \V) = \mathcal{N}\left(\bmu_t^{\mathrm{algo}},\bSigma_t^{\mathrm{algo}}\right)$, they satisfy the recursive relations
\begin{equation}
\label{eq:Sigma_t_A_t}
\begin{aligned}
\bmu_T^\mathrm{algo} &= 0, \\
\bmu_{t-1}^\mathrm{algo}
&=
\A_t^{\mathrm{algo}}
\bmu_t^\mathrm{algo}
+ \bb_t^{\mathrm{algo}}
\\
\bSigma_T^\mathrm{algo} &= \I, \\
\bSigma_{t-1}^\mathrm{algo}
&=
\A_t^{\mathrm{algo}}
\bSigma_t^\mathrm{algo}
(\A_t^{\mathrm{algo}})^T
+\beta_t \I.
\end{aligned}
\end{equation}
\end{prop}

\begin{sketchofproof}
Considering the equivalent formulation of the different algorithms in Algorithm~\ref{algo:backward_conditional_DDPM}, the score term is linear in this setting. It therefore remains to show that the gradient term is also linear. Using the expression of $\widehat{\x}_0$ given in \eqref{eq:Gaussian_x_0_chap}, we obtain
\begin{equation}
\begin{aligned}
\frac{1}{2}&\nabla_{\y_t} \|\A \widehat{\x}_0(\y_t)-\V\|^2_{\bC_{\V \mid t}^{-1}} \\
&= \sqrt{\alphabar_t}\,\bSigma \bSigma_t^{-1}\A^T\bC_{\V \mid t}^{-1}
\bigl(\A\widehat{\x}_0(\y_t)-\V\bigr).
\end{aligned}
\end{equation}
The detailed derivation is provided in Appendix~\ref{appendix:proof:prop:compute_recursively_cov_mus_algos}.
\end{sketchofproof}

\begin{remark}
	Observing equations of \Cref{prop:compute_recursively_cov_mus_algos}, we see that the iterative means $\left(\bmu_t^{\mathrm{algo}}\right)_{0 \leq t \leq T}$ of the different algorithms can be obtained by applying the corresponding algorithm to the zero vector with no adding noise (i.e. setting $\z_t = 0$ in Equation~\eqref{eq:Sigma_t_A_t}).
\end{remark}

\subsection{Comparison of the algorithms on toy models}

\paragraph{Comparison using the 2-Wasserstein distance}

It was established that, under the Gaussian assumption, the processes generated by DPS, $\Pi$GDM, and CGDM are Gaussian, with explicitly computable means and covariance matrices.
As a consequence, these algorithms can be compared using the 2-Wasserstein distance, which admits a closed-form expression in the Gaussian setting \cite{Peyre_OT}. For two Gaussian distributions $\mathcal{N}(\bmu_1,\bSigma_1)$ and $\mathcal{N}(\bmu_2,\bSigma_2)$,
\begin{equation}
\label{eq:W2_Gaussian}
\begin{aligned}
	& \Wass^2\left(\mathcal{N}(\bmu_1,\bSigma_1),\mathcal{N}(\bmu_2,\bSigma_2)\right) \\
    & = \|\bmu_1-\bmu_2\|^2
    + \mathrm{Tr}\big(
    \bSigma_1+\bSigma_2
    -2(\bSigma_1^{1/2}\bSigma_2\bSigma_1^{1/2})^{1/2}
    \big).
\end{aligned}
\end{equation}
Thanks to the expressions given in \Cref{prop:compute_recursively_cov_mus_algos}, in what follows we will report \textbf{exact Wasserstein errors of each algorithm}  at any time step $t$, that is,  
\begin{equation}
\label{eq:Wasserstein_error_algo}
\begin{aligned}
&\Wass\left(p_t^\mathrm{algo}(\y_t \mid \V) ,	p_t(\x_t \mid \V)\right)
\\
& =
\Wass\left(
	\mathcal{N}(\bmu_t^{\mathrm{algo}},\bSigma_t^{\mathrm{algo}}),
	\mathcal{N}({\bmu}_{t\mid \V},{\bC}_{t\mid \V})
	\right)
\end{aligned}
\end{equation}
the Wasserstein distance between the noisy posterior distributions $p_t^\mathrm{algo}(\y_t \mid \V)$ induced by each algorithm (\Cref{prop:compute_recursively_cov_mus_algos})
and the distributions of the targeted conditional forward process $p_t(\x_t \mid \V)$ (Equation~\eqref{eq:exact_cond_back_density}).

\paragraph{Comparison of the noisy posteriors $p_t^\mathrm{algo}(\y_t \mid \V)$ in toy models} We illustrate the comparison of the different algorithms in 2D and 3D in \Cref{fig:illustration,fig:illustration_2D_v_xt,fig:3d_illustration,fig:3d_illustration_v_xt}. We study the inpainting problem which is conditioning on a noisy part of the coordinates of the Gaussian distribution. {In order to highlight the differences between the algorithms, we consider in this section Gaussian distributions that are not scaled to lie within the usual $[-1,1]$ range commonly used for images.} In these examples, we compare the DPS, $\Pi$GDM, and CGDM algorithms. 
Notably, CGDM aligns perfectly with the true theoretical distribution, even though the 2-Wasserstein distance is not zero. Indeed, we can note that the CGDM algorithm is not exact (by the observation of the 2-Wasserstein distance): it is affected by the incorrectness of the backward process. Theoretically, for a Gaussian distribution, the exact backward process is
\begin{equation}
	\label{eq:theoretical_Gaussian_backward}
    \begin{aligned}
         \ty_T & \sim \tp_T \\
         \ty_t & = \frac{1}{\sqrt{\alpha_t}}\left(\ty_t + \beta_t \nabla \log \tp_t(\ty_t) \right) +  \sqrt{\beta}_t\z_t, \\
         &\quad 1 \leq t \leq T,  \quad\z_t\sim \mathcal{N}(\zero,\bSigma^{-1}_t\bSigma_{t-1}).
    \end{aligned}
\end{equation}
This formula is obtained in Appendix~\ref{appendix:exact_Gaussian_backward}. Consequently, two requirements are not fulfilled: First, the initialization is done with $\ty_T \sim \N$ instead of $\ty_T \sim \tp_T$, which is known as the initialization error and discussed in \cite{pierret_diffusion_models_gaussian_distributions_2024}. Second, the added noise $\z_t$ does not have the correct covariance matrix, it is not supposed to be diagonal. However, the 2-Wasserstein distance induced by these approximations is relatively low.

 \newlength{\smallgraphwidth}
\setlength{\smallgraphwidth}{0.4\linewidth}

\newlength{\biggraphwidth}
\setlength{\biggraphwidth}{\linewidth}

\begin{figure*}
         \centering
         \begin{tabular}{@{}cc@{}c@{}c@{}c@{}c@{}}
&\multicolumn{5}{c}{2-Wasserstein distance along the time}\\
&\multicolumn{5}{c}{\includegraphics[width=\biggraphwidth]{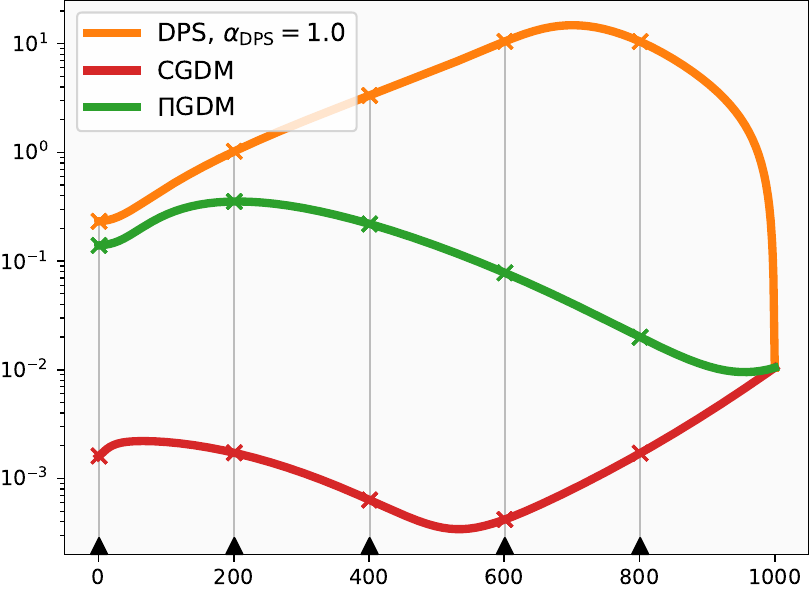}} \\
\midrule
$t$
&
0
&
200
&
400
&
600
& 
800 \\
\midrule
\rotatebox{90}{$\quad p_t^{\mathrm{algo}}(\y_t \mid \V)$}
&
\resizebox{\smallgraphwidth}{\smallgraphwidth}{\includegraphics[width=\smallgraphwidth]{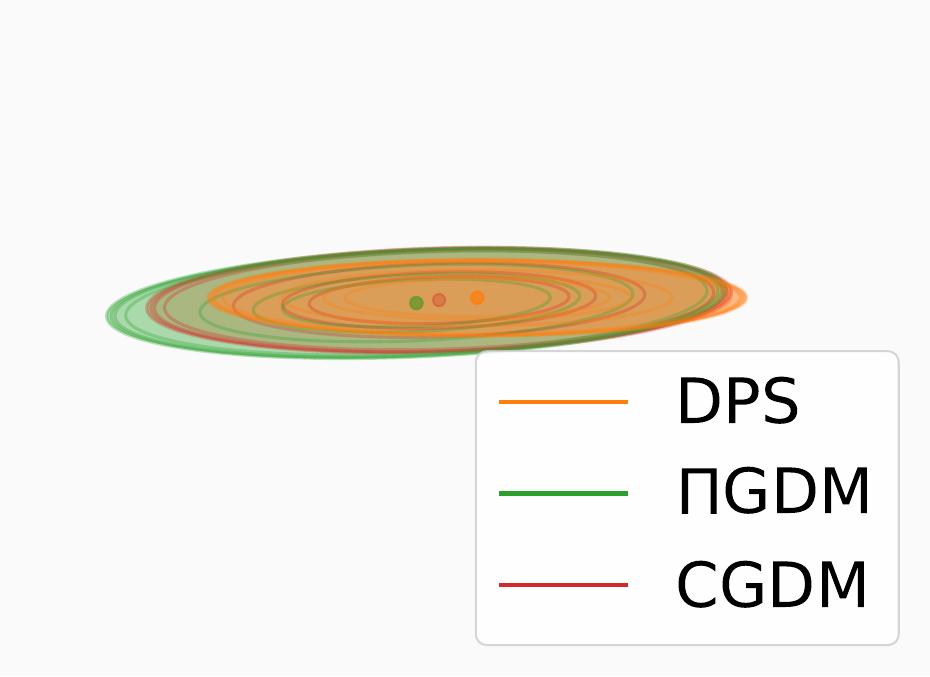} }
&
\resizebox{\smallgraphwidth}{\smallgraphwidth}{\includegraphics[width=\smallgraphwidth]{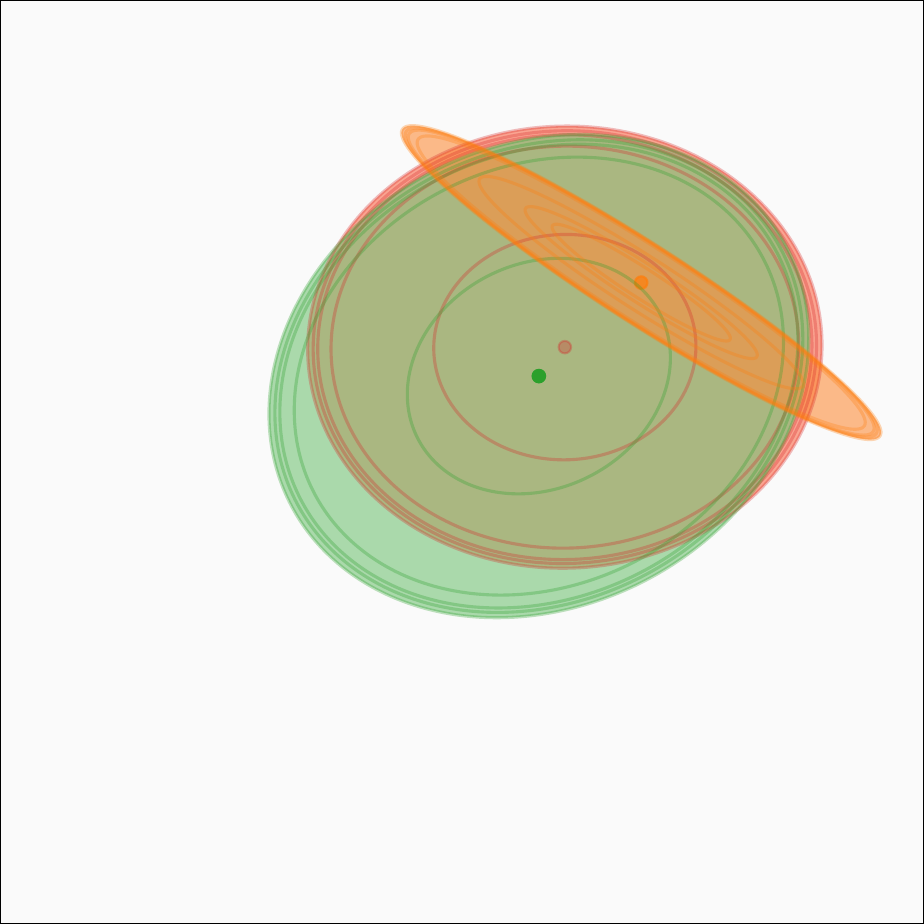} }
&
\resizebox{\smallgraphwidth}{\smallgraphwidth}{\includegraphics[width=\smallgraphwidth]{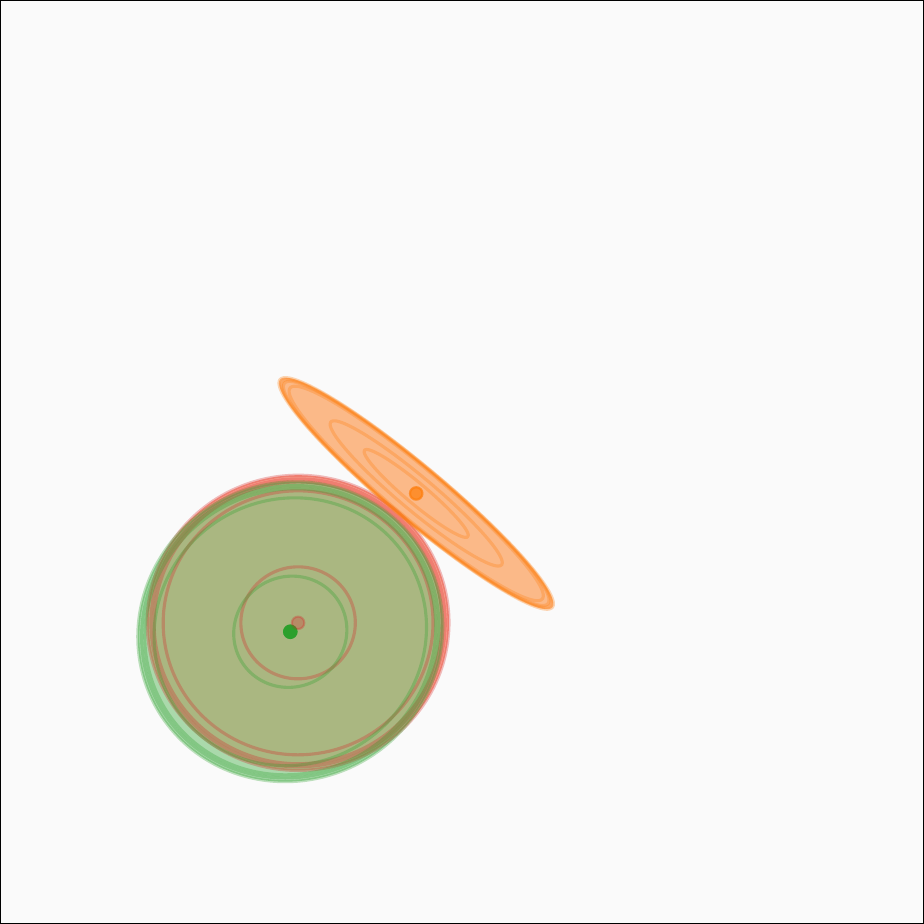} }
&
\resizebox{\smallgraphwidth}{\smallgraphwidth}{\includegraphics[width=\smallgraphwidth]{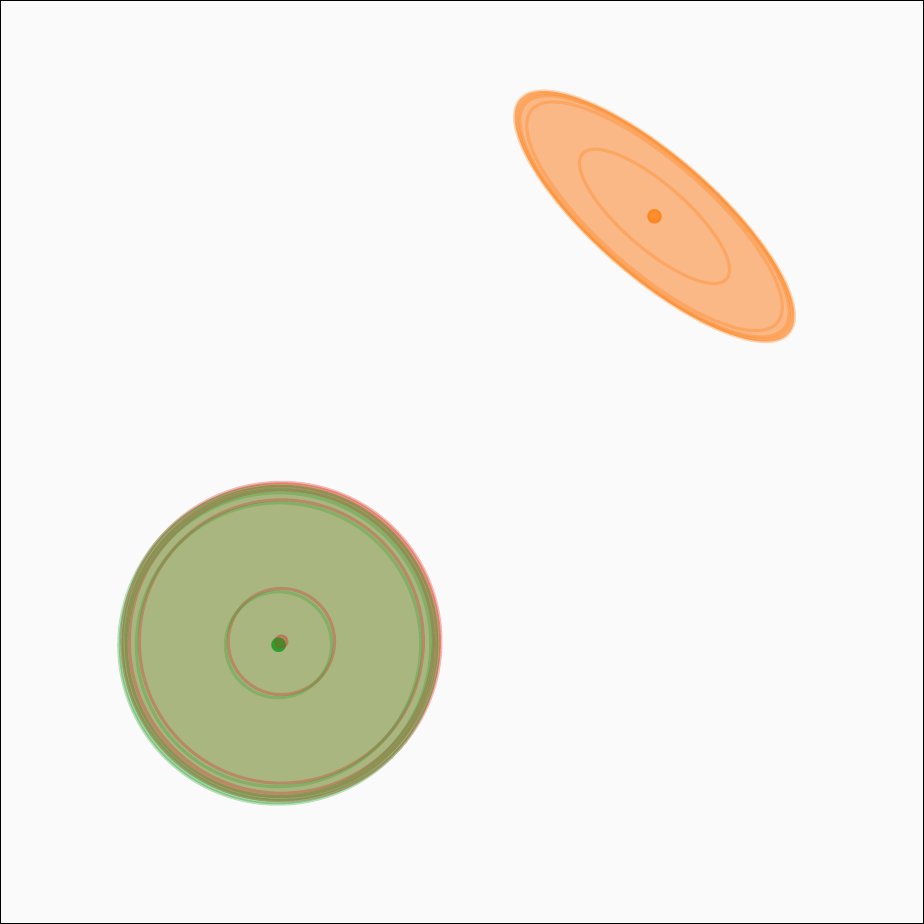} }
&
\resizebox{\smallgraphwidth}{\smallgraphwidth}{\includegraphics[width=\smallgraphwidth]{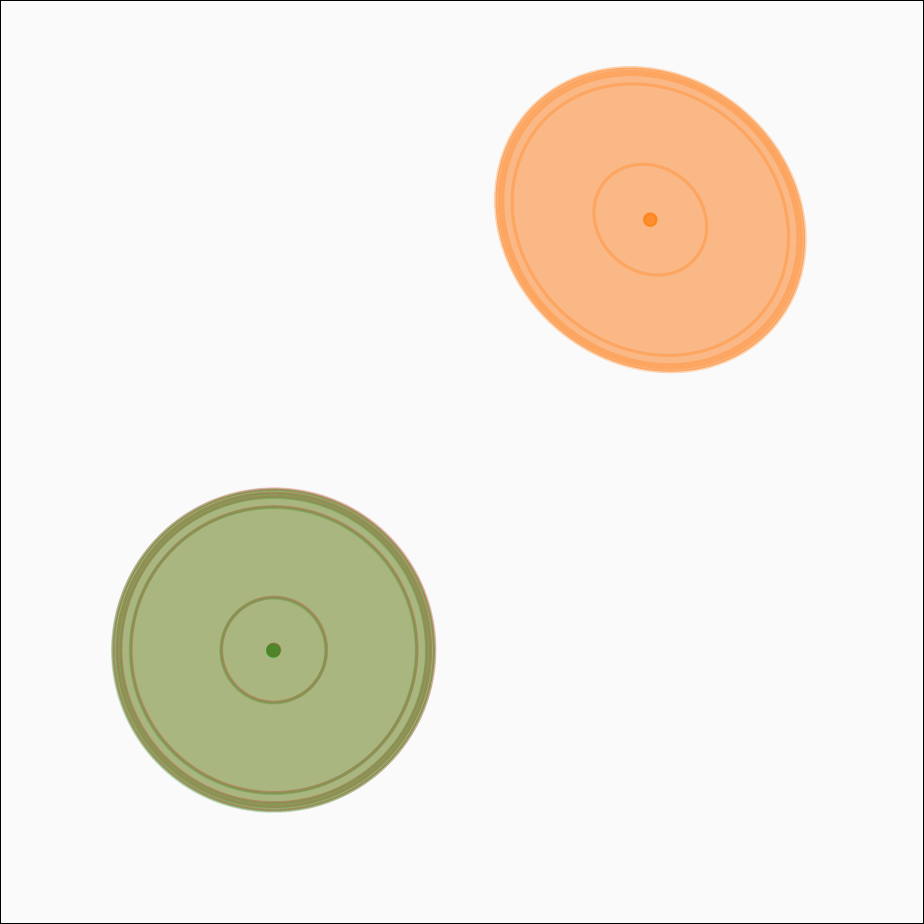} } 
\\
\bottomrule
\end{tabular}
	\caption[Illustration of the algorithms in 2D for the inpainting problem: Focus on the noisy posteriors $p^{\mathrm{algo}}_t(\y_t\mid \V)$.]{\label{fig:illustration} \textbf{Illustration of the algorithms in 2D for the inpainting problem: Focus on the noisy posteriors $p^{\mathrm{algo}}_t(\y_t\mid \V)$.} A 2D Gaussian distribution is conditioned on its first coordinate and noised at level $\sigma = 10/255$. Top: 2-Wasserstein distance along the time (from $t=1000$ to $t = 0$) for the different algorithms with respect to the theoretical forward distribution. Bottom: For different times $t$, we plot the 2D noisy posterior of the algorithms at time $t$.  Note the misalignment of the noisy posterior covariances at each time step.}
\end{figure*}

\begin{figure*}
         \centering
         \begin{tabular}{@{}cc@{}c@{}c@{}c@{}c@{}}
\midrule
$t$
&
0
&
200
&
400
&
600
& 
800 \\
\midrule
\rotatebox{90}{$\quad p^\mathrm{algo}_t(\V \mid \y_t)$}
&
\resizebox{\smallgraphwidth}{\smallgraphwidth}{\includegraphics[width=\smallgraphwidth]{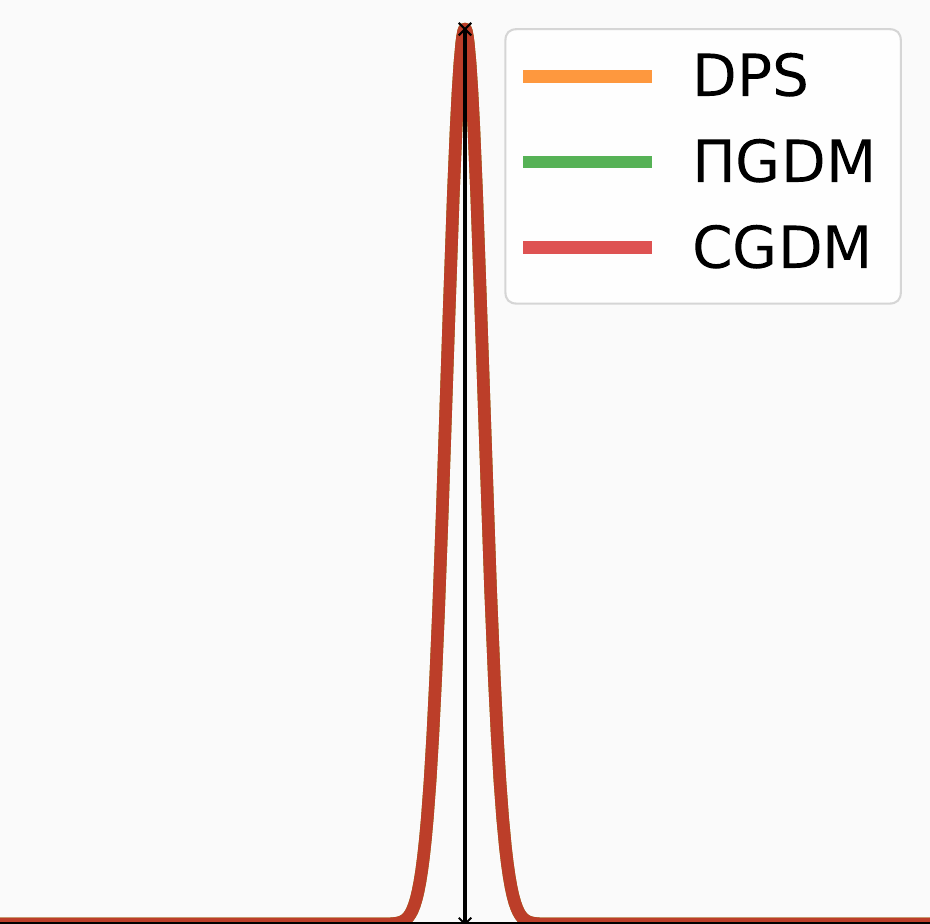}} 
&
\resizebox{\smallgraphwidth}{\smallgraphwidth}{\includegraphics[width=\smallgraphwidth]{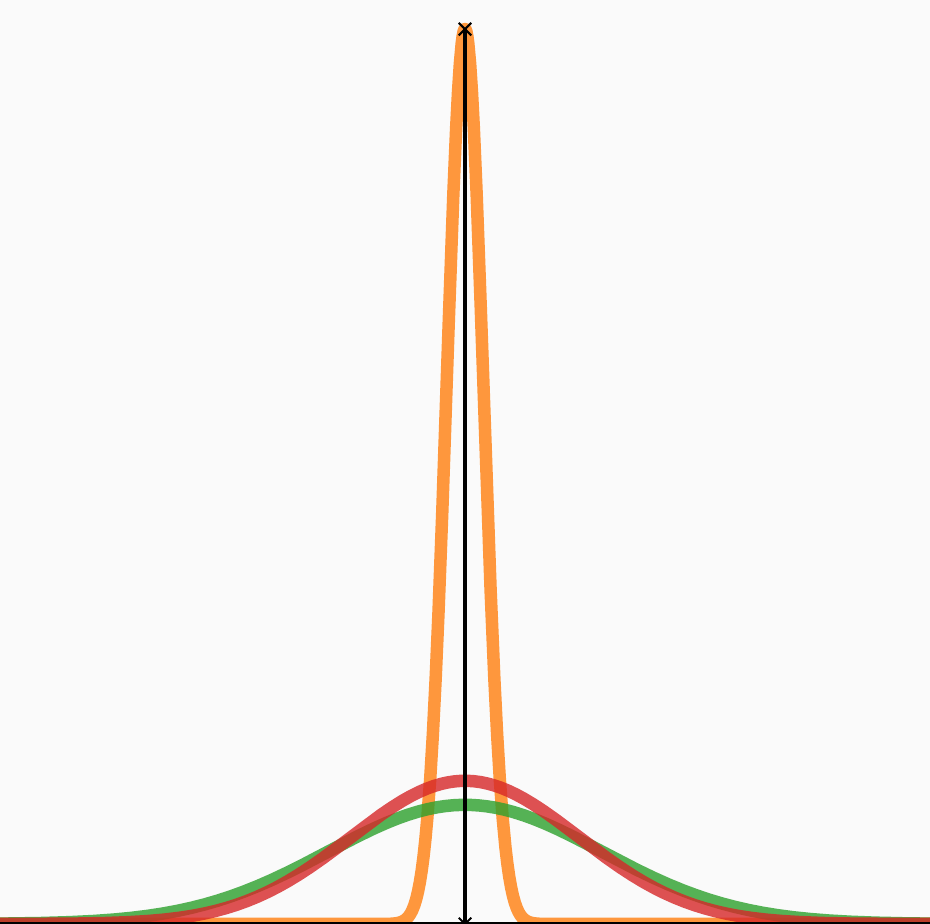}} 
&
\resizebox{\smallgraphwidth}{\smallgraphwidth}{\includegraphics[width=\smallgraphwidth]{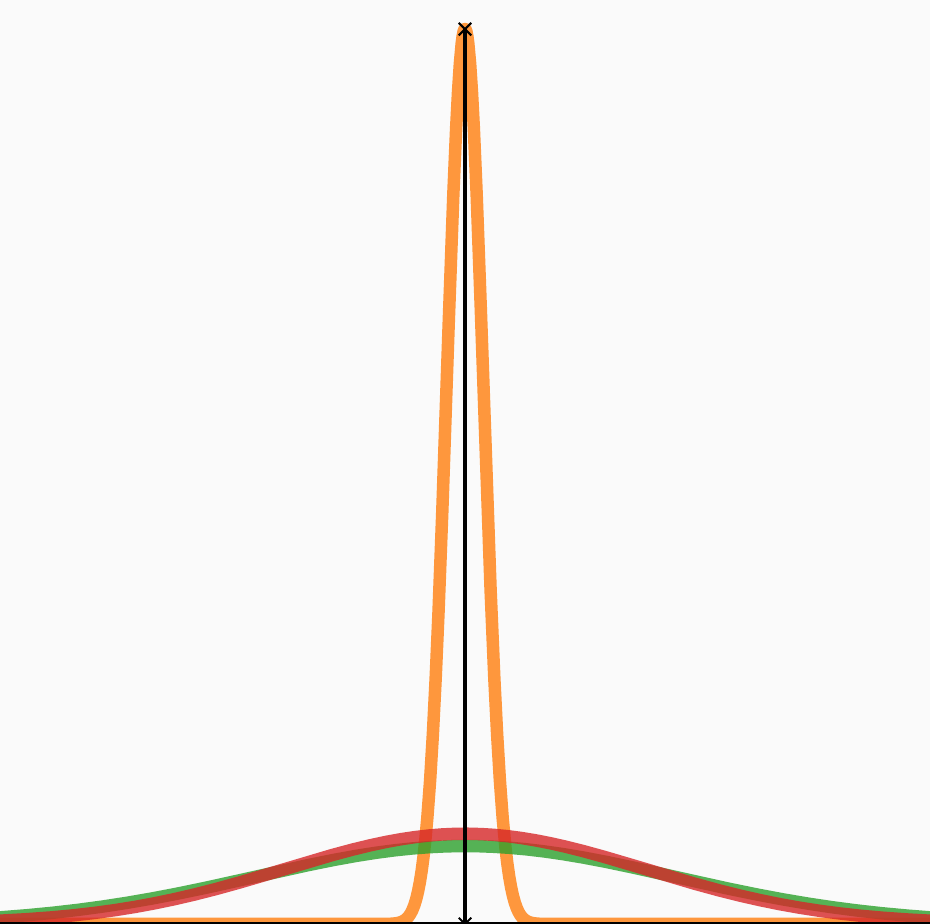}} 
&
\resizebox{\smallgraphwidth}{\smallgraphwidth}{\includegraphics[width=\smallgraphwidth]{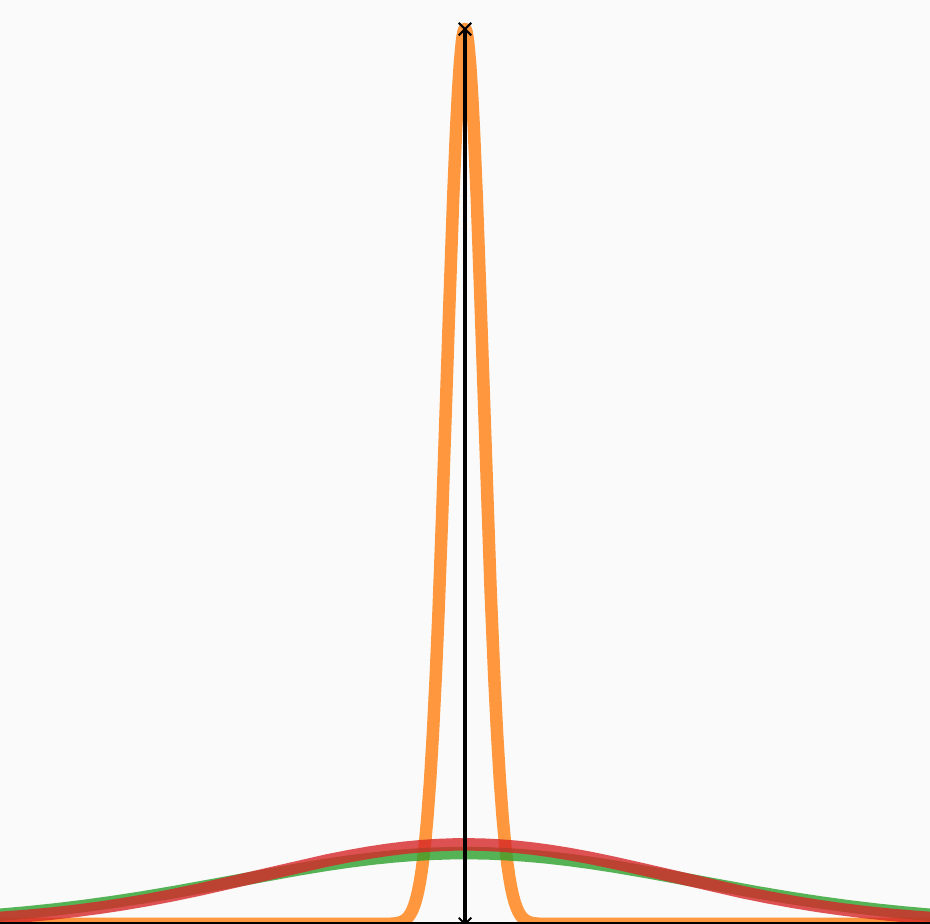}} 
&
\resizebox{\smallgraphwidth}{\smallgraphwidth}{\includegraphics[width=\smallgraphwidth]{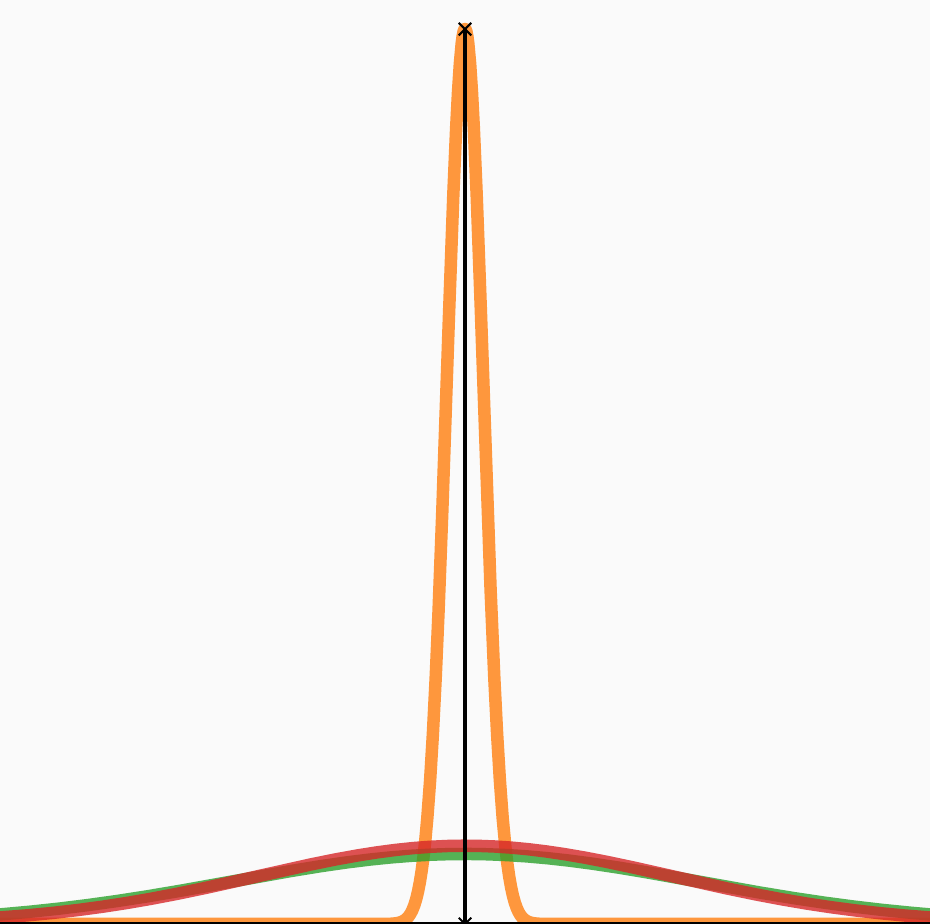}} \\ 
\bottomrule
\end{tabular}
	\caption[Illustration of the algorithms in 2D for the inpainting problem: Focus on the noisy likelihoods $p^{\mathrm{algo}}_t(\V \mid \y_t)$.]{\label{fig:illustration_2D_v_xt} \textbf{Illustration of the algorithms in 2D for the inpainting problem: Focus on the noisy likelihoods $p^{\mathrm{algo}}_t(\V \mid \y_t)$.} For different selected times $t$, we plot the 1D distribution model of $p^{\mathrm{algo}}_t(\V \mid \y_t)$, related to \Cref{fig:illustration}. Note that DPS underestimates the variance and that all three algorithms coincide at $t = 0$.}
\end{figure*}

\setlength{\smallgraphwidth}{0.37\linewidth}
\begin{figure*}
         \centering
         \begin{tabular}{@{}cc@{}c@{}c@{}c@{}c@{}c@{}}
&\multicolumn{5}{c}{2-Wasserstein distance along the time}\\
&\multicolumn{5}{c}{\includegraphics[width=\biggraphwidth]{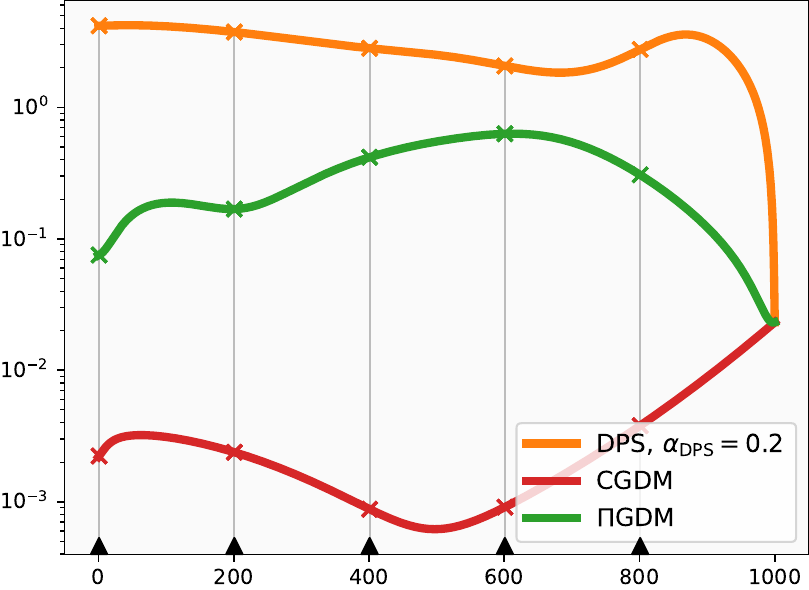}} \\
\midrule
$t$
&
0
&
200
&
400
&
600
&
800 \\
\midrule
\rotatebox{90}{\parbox{\smallgraphwidth}{\centering Projected  \\ $p^\mathrm{algo}_t(\y_t \mid \V)$ \\ on the unknown \\ coordinate}}
&
\resizebox{\smallgraphwidth}{\smallgraphwidth}{\includegraphics[width=\smallgraphwidth]{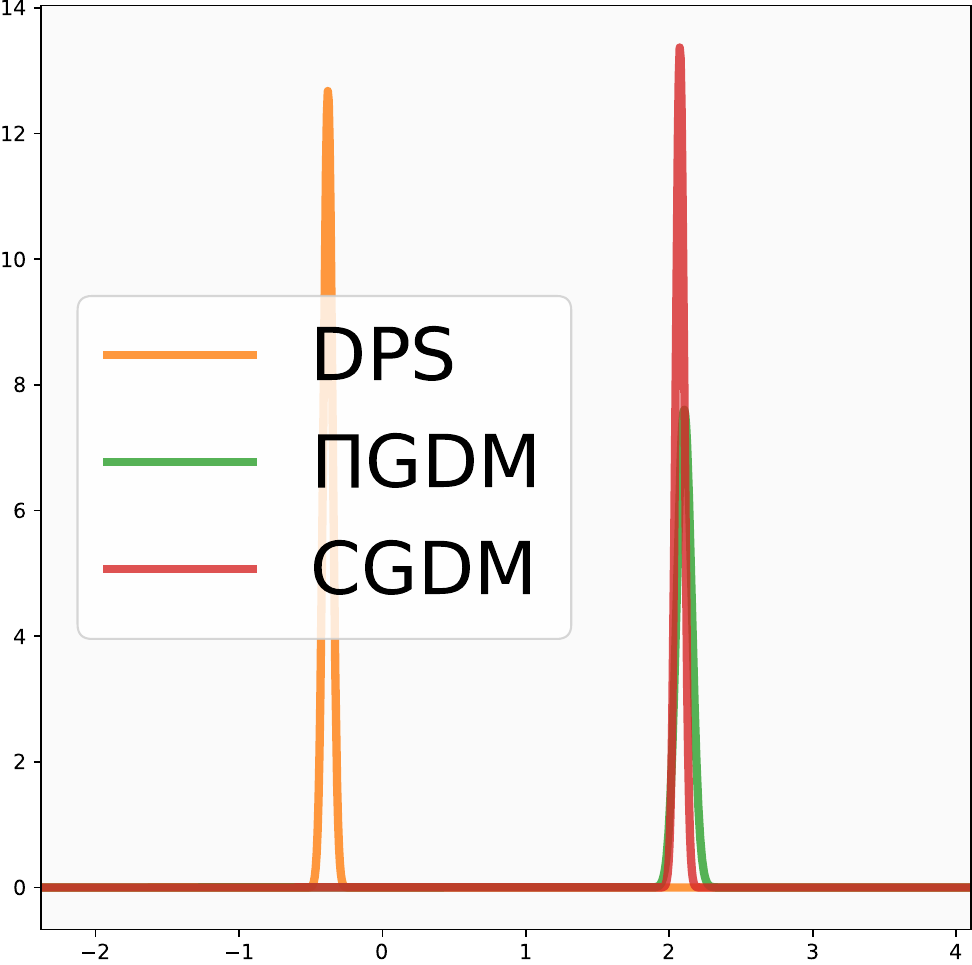} }
&
\resizebox{\smallgraphwidth}{\smallgraphwidth}{\includegraphics[width=\smallgraphwidth]{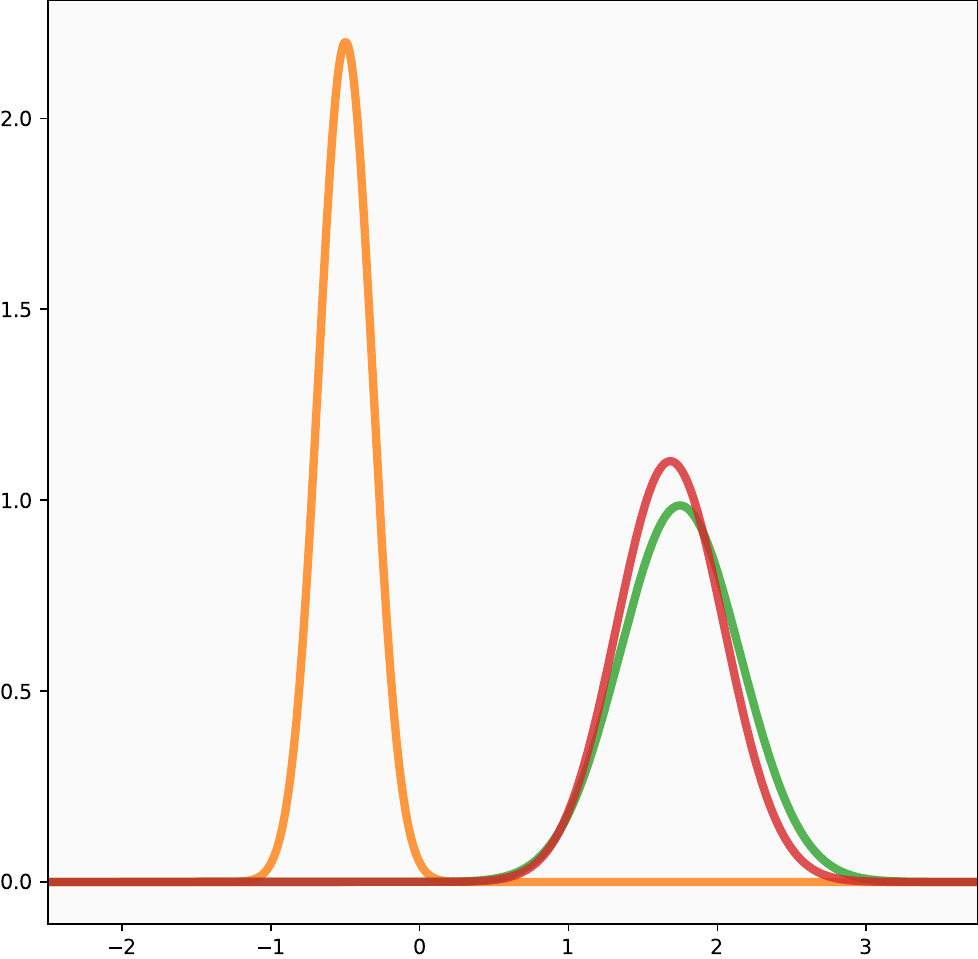} }
&
\resizebox{\smallgraphwidth}{\smallgraphwidth}{\includegraphics[width=\smallgraphwidth]{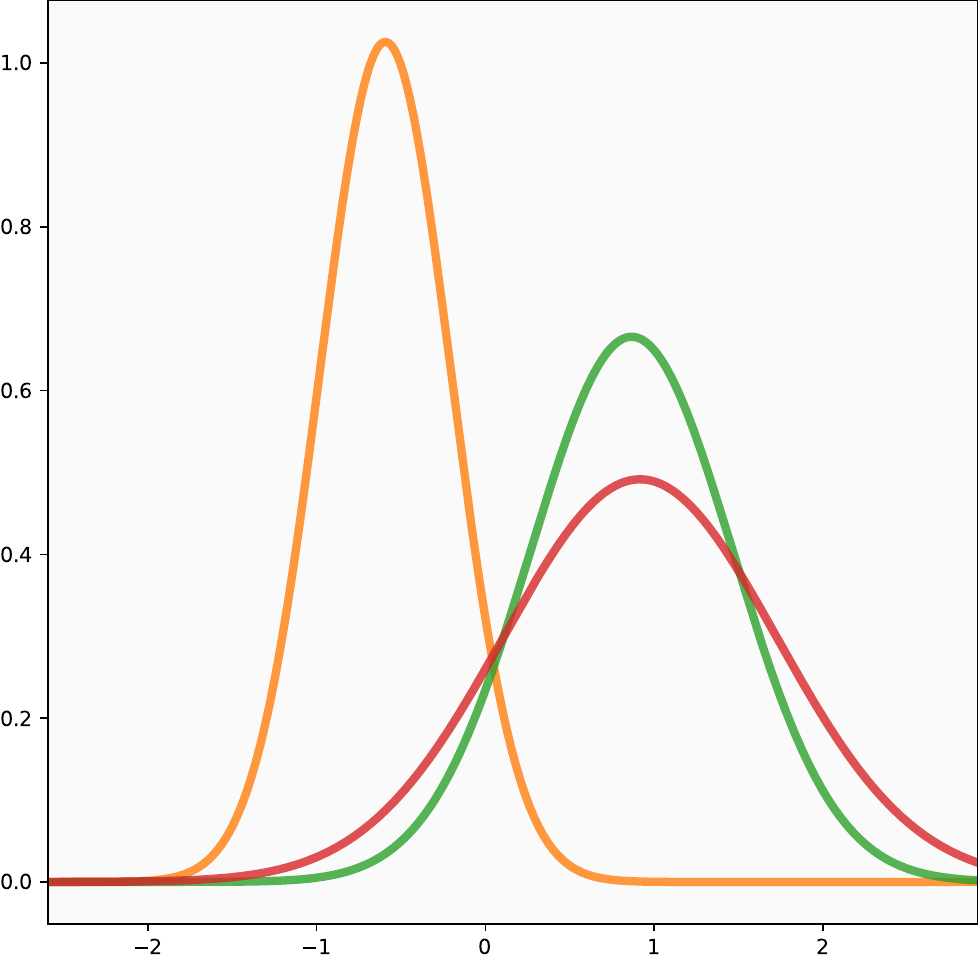} }
&
\resizebox{\smallgraphwidth}{\smallgraphwidth}{\includegraphics[width=\smallgraphwidth]{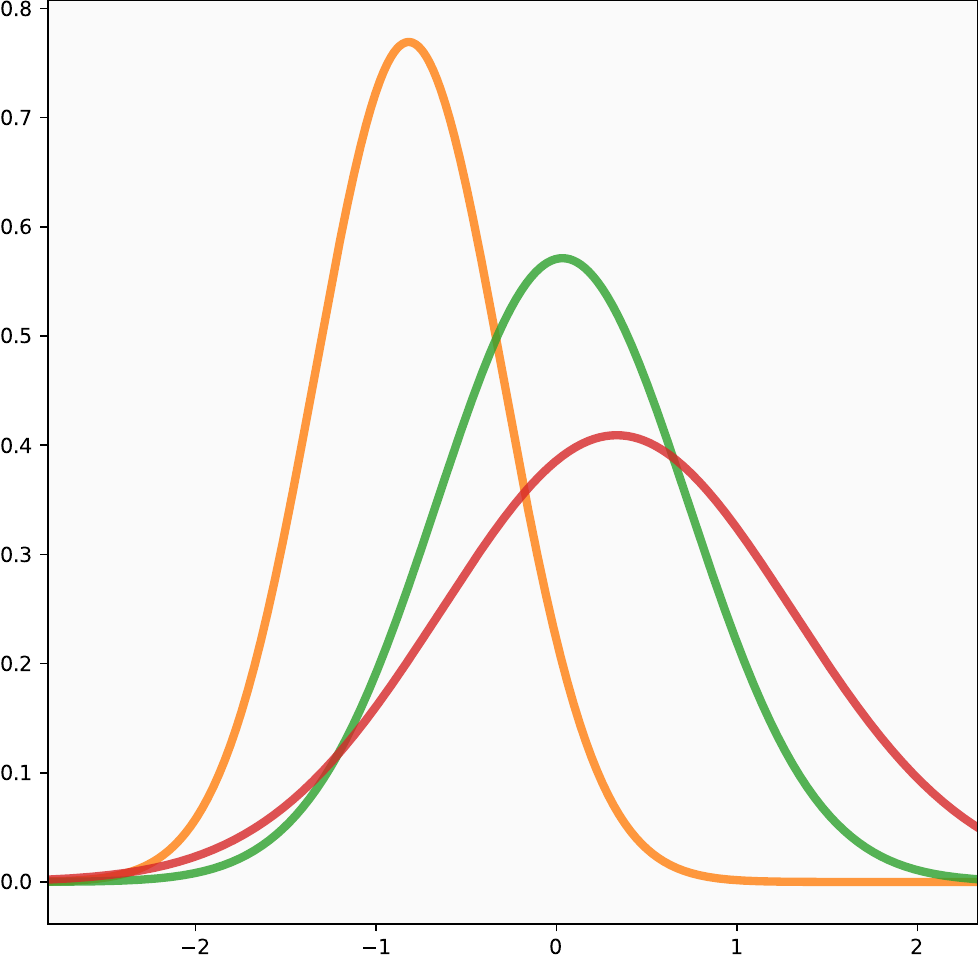} }
&
\resizebox{\smallgraphwidth}{\smallgraphwidth}{\includegraphics[width=\smallgraphwidth]{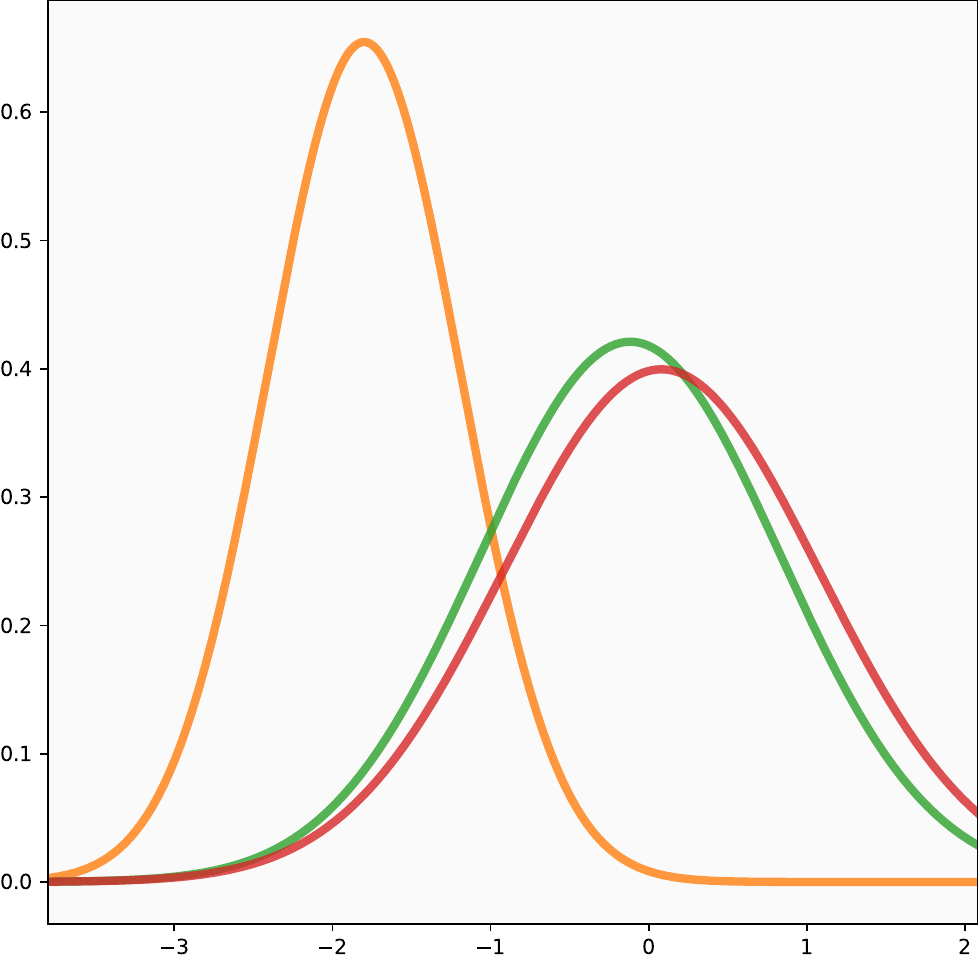} } \\
\bottomrule
\end{tabular}
	\caption[Illustration of the algorithms in 3D for the inpainting problem: Focus on the noisy posteriors.]{\label{fig:3d_illustration}\textbf{Illustration of the algorithms in 3D for the inpainting problem: Focus on the the noisy posteriors $p^{\mathrm{algo}}_t(\y_t\mid \V)$.} A 3D Gaussian distribution is conditioned on its two first coordinates and noised at level $\sigma = 10/255$. Top: 2-Wasserstein distance along the time for the different algorithms with respect to to the theoretical forward distribution. Bottom: For different times $t$, we plot the 1D algorithms' backward distribution of the unknown coordinate at time $t$. We can observe the bias introduced by the DPS and $\Pi$GDM algorithms over time.}
\end{figure*}
\setlength{\smallgraphwidth}{0.4\linewidth}
\begin{figure*}[t]
         \centering
         \begin{tabular}{@{}cc@{}c@{}c@{}c@{}c@{}}
\midrule
$t$
&
0
&
200
&
400
&
600
& 
800 \\
\midrule
\rotatebox{90}{\parbox{\smallgraphwidth}{\centering $p^\mathrm{algo}_t(\V \mid \y_t)$}}
&
\resizebox{\smallgraphwidth}{\smallgraphwidth}{\includegraphics[width=\smallgraphwidth]{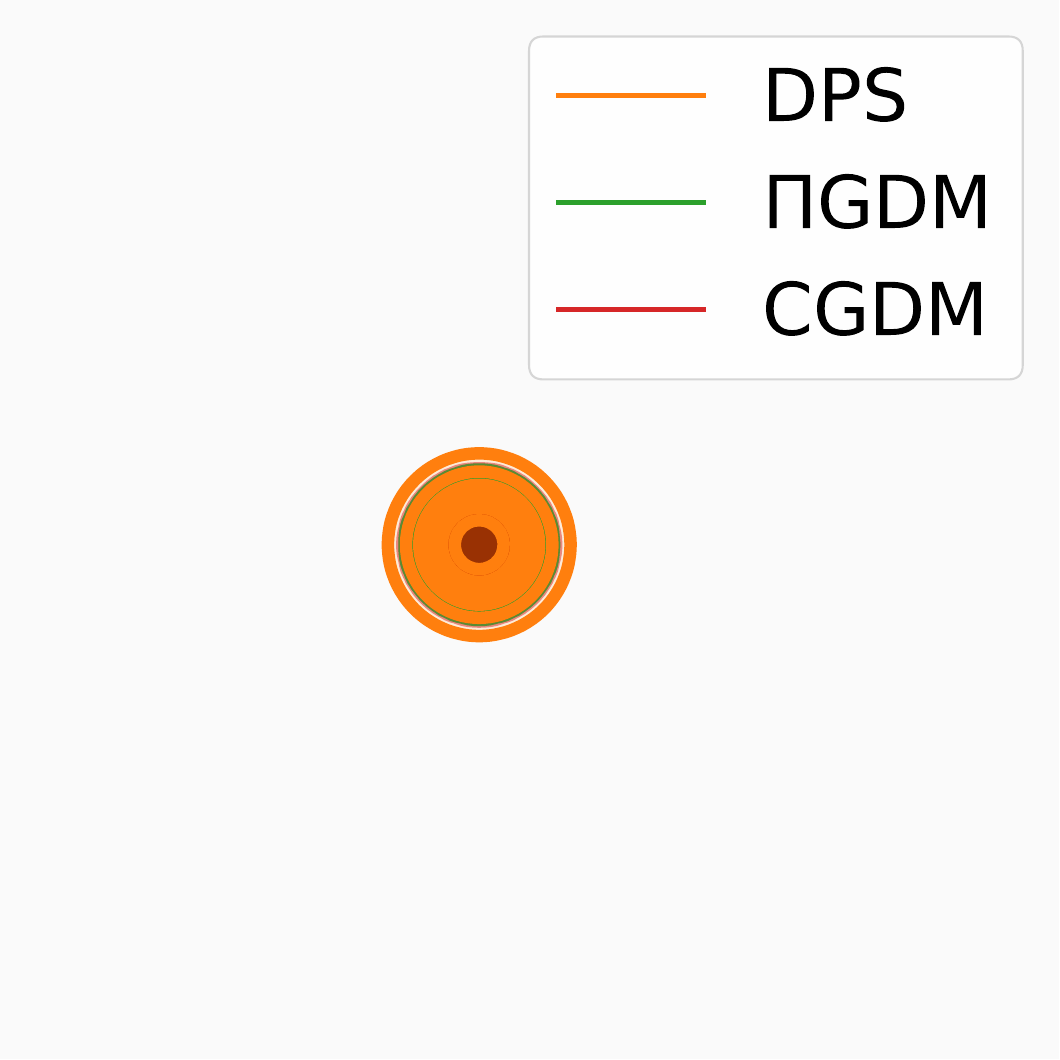}}
&
\resizebox{\smallgraphwidth}{\smallgraphwidth}{\includegraphics[width=\smallgraphwidth]{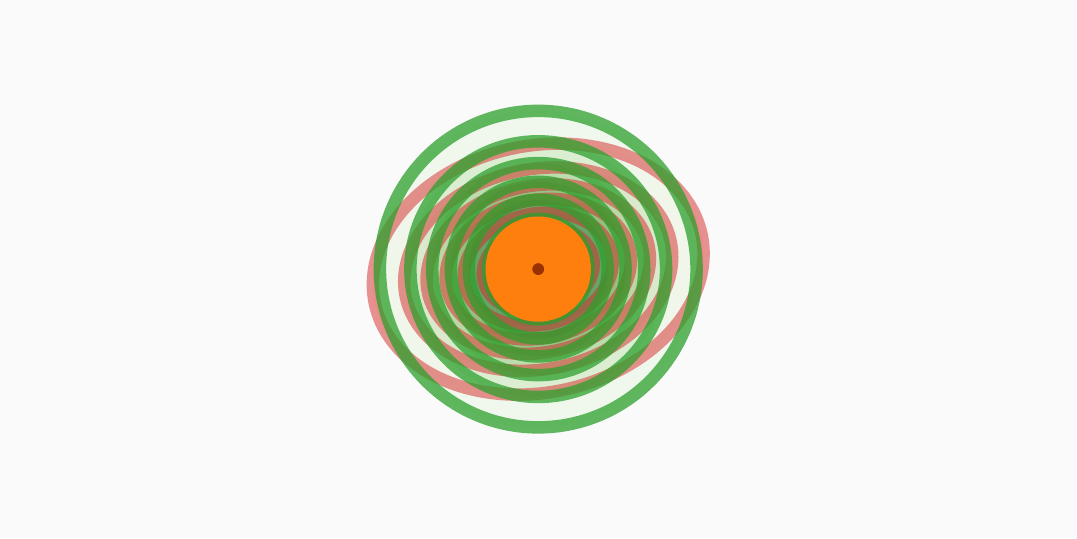}}
&
\resizebox{\smallgraphwidth}{\smallgraphwidth}{\includegraphics[width=\smallgraphwidth]{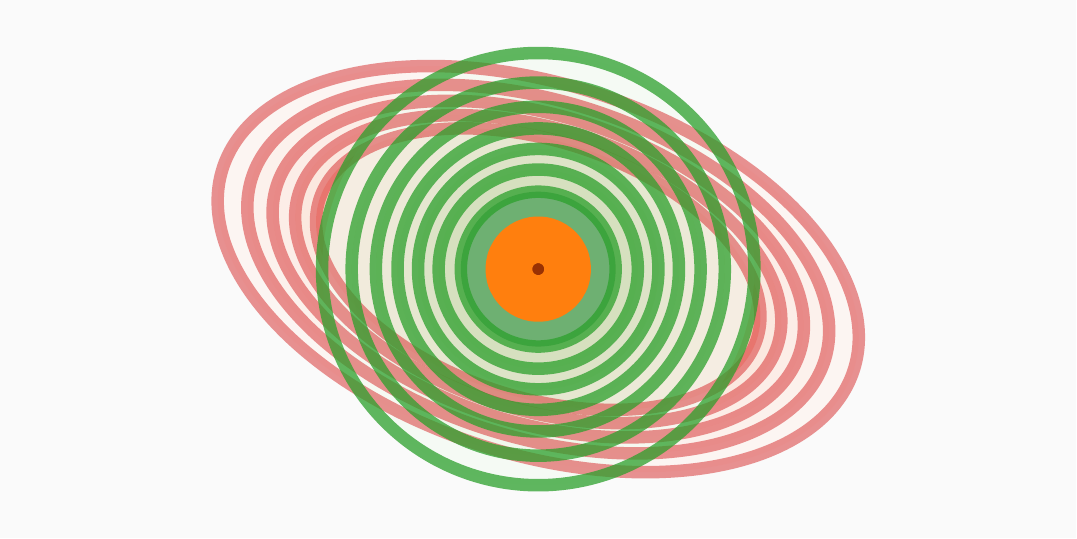}}
&
\resizebox{\smallgraphwidth}{\smallgraphwidth}{\includegraphics[width=\smallgraphwidth]{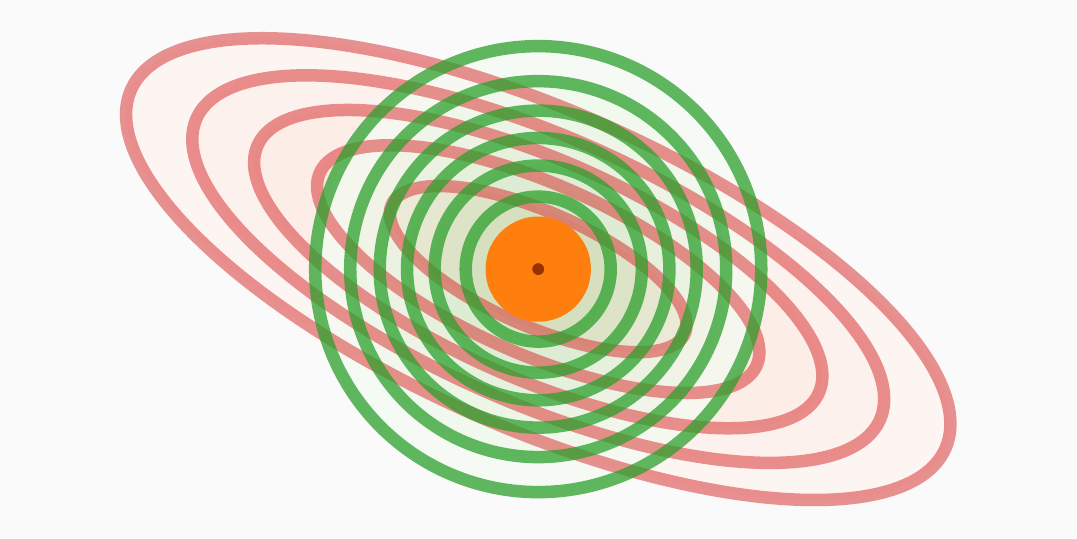}}
&
\resizebox{\smallgraphwidth}{\smallgraphwidth}{\includegraphics[width=\smallgraphwidth]{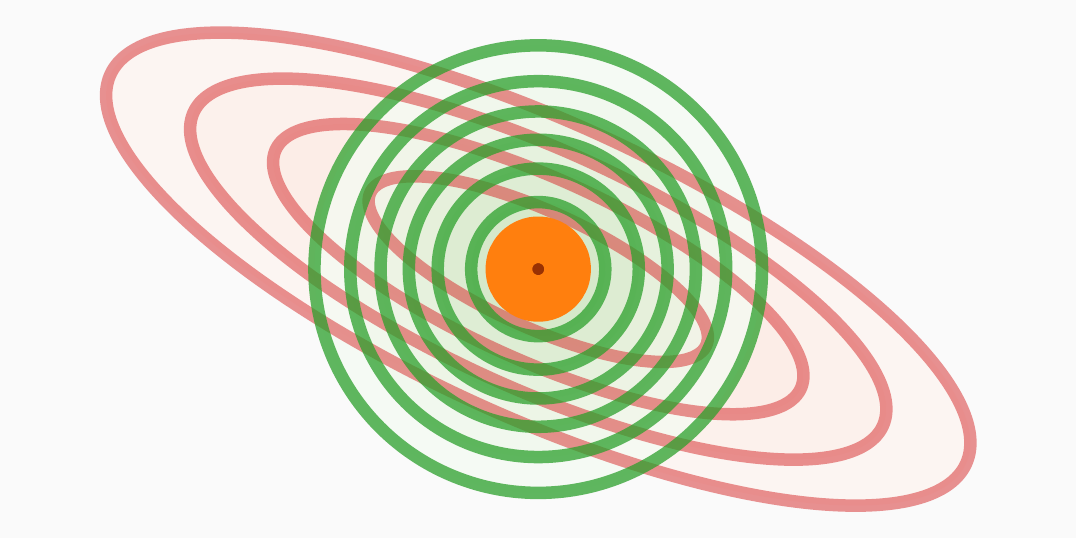}} \\ 
\bottomrule
\end{tabular}
	\caption[Illustration of the algorithms in 3D for the inpainting problem: Focus on the noisy likelihoods $p^{\mathrm{algo}}_t(\V\mid \y_t)$.]{\label{fig:3d_illustration_v_xt}\textbf{Illustration of the algorithms in 3D for the inpainting problem: Focus on the noisy likelihood $p^{\mathrm{algo}}_t(\V\mid \y_t)$.} A 3D Gaussian distribution is conditioned on its two first coordinates and noised at level $\sigma = 10/255$. For different times $t$, we plot the 2D distribution model of $p^{\mathrm{algo}}_t(\V \mid \y_t)$, related to \Cref{fig:3d_illustration}. The different algorithms exhibit alignment near the final time $t = 0$.}
\end{figure*}

 In \Cref{fig:illustration}, the distributions along the time of the algorithms (2D bottom graphs) show that the DPS backward distribution moves into the space with unexact mean and covariance estimations along the time. The $\Pi$GDM algorithm is very faithful to the theoretical backward process in terms of mean but has not a perfect covariance information. These two facts are observable in the 2-Wasserstein distance graph: {the 2-Wasserstein distance for CGDM remains consistently low, within the range of $10^{-3}$ to $10^{-2}$, while $\Pi$GDM varies between $10^{-2}$ and $10^{-1}$. In contrast, DPS shows significantly higher deviation, reaching values above $10^1$, highlighting its instability and divergence from the true posterior distribution.} Similar observations can be made in \Cref{fig:3d_illustration}. Note that DPS is tested with $\alpha_{\text{\tiny DPS}} = 0.2$ because it becomes unstable for higher values, although $\alpha_{\text{\tiny DPS}} = 1$ would be the natural choice. At the final time $t=0$, both DPS and $\Pi$GDM exhibit a non-zero bias.

The ability of the DPS and $\Pi$GDM algorithms to exhibit a decreasing error near the final time $t = 0$ can be understood from the remarks in \Cref{sec:comp_algo_under_Gaussian_assumption}, as the expressions of the noisy posteriors for the different algorithms closely approximate the true noisy posterior when $t$ is close to zero.

\paragraph{Comparison of the estimated noisy likelihood $p_t^\mathrm{algo}(\V\mid\y_t)$} Each algorithm is distinguished by its modeling choice for the covariance matrix of the noisy likelihood $p_t^\mathrm{algo}(\V \mid \y_t)$ (see \Cref{tab:comparison_expression_grad_cond}). In \Cref{fig:3d_illustration_v_xt,fig:illustration_2D_v_xt}, we compare these choices.
More precisely, since $p_t(\V \mid \y_t)$ depends on both $\y_t$ and $\V$, where $\V$ is related to $\x_0$ via \Cref{eq:inverse_problem}, we proceed as follows: we first fix a sample $\y_T$ drawn from the distribution $p_T$ and compute the corresponding backward trajectory $(\y_t)_{0 \leq t \leq T}$. Then, we generate a noisy observation $\V$ of the associated sample $\y_0$, and we plot $p_t^\mathrm{algo}(\V \mid \y_t)$ at selected time steps. This allows us to visualize the model for $\bC^\mathrm{algo}_{\V \mid t}$ defined by each algorithm.
In \Cref{fig:illustration_2D_v_xt}, we observe that the variance of $p_t(\V \mid \x_t)$ is significantly underestimated by the DPS algorithm. This may explain the instabilities observed in higher-dimensional settings: in this algorithm, the gradient term $\nabla_{\y_t}\|\A\widehat{\x}_0(\y_t)\|^2$ is divided by the low variance $\sigma^2$, amplifying its magnitude.
Similarly, in \Cref{fig:3d_illustration_v_xt}, the DPS algorithm again severely underestimates the variance and also introduces a substantial bias. The $\Pi$GDM algorithm also suffers from inaccuracies in modeling $p_t(\V \mid \x_t)$, as its covariance model $\bC^{{\Pi\mathrm{GDM}}}_{\V \mid t}$ does not incorporate the true covariance structure.
However, the three algorithms are aligned at $t = 0$, as also discussed in \Cref{subsec:exact_Gaussian_formulas}.

\section{Scaling to higher dimension}

\label{sec:study_case_deblurring}

Previously, we compared the three algorithms on toy examples, where the underlying distributions are both computable and directly observable, typically through the visualization of Gaussian covariance ellipses. However, these models are ultimately intended for the generation of large-scale images, and we now aim to investigate the behavior of the algorithms in this high-dimensional setting.
The main difficulties in such an analysis are twofold: first, the absence of a tractable ground-truth distribution against which the algorithms can be compared; second, the impossibility of computing the exact covariance matrices associated with the distributions manipulated by the algorithms.
Indeed, even storing such covariance matrices becomes prohibitive. For RGB images of resolution \(256 \times 256\), the covariance matrix has size
$
(3 \times 256 \times 256)^2 > 3 \times 10^{10}.
$
As a consequence, applying these matrices without additional structural assumptions is computationally infeasible, which in turn prevents the exact computation of Wasserstein distances.
To address these issues, we focus on inverse problems involving stationary Gaussian textures. First, a kriging-based framework has been developed in the literature to solve certain inverse problems exactly \cite{Galerne_Leclaire_gaussian_inpainting_2017_siims, pierret_stochastic_superresolution_gaussian_microtextures_2024}. Second, the stationarity assumption enables a compact representation and storage of the associated covariance matrices, as detailed below.

\subsection{Stationary Gaussian textures and kriging}

We consider the asymptotic discrete spot noise (ADSN) distribution~\cite{Galerne_Gousseau_Morel_random_phase_textures_2011_IEEE} associated with an RGB texture $\U\in\R^{3\times \OMN}$, with $\OMN = \{0,\ldots,M-1\}\times \{0,\ldots,N-1\}$, which is defined as the stationary Gaussian distribution whose covariance equals the autocorrelation of $\U$. More precisely, this distribution is sampled using convolution with a white Gaussian noise:
Denoting $m\in\R^3$ the channelwise mean of $\U$ and $\bt_c = \frac{1}{\sqrt{MN}}(\U_c-m_c)$, $1 \leq c \leq 3$, its associated \emph{texton}, for $\w \sim \N$ of size $M \times N$
the channelwise convolution 
\begin{equation}
\label{eq:sample_ADSN}
    \x = m + \bt \star \w \in \R^{3 \OMN}
\end{equation} follows $\ADSN(\U)$.
 This distribution is the Gaussian $\mathcal{N}(m,\bSigma)$ where $\bSigma$ is induced by an RGB convolution operator i.e. in the form $\bSigma = \bC_t \bC_t^T \in \R^{3\OMN \times \OMN}$ with $\bC_t :=\begin{pmatrix}
	\bC_{\bt_1}^T 
	&\bC_{\bt_2}^T 
	&\bC_{\bt_3}^T
\end{pmatrix}^T \in \R^{ \OMN \times  3\OMN}$ where $\bC_{t_c} \in \R^{\OMN \times \OMN}$ is the convolution by the $c$-th channel of the texton $\bt$ for $1 \leq c \leq 3$. This correlation is induced by the fact that the white noise consider in \Cref{eq:sample_ADSN} is the same on each channel. In the Fourier domain, for $\x \in \R^{3 \times \OMN}$, $\xi \in \OMN$, $1 \leq c \leq 3$,
\begin{equation}
	\widehat{\bSigma x}_c(\xi) = \widehat{\bt}_c(\xi) \sum_{j=1}^3 \overline{\widehat{\bt}}_j(\xi) \widehat{\x}(\xi) = \widehat{\bt}(\xi)[\overline{\widehat{\bt}}(\xi)]^T \widehat{\x}(\xi)
\end{equation}
where $\widehat{\x}$ is the Fourier transform of $\x$ and $\widehat{\x}(\xi) := (\begin{smallmatrix}
	\widehat{\x}_1(\xi) 
	& \widehat{\x}_2(\xi) 
	& \widehat{\x}_3(\xi) 
\end{smallmatrix})^T \in \R^3$. Consequently, the matrix $\bSigma$ acts as a rank-1 3D matrix on each frequency $\xi \in \OMN$ of $\x$ in the Fourier domain. Denoting by $\widehat{\bSigma}(\xi)$ the action of the matrix $\bSigma$ in the Fourier basis on the frequency $\xi$, we can write \cite{Xia_synthesizing_mixing_Gaussian_texture_models_2014_SIAM}
\begin{equation}
	\widehat{\bSigma}(\xi) = \widehat{\bt}(\xi)\left[\widehat{\bt}(\xi)\right]^T.
\end{equation}
We can provide the eigenvalues of $\bSigma$, that are $\left(\sum_{i=1}^3 \left|\bt_i(\xi)\right|^2\right)_{\xi \in \OMN}$ and $0$ with multiplicity $2MN$.
 The score matrix $\bSigma_t = \alphabar_t \bSigma + (1-\alphabar_t)\I$ has the same structure as $\bSigma$ and we can write 
\begin{equation}
\label{eq:score_matrix_3d}
	\widehat{\bSigma}_t(\xi) = \alphabar_t \widehat{\bt}(\xi)\left[\widehat{\bt}(\xi)\right]^T + (1-\alphabar_t)\I_3.
\end{equation}
{As already done in \cite{pierret_diffusion_models_gaussian_distributions_2024}, the score can be exactly applied in the context of Gaussian microtextures. The operations are detailed in Appendix~\ref{appendix:ADSN_score_application}}.
As a consequence, we are able to implement a diffusion model with an exact score on ADSN microtextures, as illustrated in Figure~\ref{fig:uncond_samples_ADSN}. This direction is explored in the remainder of this section to analyze the DPS, $\Pi$GDM and CGDM algorithms in the context of high-dimensional inverse problems. The covariance matrix structure will allow us to efficiently compute 2-Wasserstein distances in the context of deblurring.

\setlength{\tabcolsep}{1pt}
\newlength{\uncondsamplewidth}
\setlength{\uncondsamplewidth}{0.19\linewidth}
\begin{figure}
\centering
	\begin{tabular}{ccccc}
	\small $t=1000$
	&
	\small$t=200$
	&
	\small$t=50$
	&
	\small$t=10$
	&
	\small$t=0$
	\\
	\includegraphics[width=\uncondsamplewidth]{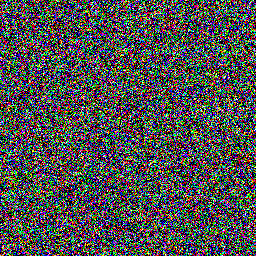}
	&\includegraphics[width=\uncondsamplewidth]{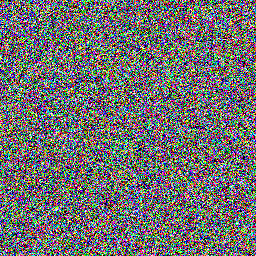}
	&\includegraphics[width=\uncondsamplewidth]{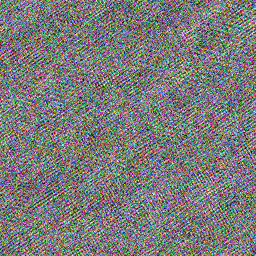}
	&\includegraphics[width=\uncondsamplewidth]{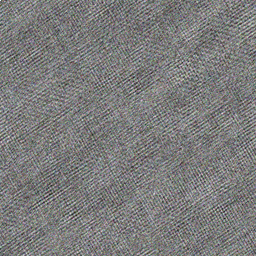}
	&\includegraphics[width=\uncondsamplewidth]{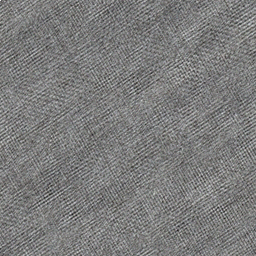} \\
	\includegraphics[width=\uncondsamplewidth]{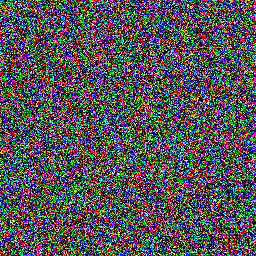}
	&\includegraphics[width=\uncondsamplewidth]{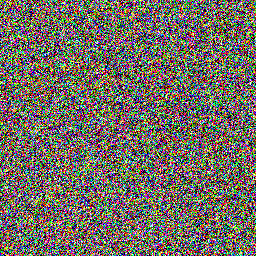}
	&\includegraphics[width=\uncondsamplewidth]{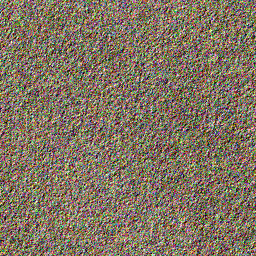}
	&\includegraphics[width=\uncondsamplewidth]{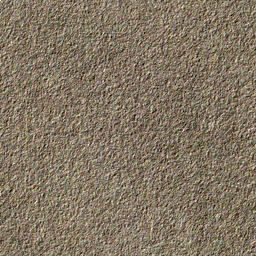}
	&\includegraphics[width=\uncondsamplewidth]{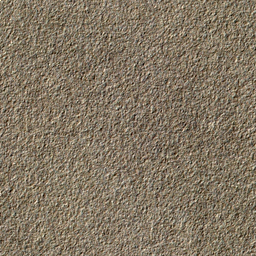}	
	\end{tabular}
	\caption[Illustration of the application of a DDPM on ADSN microtextures, with an exact score function.]{\label{fig:uncond_samples_ADSN} Illustration of the application of a DDPM on ADSN microtextures, with an exact score function.}
\end{figure}

\subsection{Study of the deblurring problem: Exact computation of the 2-Wasserstein distance}

The degradation operator $\A$ of the deblurring inverse problem is a channelwise convolution by a given blur kernel $\bc \in \R^{\OMN}$. In the following, we denote by $\bC$, the block diagonal RGB convolution for which $\bC_{\bc}$ is on each block and where $\bc \in \R^{\OMN}$ is a blur kernel. In other terms, $\bC$ applies the same convolution by the blur kernel $\bc$ on each channel of an image. We focus on three blur kernels: the zoom-out bicubic kernel with a factor $r=16$, which is the convolution part of the super-resolution (SR) problem and two motion blur kernels, generated by an online code\footnote{\url{https://github.com/LeviBorodenko/motionblur}} used in \cite{Chung_DPS_ICLR_2023}. 

\setlength{\tabcolsep}{1pt}
\newlength{\blurkerwidth}
\setlength{\blurkerwidth}{0.24\linewidth}

Exploiting that the degradation operator associated with the deblurring inverse problem preserves the structure of the covariance matrix of the ADSN models given in \Cref{eq:score_matrix_3d} we can state the following proposition.

\begin{prop}[Simultaneous diagonalizability of the Gaussian backward processes associated with the different algorithms]
\label{prop:simult_diago_cov_backwards}
For the deblurring problem involving ADSN microtextures, the covariance matrices associated with the forward process $({\bC}_{t\mid \V})_{0 \leq t \leq T}$ and the backward processes of the different algorithms—$(\bSigma_t^\mathrm{DPS})_{0 \leq t \leq T}$, $(\bSigma_t^{\Pi\mathrm{GDM}})_{0 \leq t \leq T}$, and $(\bSigma_t^\mathrm{CGDM})_{0 \leq t \leq T}$—are all diagonalizable in the same orthogonal basis as $\bSigma$.
\end{prop}
\begin{proof}
	See Appendix~\ref{appendix:proof_proposition_simultaneous_diagonalizable}
\end{proof}

\Cref{prop:simult_diago_cov_backwards} is key to computing efficiently the exact 2-Wasserstein distances associated with the different algorithms.
Indeed, if $\bSigma_1$ and $\bSigma_2$ are simultaneously diagonalizable, with respective eigenvalues $\left(\lambda_{i,1}\right)_{1\leq i \leq d}$ and $\left(\lambda_{i,2}\right)_{1\leq i \leq d}$, then Equation~\eqref{eq:W2_Gaussian} boils down to
\begin{equation}
\label{eq:W2_Gaussian_diagonalizable}
\begin{aligned}
	&\Wass^2\left(\mathcal{N}(\bmu_1,\bSigma_1),\mathcal{N}(\bmu_2,\bSigma_2)\right) \\
	& =
    \|\bmu_1-\bmu_2\|^2
	+ \sum_{1\leq i \leq d}
    \left(
    \sqrt{\lambda_{i,1}}
    -\sqrt{\lambda_{i,2}}
    \right)^2.
\end{aligned}
\end{equation}
To compute the 2-Wasserstein distance~\eqref{eq:W2_Gaussian_diagonalizable}, it suffices to compute the eigenvalues of the involved covariance matrices $(\bSigma_t^\mathrm{DPS})_{0 \leq t \leq T}$, $(\bSigma_t^{\Pi\mathrm{GDM}})_{0 \leq t \leq T}$, and $(\bSigma_t^\mathrm{CGDM})_{0 \leq t \leq T}$ at each time step, as done in~\cite{pierret_diffusion_models_gaussian_distributions_2024} for the continuous non conditional case. This approach ensures fast and efficient computation of these distances.
However, to allow a fine experimental comparison between the three algorithms under study, one needs to further decompose the Wasserstein error into two parts, as detailed by the following proposition.

\begin{prop}[Orthogonal decomposition of the deblurring Wasserstein errors]
\label{prop:cut_Wass_in_two_parts}
For the deblurring problem involving ADSN microtextures, the backward processes of the different algorithms are all compatible with the orthogonal decomposition $\R^d = \ker \bSigma \oplus \left(\ker \bSigma\right)^\perp$ (in the sense that each projected component are independent Gaussian processes).
As a consequence, for each $\algo \in \{\mathrm{DPS},\Pi\mathrm{GDM},\mathrm{CGDM}\}$,
the square of the Wasserstein distance of the $t$-th iteration of the algorithm wrt to the true conditional Gaussian algorithm decomposes into
	\begin{equation}
		\begin{aligned}
	&\Wass^2\left(p_t^\mathrm{algo}(\y_t \mid \V) ,	p_t(\x_t \mid \V)\right) 
	\\
	& = \left(\WasskerS^\mathrm{algo}\right)^2+\left(\WassorthokerS^\mathrm{algo}\right)^2.
		\end{aligned}
	\end{equation}
The first component $\WasskerS^\mathrm{algo}$ of the error is the same for each algorithm.
\end{prop}
\begin{proof}
	See Appendix~\ref{appendix:proof:prop:cut_Wass_in_two_parts}.
\end{proof}

The $\WasskerS^\mathrm{algo}$ part of the error is due to the fact that the DDPM scheme is unable to retrieve the rank of $\bSigma$, as observed for the discrete sampling schemes in \cite{pierret_diffusion_models_gaussian_distributions_2024}. 
Note that this error can be managed by modifying the noise schedule, as explored in \cite{Strasman_analysis_noise_schedule_SGM_2025_TMLR}.
In view of Proposition~\ref{prop:cut_Wass_in_two_parts}, 
in what follows we only report the $\WassorthokerS^\mathrm{algo}$ component of the Wassertein error for a better comparison of the algorithms.

\subsection{Numerical study of the deblurring problem}

We can now observe the exact error $\WassorthokerS^\mathrm{algo}$ defined by Proposition~\ref{prop:cut_Wass_in_two_parts} for the different algorithms DPS, $\Pi$GDM, CGDM for three different blur kernels examples with measurement noise level $\sigma = 10/255$, as illustrated in \Cref{fig:W2_SR16,fig:W2_blur1,fig:W2_blur6}.


The hyperparameter $\alpha_{\mathrm{DPS}}$ respects the following rule: the lowest stable value provides the lowest 2-Wasserstein distance and we use it in  \Cref{fig:W2_SR16,fig:W2_blur1,fig:W2_blur6}.
The algorithms are ranked in this order in terms of performance: CGDM, $\Pi$GDM and DPS along the time. The superior performance of CGDM was expected in this Gaussian setting. For the bicubic kernel and each texture, the $\Pi$GDM is quite close to the perfect CGDM algorithm at the end of the process while DPS presents a high 2-Wasserstein error along all the iterations. For the two motion blur kernels, $\Pi$GDM stays relatively far from the true conditional algorithm.

Samples shown in \Cref{fig:samples_algorithms_Fabric3_motion_blur1_sig_10,fig:samples_algorithms_fabric_motion_blur1_sig_10} for two texture examples show that the samples generated by the different algorithms seem very similar.
{However,} the corresponding mean demonstrates that the two algorithms are significantly biased, especially the DPS one. \Cref{fig:means_Fabric_SR16}  shows the mean along the backward process for one texture example and, as observed in our toy examples in 2D and 3D, the DPS and $\Pi$GDM algorithms are biased, and DPS more severely.

We compare the models of the noisy likelihood covariance of $p_t(\V \mid \y_t)$ in \Cref{fig:ker_Fabric_motion_blur1} for the first motion kernel, for the first fabric texture. We can observe that the constant DPS is really far from the true theoretical distribution while $\Pi$GDM approximation becomes less harmful along the time which can be explained by our empirical observations in \Cref{sec:comp_algo_under_Gaussian_assumption}.

However, the covariance conditional distribution are very close for the three algorithms for lower times, as illustrated in \Cref{fig:cond_distrib_Fabric_SR16}{, which is related to our observations in \Cref{sec:comp_algo_under_Gaussian_assumption}: For $t$ close to $0$, the algorithms converge towards the correct conditional distribution $p_0(\cdot \mid \V)$.}

\setlength{\tabcolsep}{1pt}
\newlength{\obswidth}
\setlength{\obswidth}{0.4\linewidth}
\begin{figure*}[t]
\centering
\begin{tabular}{*{5}{>{\centering\arraybackslash}p{\obswidth}}}
Blur kernel
&
Image $\U$
&
DFT of the texton $\bt$ 
&
Blurred image $\V$ 
&
$\WassorthokerS^\mathrm{algo}$
\\
	\includegraphics[width=\obswidth]{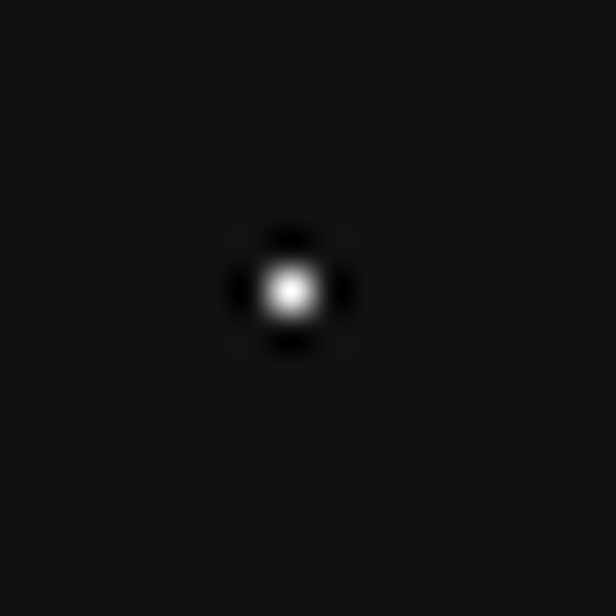}
	&
	\includegraphics[width=\obswidth]{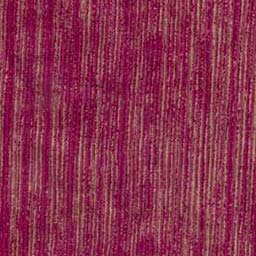}
	&
	\includegraphics[width=\obswidth]{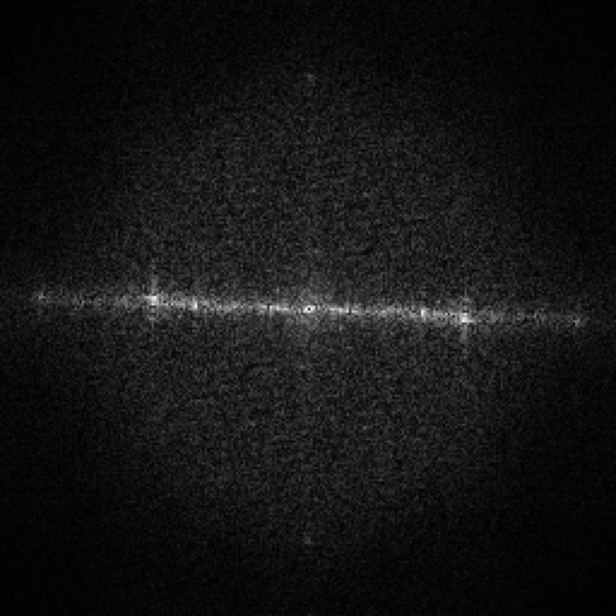}
	&
	\includegraphics[width=\obswidth]{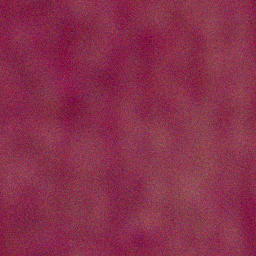}
	&
	\includegraphics[width=\obswidth]{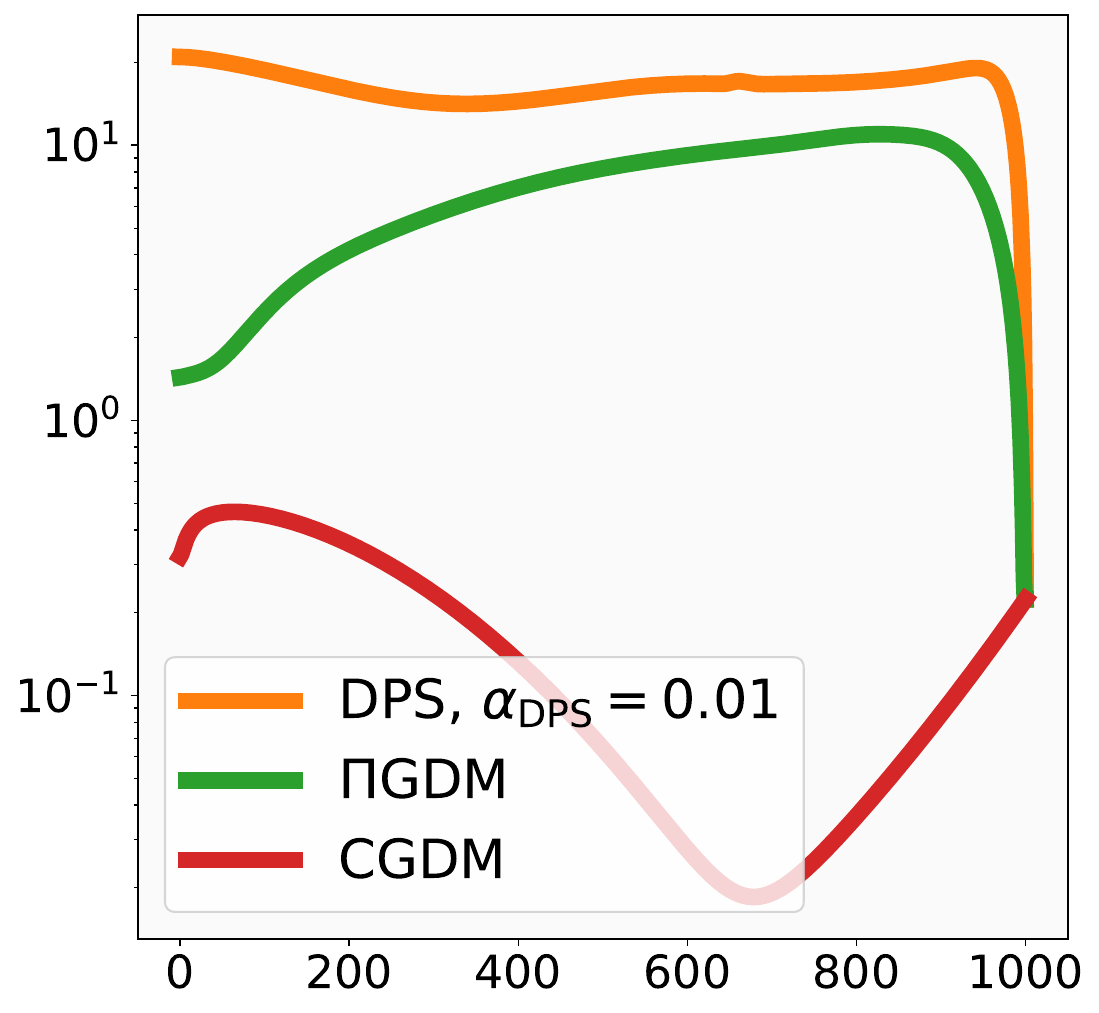} \\
	&
	\includegraphics[width=\obswidth]{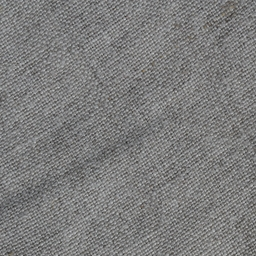}
	&
	\includegraphics[width=\obswidth]{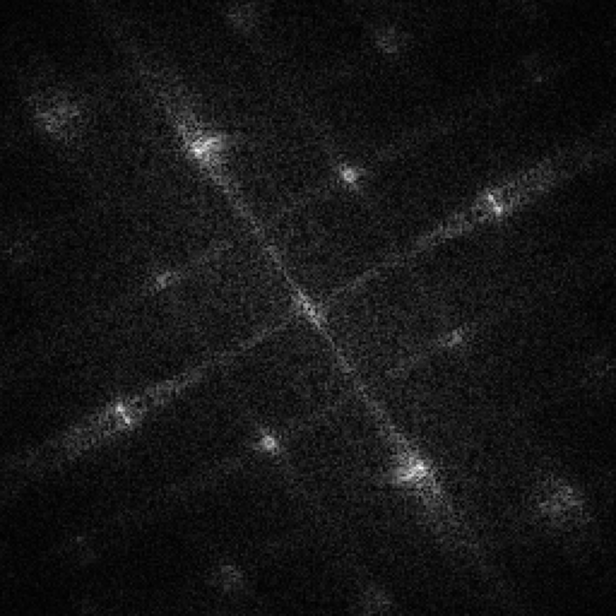}
	&
	\includegraphics[width=\obswidth]{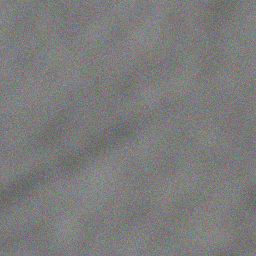}
	&
	\includegraphics[width=\obswidth]{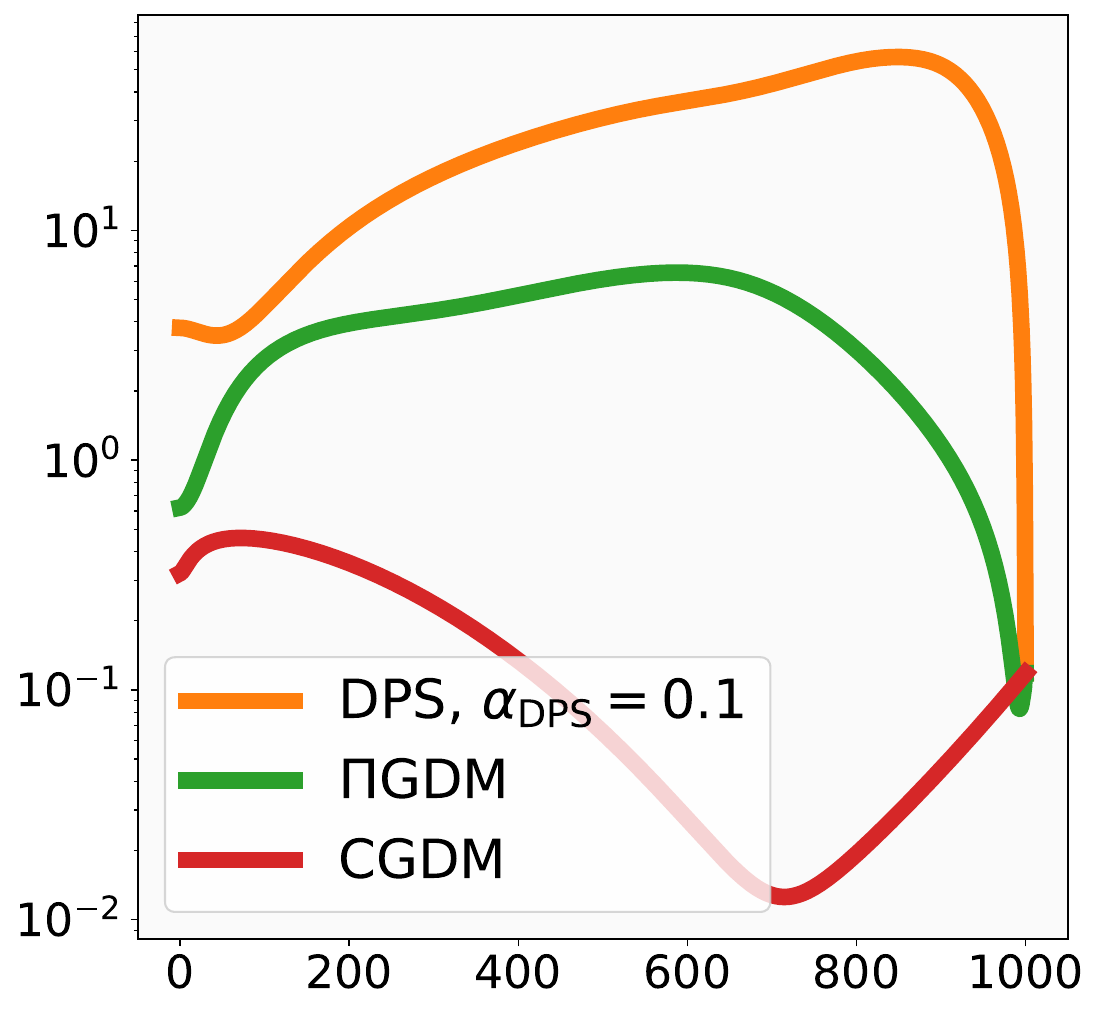} \\
	&
	\includegraphics[width=\obswidth]{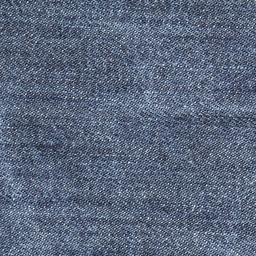}
	&
	\includegraphics[width=\obswidth]{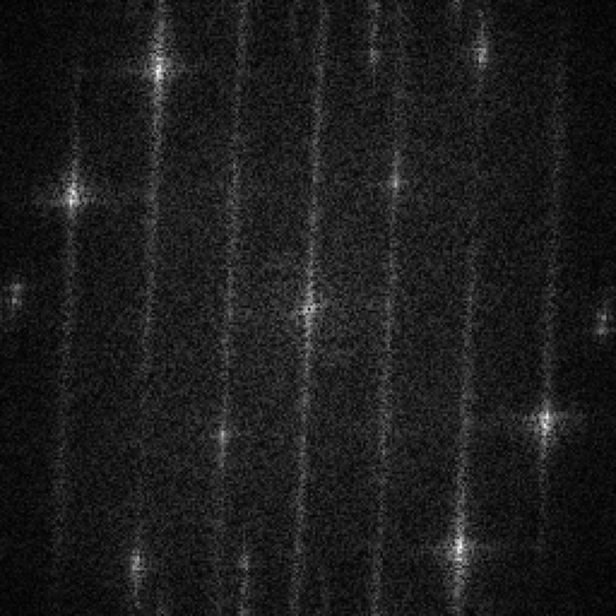}
	&
	\includegraphics[width=\obswidth]{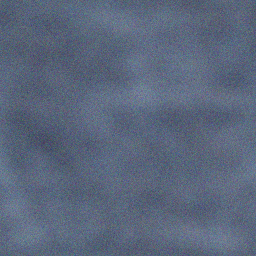}
	&
	\includegraphics[width=\obswidth]{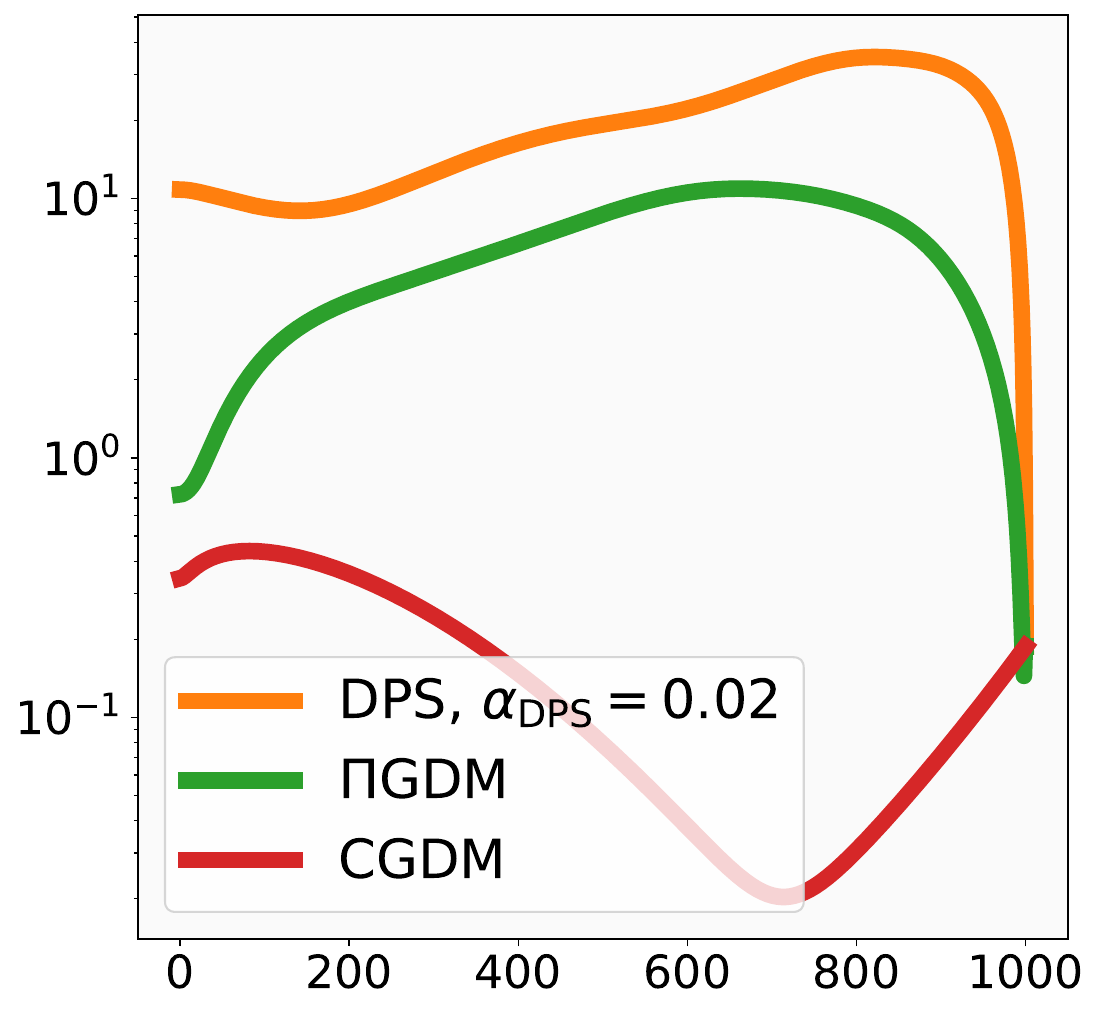} \\
	&
	\includegraphics[width=\obswidth]{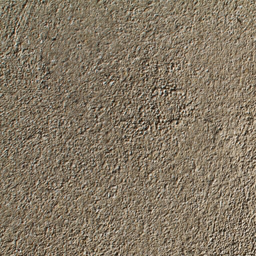}
	&
	\includegraphics[width=\obswidth]{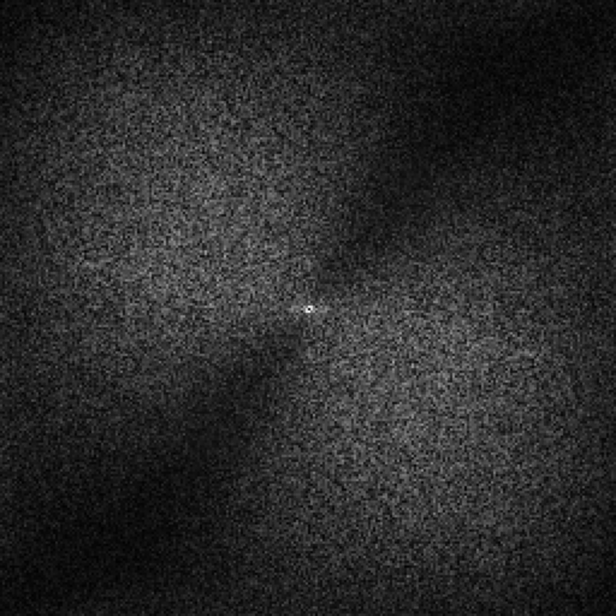}
	&
	\includegraphics[width=\obswidth]{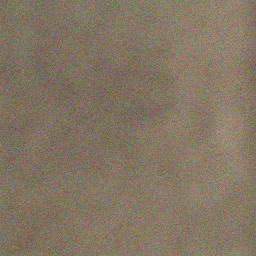}
	&
	\includegraphics[width=\obswidth]{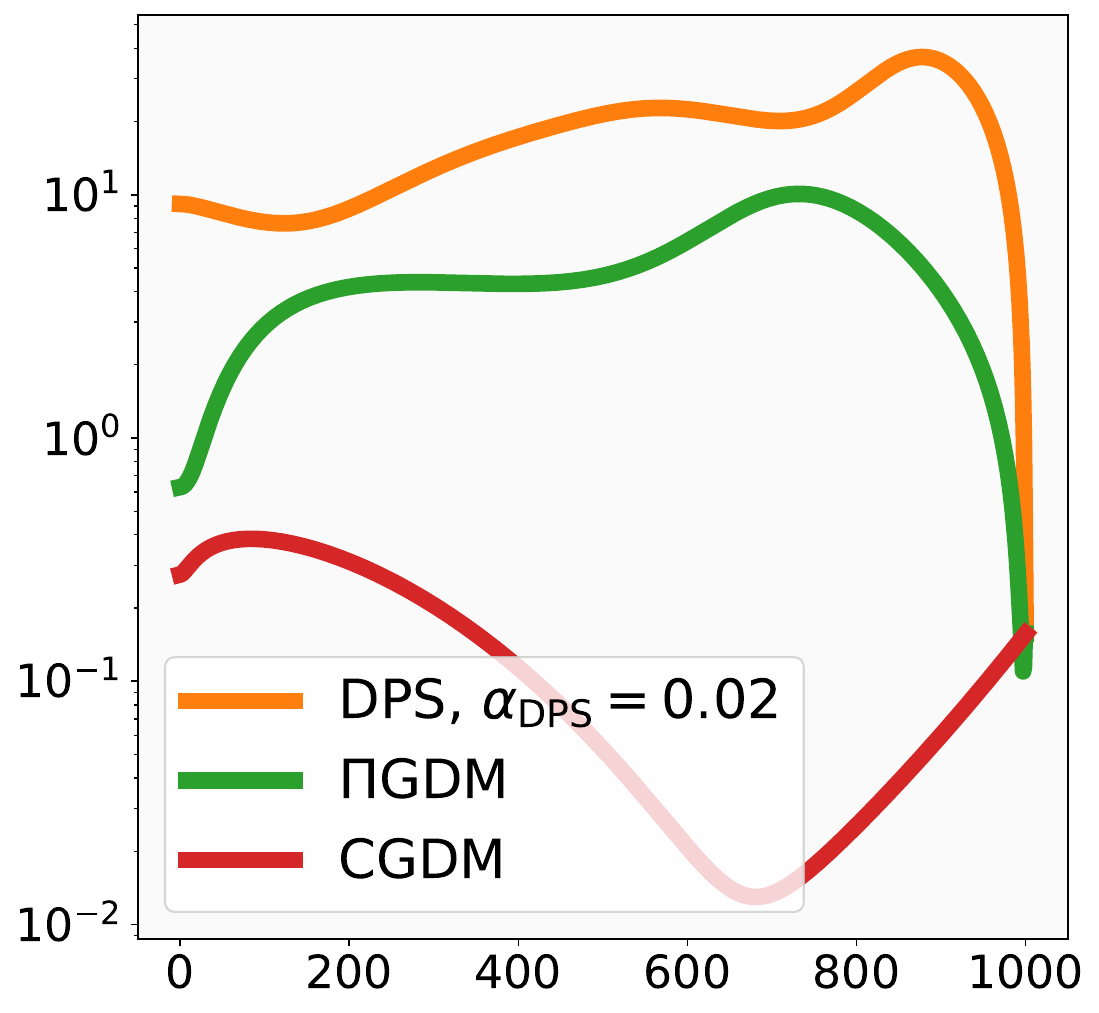} \\
	 \\
\end{tabular}
\caption[2-Wasserstein distance evolution of the different algorithms for the bicubic kernel]{\label{fig:W2_SR16} \textbf{2-Wasserstein distance evolution of the different algorithms for the bicubic kernel.} From left to right: blur kernel, image $\U$ associated with the ADSN distribution, log modulus of the DFT of the texton $\bt$, blurred image $\V$, 2-Wasserstein distance of the different algorithms with respect to the forward process, along the time. We observe a consistent ranking of the algorithms in terms of performance—DPS, $\Pi$GDM, and CGDM—from lowest to highest, across all kernels and throughout the diffusion process.}
\end{figure*}

\begin{figure*}[t]
\centering
\begin{tabular}{*{5}{>{\centering\arraybackslash}p{\obswidth}}}
Blur kernel
&
Image $\U$
&
DFT of the texton $\bt$ 
&
Blurred image $\V$
&
$\WassorthokerS^\mathrm{algo}$
\\
	\includegraphics[width=\obswidth]{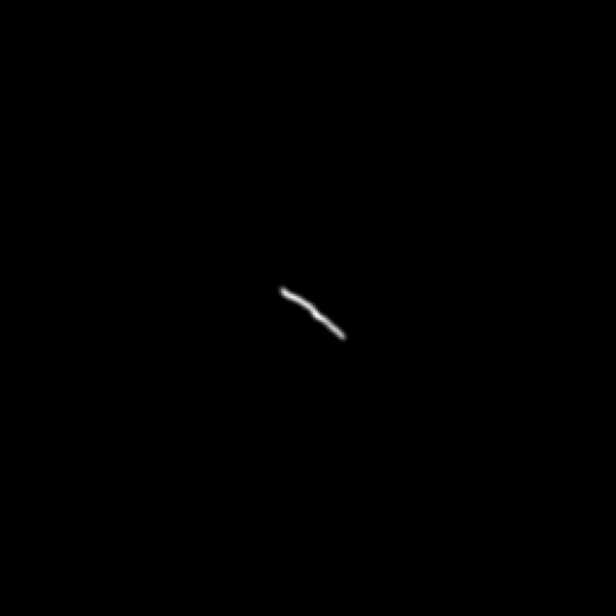}
	&
	\includegraphics[width=\obswidth]{Figures_conditional_Gaussian/ADSN_blur/Textures/Fabric.png}
	&
	\includegraphics[width=\obswidth]{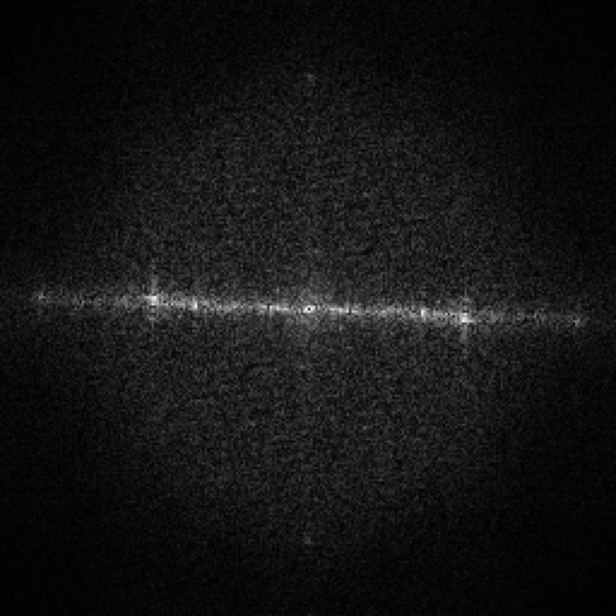}
	&
	\includegraphics[width=\obswidth]{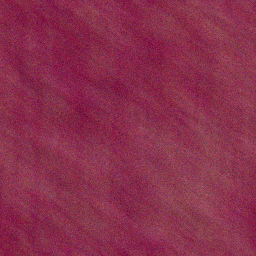}
	&
	\includegraphics[width=\obswidth]{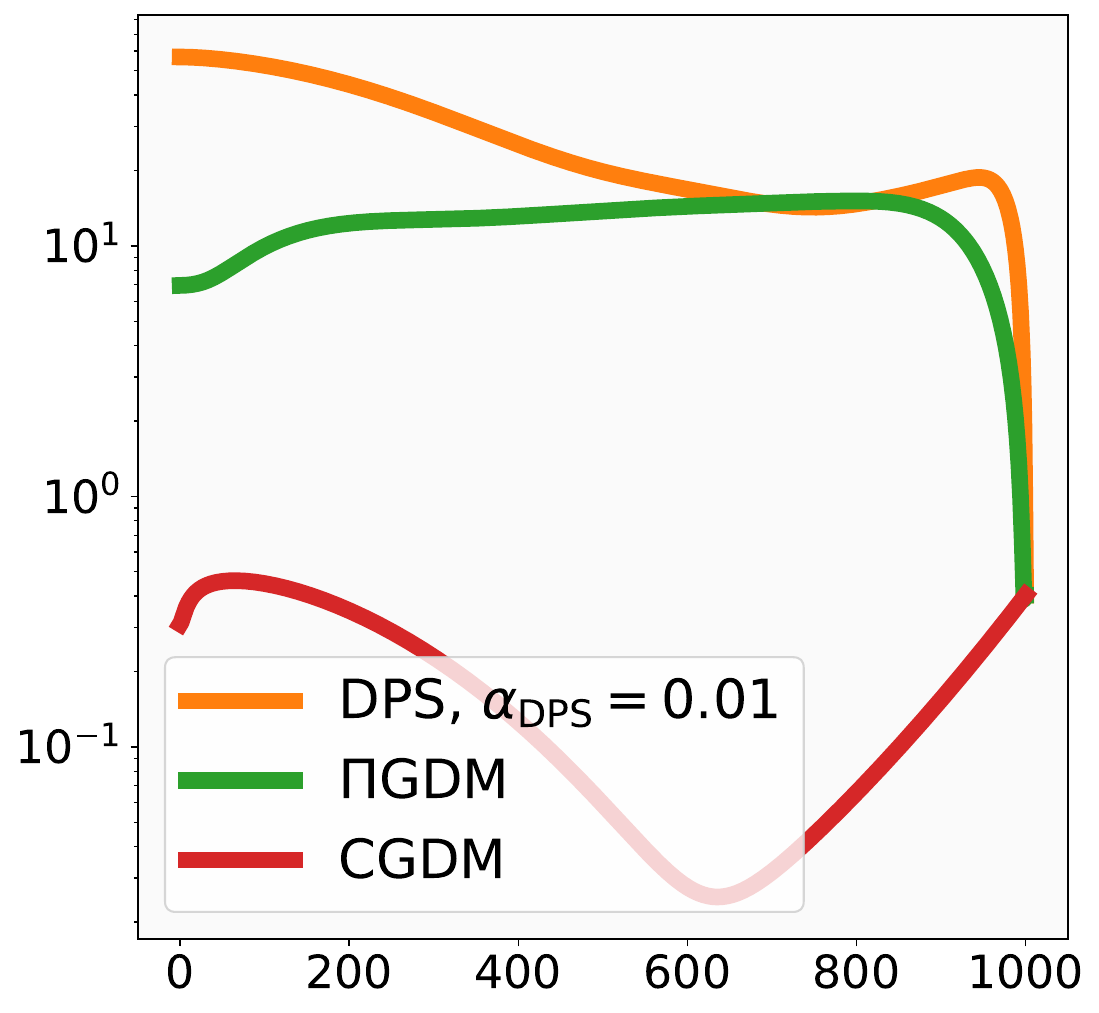} \\
	&
	\includegraphics[width=\obswidth]{Figures_conditional_Gaussian/ADSN_blur/Textures/Fabric3.png}
	&
	\includegraphics[width=\obswidth]{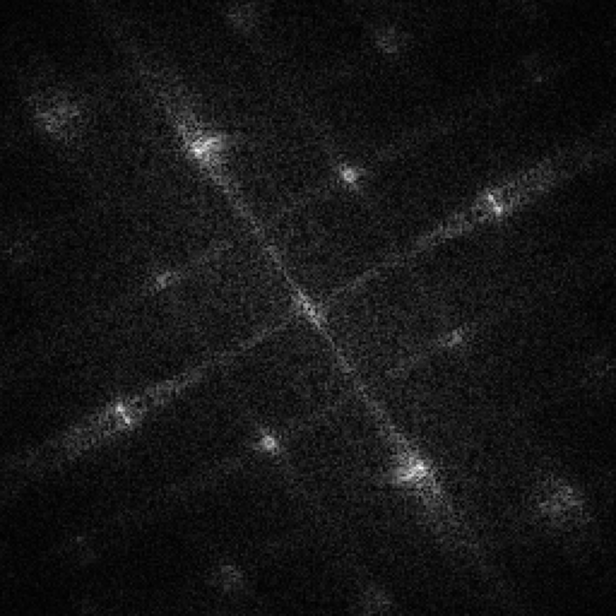}
	&
	\includegraphics[width=\obswidth]{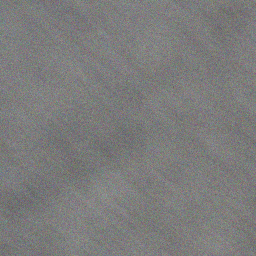}
	&
	\includegraphics[width=\obswidth]{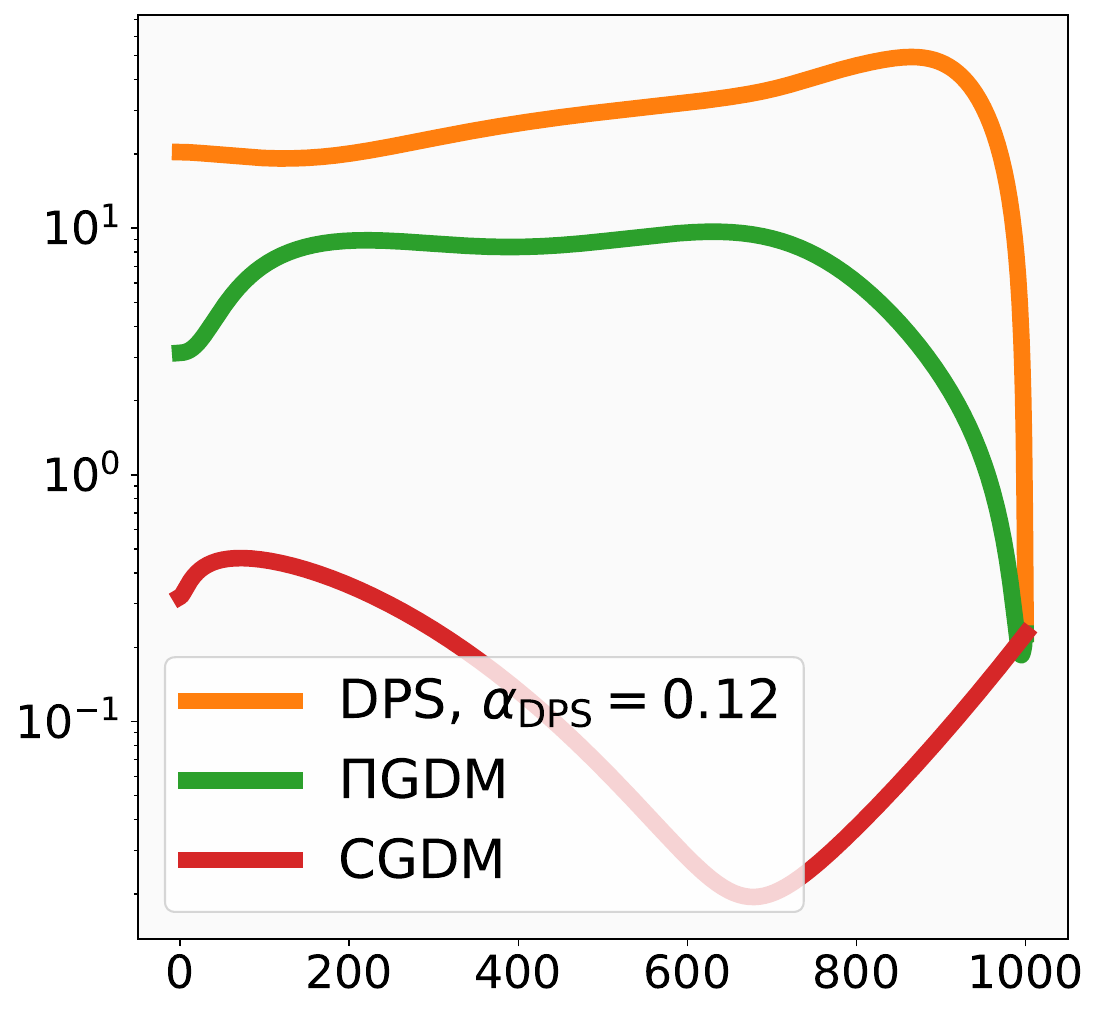} \\
	&
	\includegraphics[width=\obswidth]{Figures_conditional_Gaussian/ADSN_blur/Textures/Jean2.png}
	&
	\includegraphics[width=\obswidth]{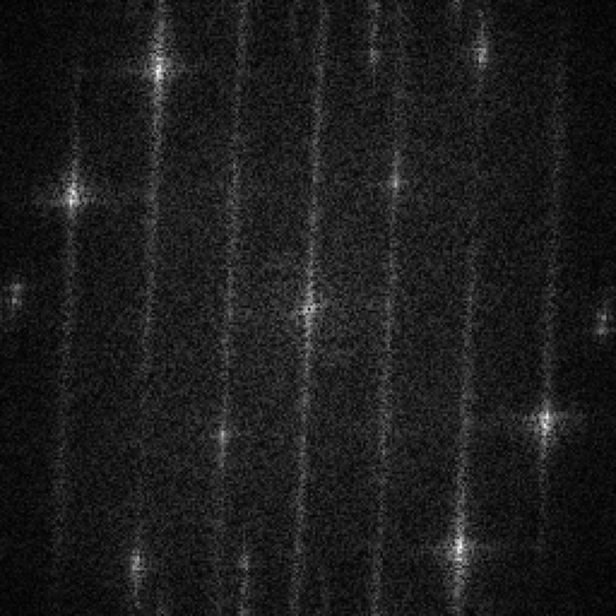}
	&
	\includegraphics[width=\obswidth]{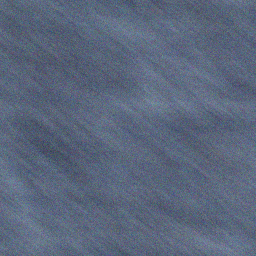}
	&
	\includegraphics[width=\obswidth]{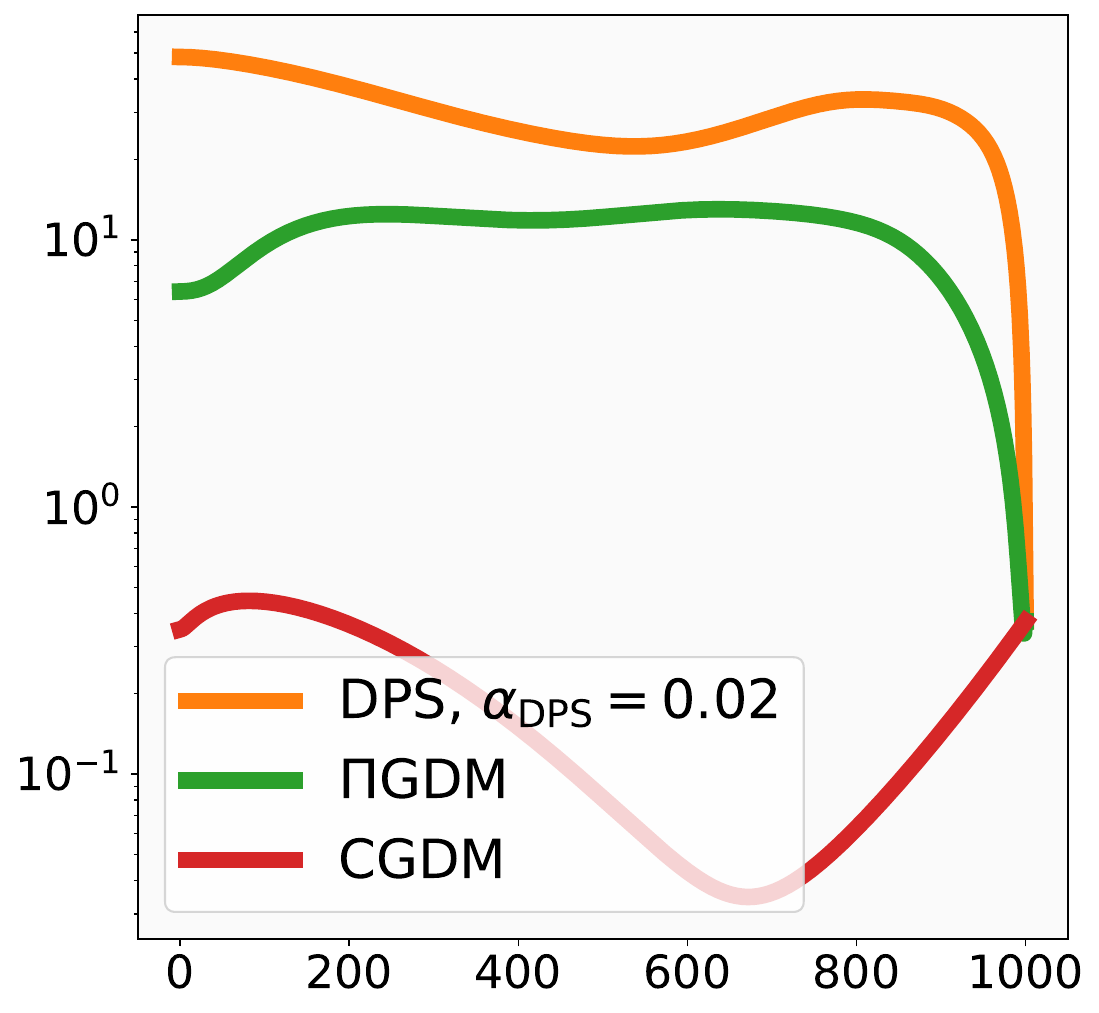} \\
	&
	\includegraphics[width=\obswidth]{Figures_conditional_Gaussian/ADSN_blur/Textures/Wall2.png}
	&
	\includegraphics[width=\obswidth]{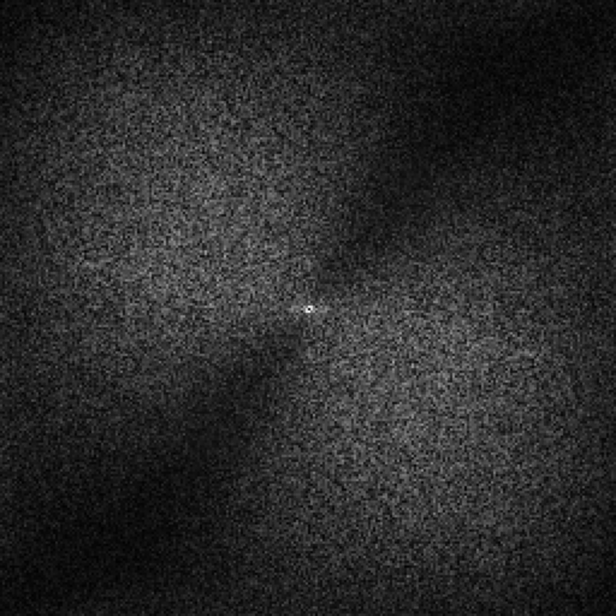}
	&
	\includegraphics[width=\obswidth]{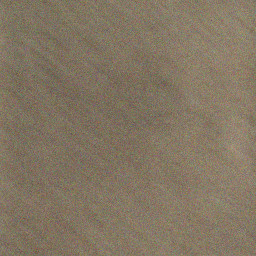}
	&
	\includegraphics[width=\obswidth]{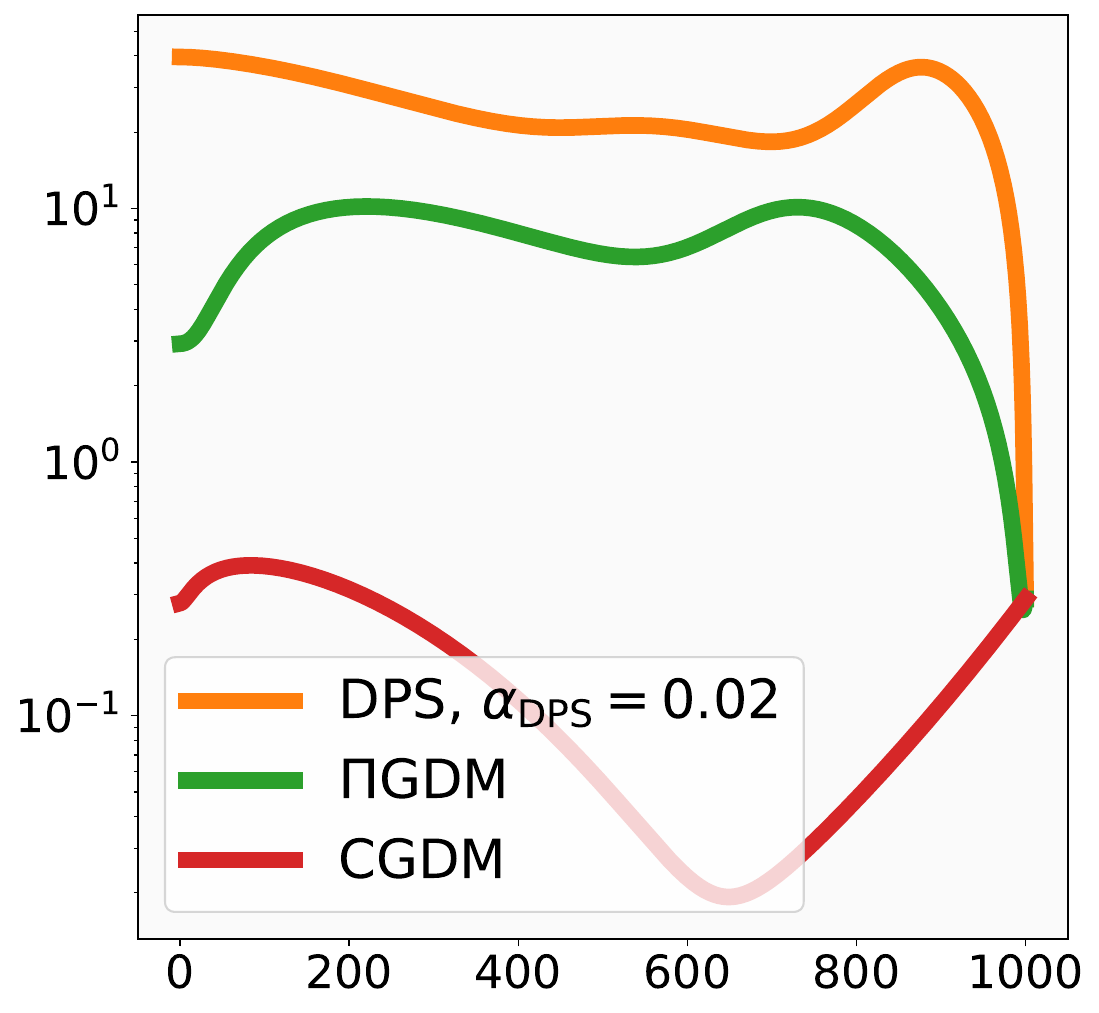} \\
\end{tabular}
\caption[2-Wasserstein distance evolution of the different algorithms for the first motion blur kernel]{\label{fig:W2_blur1} \textbf{2-Wasserstein distance evolution of the different algorithms for the first motion blur kernel.} From left to right: blur kernel, image $\U$ associated with the ADSN distribution, log modulus of the DFT of the texton $\bt$, blurred image $\V$, 2-Wasserstein distance of the different algorithms with respect to the forward process, along the time. We observe a consistent ranking of the algorithms in terms of performance—DPS, $\Pi$GDM, and CGDM—from lowest to highest, across all kernels and throughout the diffusion process, except for the first example around intermediate time steps.}
\end{figure*}

\begin{figure*}[t]
\centering
\begin{tabular}{*{5}{>{\centering\arraybackslash}p{\obswidth}}}
Blur kernel
&
Image $\U$
&
DFT of the texton $\bt$ 
&
Blurred image $\V$
&
$\WassorthokerS^\mathrm{algo}$
\\
	\includegraphics[width=\obswidth]{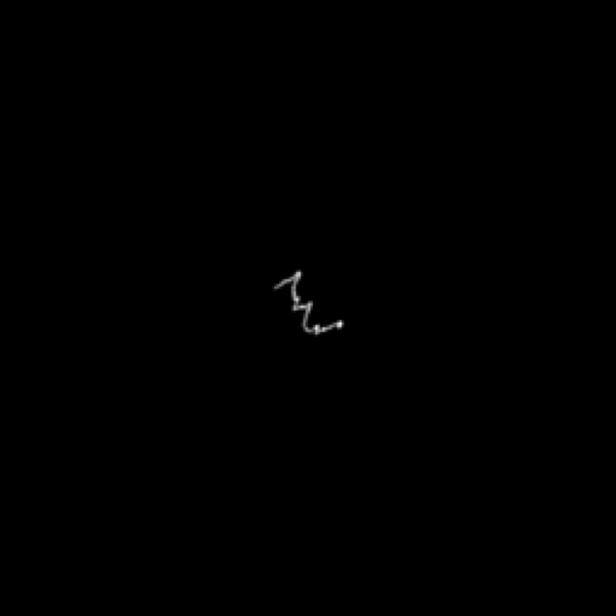}
	&
	\includegraphics[width=\obswidth]{Figures_conditional_Gaussian/ADSN_blur/Textures/Fabric.png}
	&
	\includegraphics[width=\obswidth]{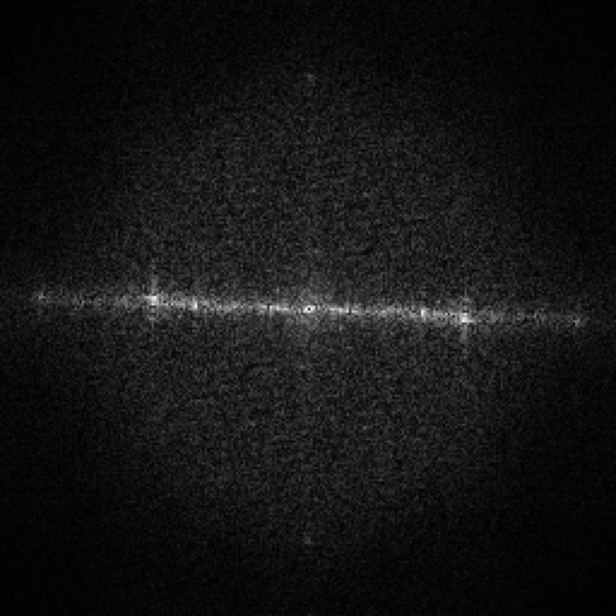}
	&
	\includegraphics[width=\obswidth]{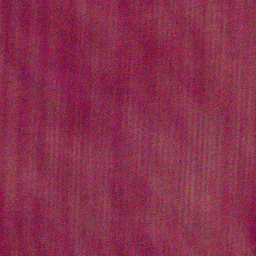}
	&
	\includegraphics[width=\obswidth]{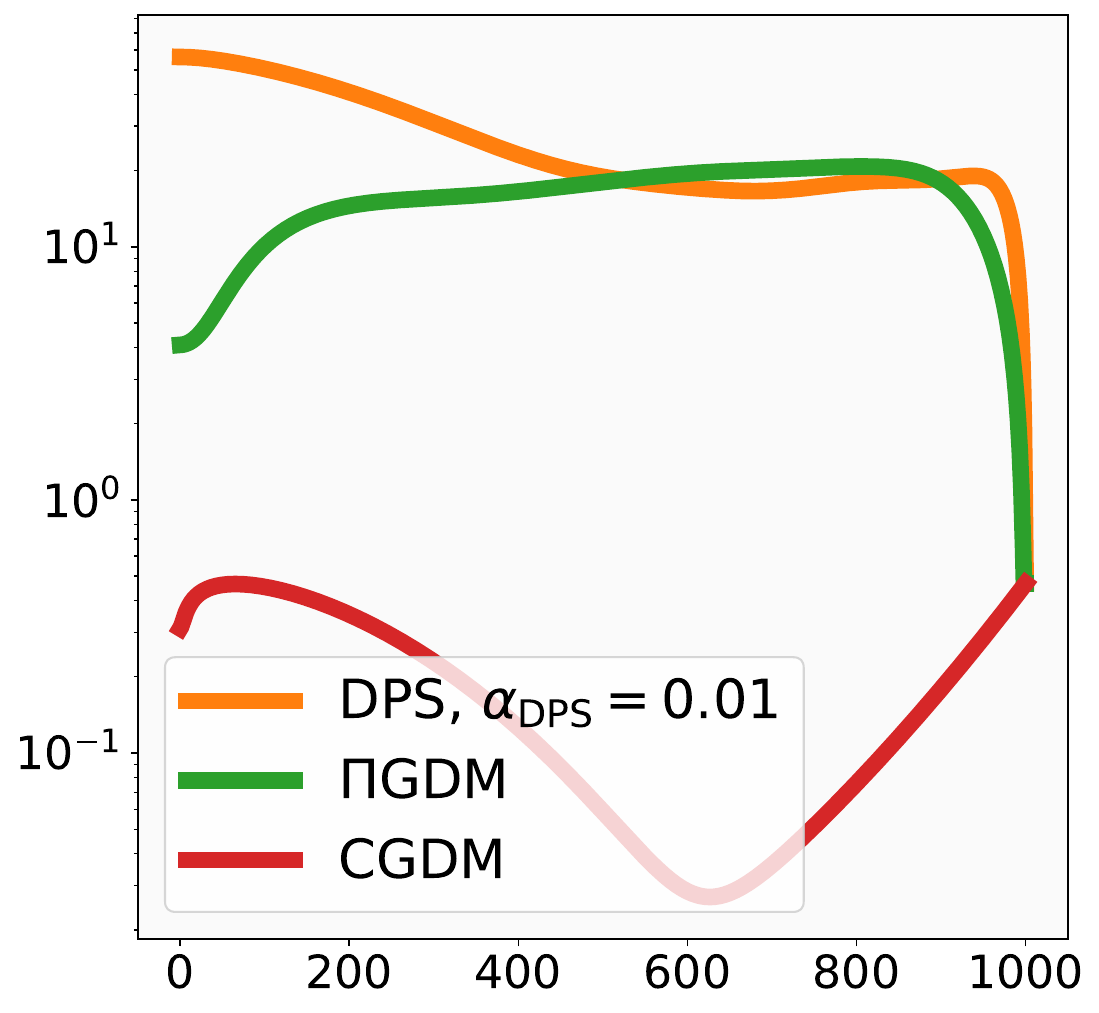} \\
	&
	\includegraphics[width=\obswidth]{Figures_conditional_Gaussian/ADSN_blur/Textures/Fabric3.png}
	&
	\includegraphics[width=\obswidth]{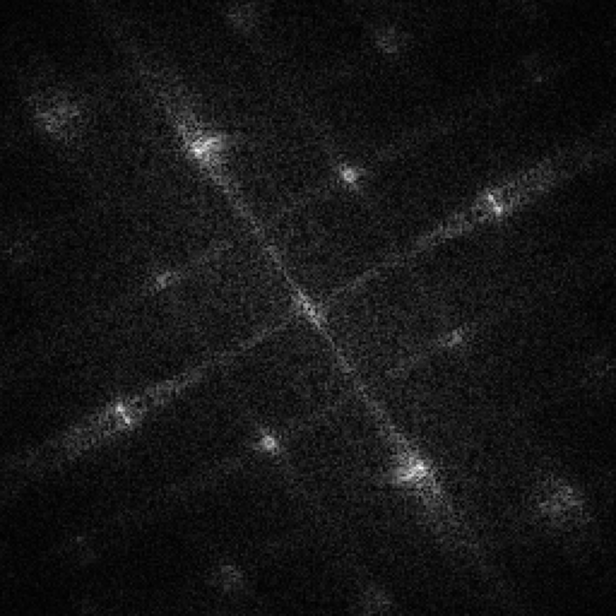}
	&
	\includegraphics[width=\obswidth]{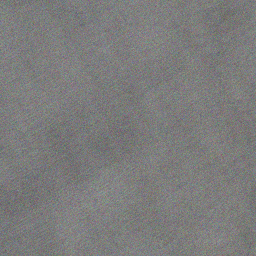}
	&
	\includegraphics[width=\obswidth]{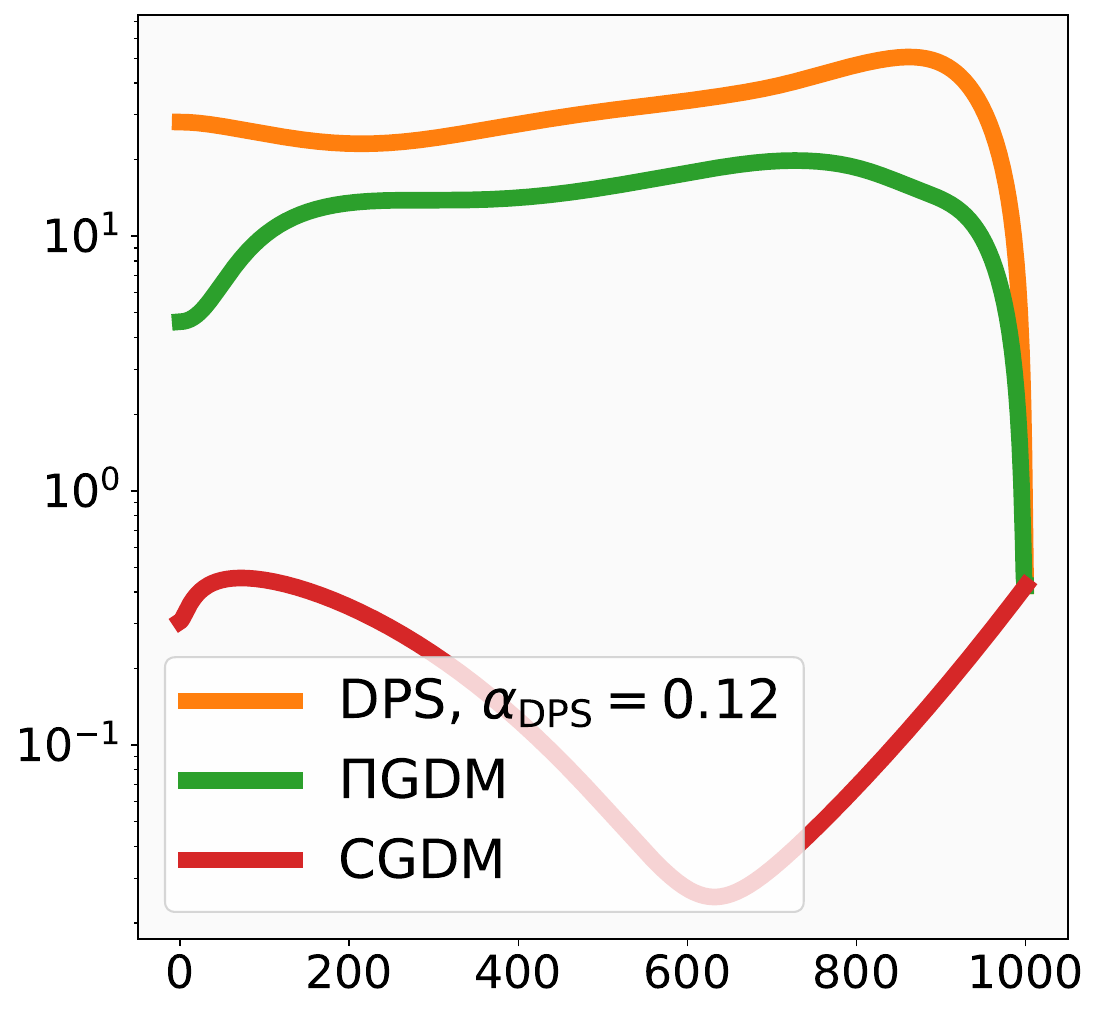} \\
	&
	\includegraphics[width=\obswidth]{Figures_conditional_Gaussian/ADSN_blur/Textures/Jean2.png}
	&
	\includegraphics[width=\obswidth]{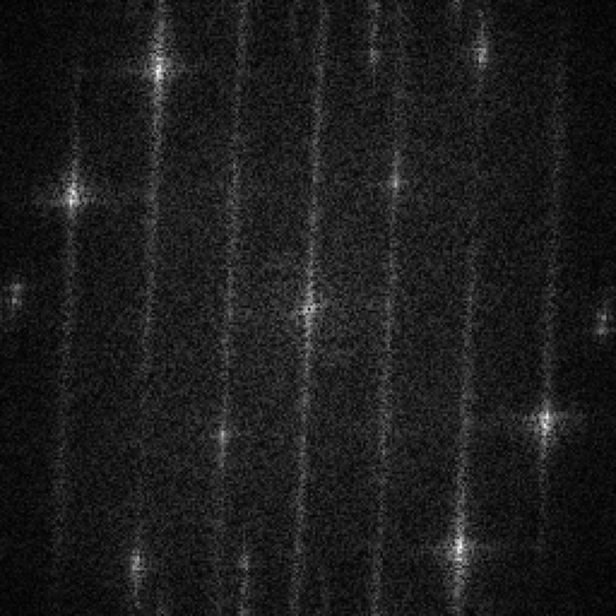}
	&
	\includegraphics[width=\obswidth]{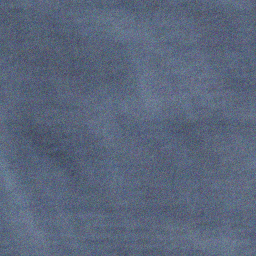}
	&
	\includegraphics[width=\obswidth]{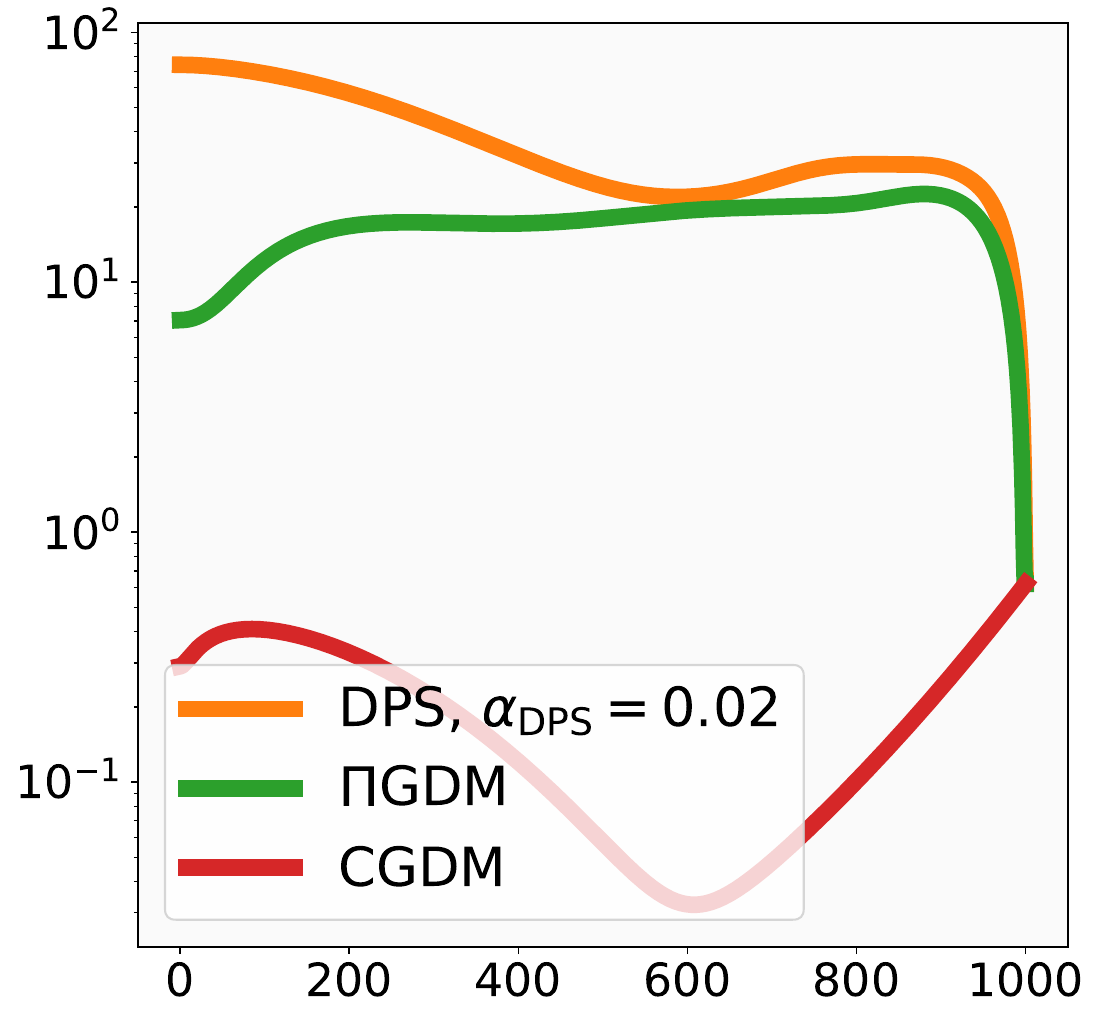} \\
	&
	\includegraphics[width=\obswidth]{Figures_conditional_Gaussian/ADSN_blur/Textures/Wall2.png}
	&
	\includegraphics[width=\obswidth]{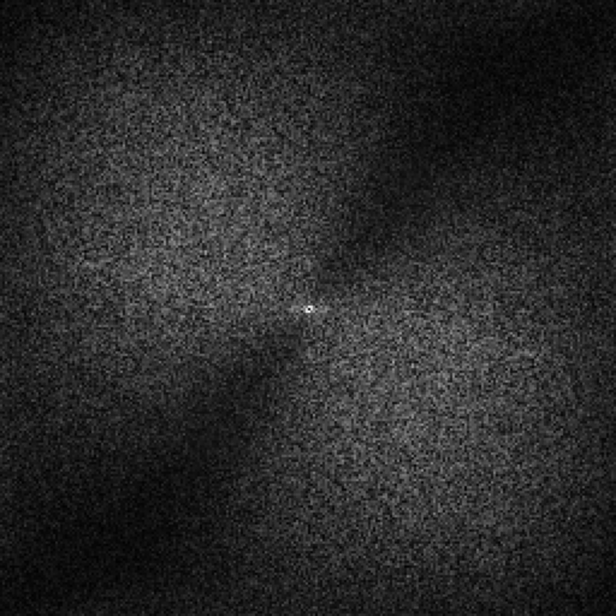}
	&
	\includegraphics[width=\obswidth]{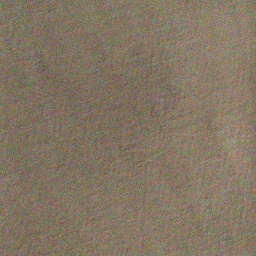}
	&
	\includegraphics[width=\obswidth]{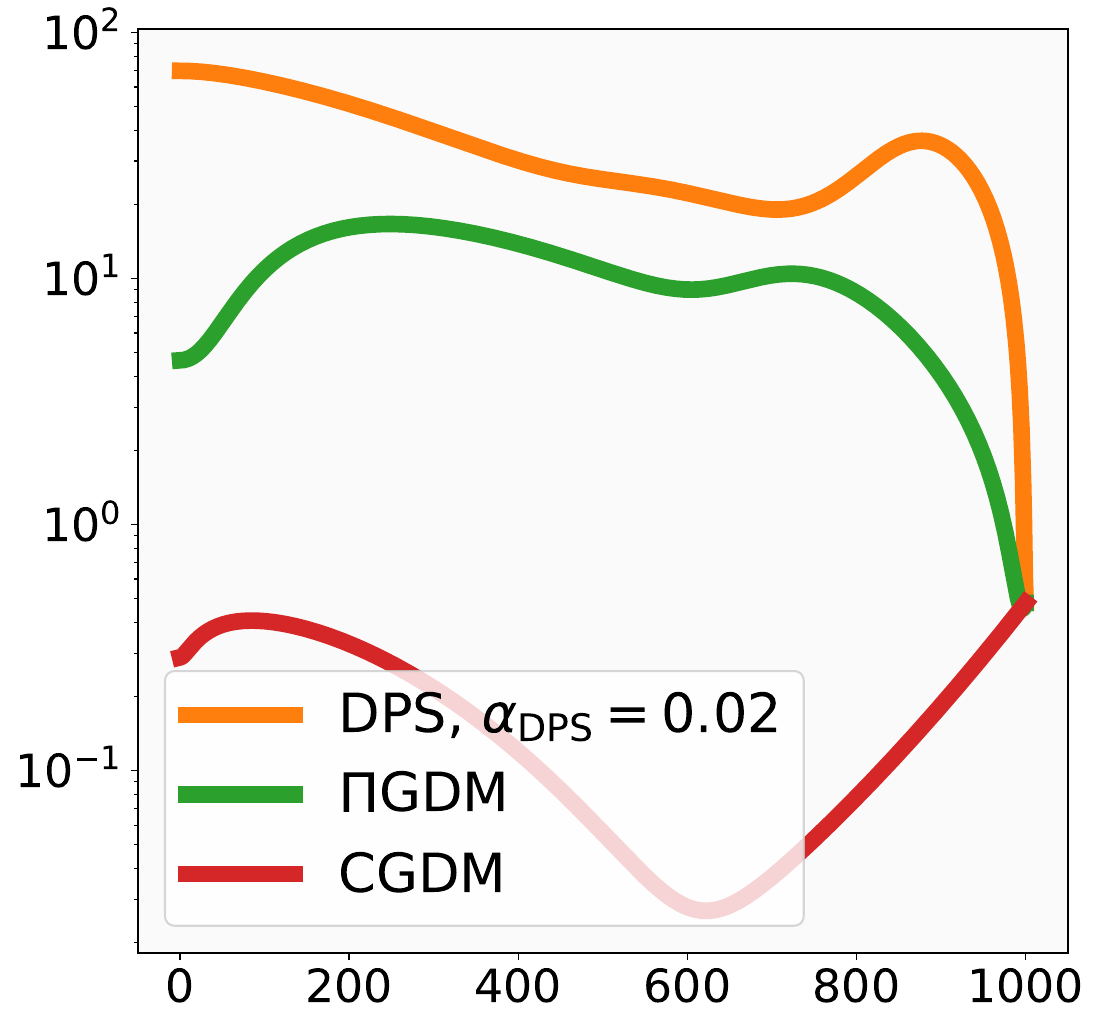} \\
	 \\
\end{tabular}
\caption[2-Wasserstein distance evolution of the different algorithms for the second motion blur kernel]{\label{fig:W2_blur6} \textbf{2-Wasserstein distance evolution of the different algorithms for the second motion blur kernel.} From left to right: blur kernel, image $\U$ associated with the ADSN distribution, log modulus of the DFT of the texton $\bt$, blurred image $\V$, 2-Wasserstein distance of the different algorithms with respect to the forward process, along the time. We observe a consistent ranking of the algorithms in terms of performance—DPS, $\Pi$GDM, and CGDM—from lowest to highest, across all kernels and throughout the diffusion process, except for the first example around intermediate time steps.}
\end{figure*}

\newlength{\kernelwidth}
\setlength{\kernelwidth}{0.5\linewidth}

\setlength{\tabcolsep}{1pt}
\begin{figure*}[t]
\centering
	\begin{tabular}{llll}
	\multicolumn{1}{c}{Image $\U$}
	& \multicolumn{1}{c}{Texton $\bt$}
	& \multicolumn{1}{c}{Blur kernel}
	& \multicolumn{1}{c}{$\V$} \\
	\includegraphics[width=\kernelwidth]{Figures_conditional_Gaussian/ADSN_blur/Textures/Fabric.png}
	& \includegraphics[width=\kernelwidth]{Figures_conditional_Gaussian/ADSN_blur/Results/Fabric/motion_blur1_sig_10/t_fft.png} 
	&
	\includegraphics[width=\kernelwidth]{Figures_conditional_Gaussian/ADSN_blur/Results/Fabric/motion_blur1_sig_10/kernel_not_fft.png}
	&
	\includegraphics[width=\kernelwidth]{Figures_conditional_Gaussian/ADSN_blur/Results/Fabric/motion_blur1_sig_10/v.png}
	\\
	\multicolumn{4}{c}{Algorithms means} \\
	\midrule
	\multicolumn{1}{c}{Conditional distribution} & 
	\multicolumn{1}{c}{CGDM} & 
	\multicolumn{1}{c}{$\Pi$GDM} & 
	\multicolumn{1}{c}{DPS} \\
	\includegraphics[width=\kernelwidth]{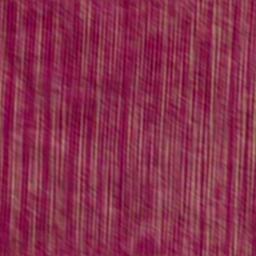}
	& \includegraphics[width=\kernelwidth]{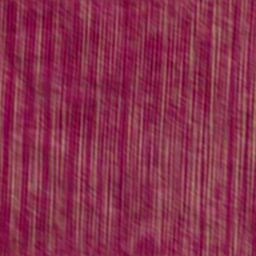} 
	&
	\includegraphics[width=\kernelwidth]{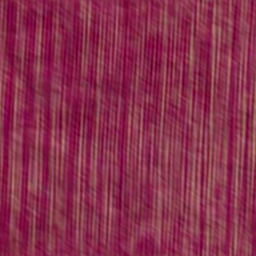} 
	&
	\includegraphics[width=\kernelwidth]{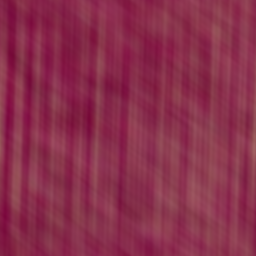}
	\\
	\multicolumn{4}{c}{Algorithms samples} \\
	\midrule
	\multicolumn{1}{c}{Conditional distribution} & 
	\multicolumn{1}{c}{CGDM} & 
	\multicolumn{1}{c}{$\Pi$GDM} & 
	\multicolumn{1}{c}{DPS} \\
	\includegraphics[width=\kernelwidth]{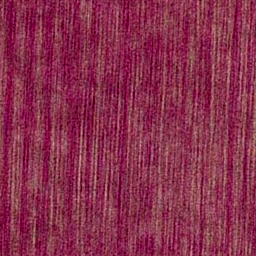}
	& \includegraphics[width=\kernelwidth]{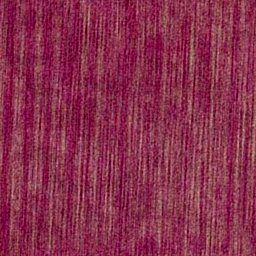} 
	&
	\includegraphics[width=\kernelwidth]{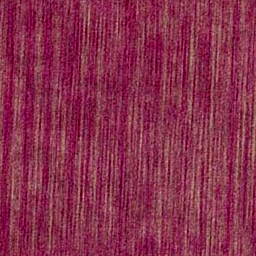}
	&
	\includegraphics[width=\kernelwidth]{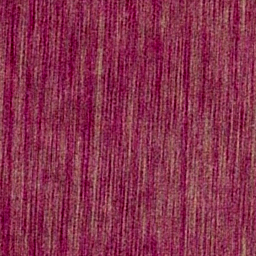}
	\\		
	\includegraphics[width=\kernelwidth]{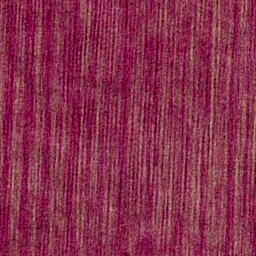}
	& \includegraphics[width=\kernelwidth]{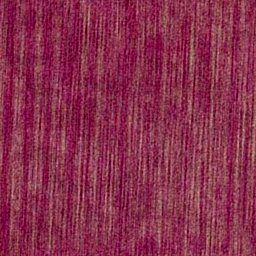} 
	&
	\includegraphics[width=\kernelwidth]{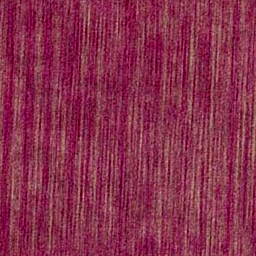}
	&
	\includegraphics[width=\kernelwidth]{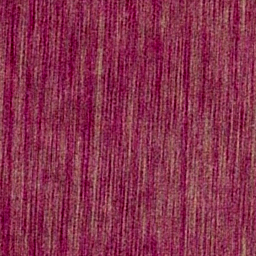}
	\\		
	\end{tabular}
	\caption[Samples and means generated by the different algorithms.]{\label{fig:samples_algorithms_fabric_motion_blur1_sig_10} \textbf{Samples and means generated by the different algorithms.} CGDM, $\Pi$GDM, DPS samples are generated from the same seed at each step and are very similar. However, the mean of the DPS algorithm contains less texture information than the other alglorithms which illustrates its bias. 
    }
\end{figure*}

\newlength{\conddirtribwidth}
\setlength{\conddirtribwidth}{0.27\linewidth}

\begin{figure*}[t]
\centering
	\begin{tabular}{c>{\centering\arraybackslash}p{\conddirtribwidth}>{\centering\arraybackslash}p{\conddirtribwidth}>{\centering\arraybackslash}p{\conddirtribwidth}>{\centering\arraybackslash}p{\conddirtribwidth}>{\centering\arraybackslash}p{\conddirtribwidth}>{\centering\arraybackslash}p{\conddirtribwidth}>{\centering\arraybackslash}p{\conddirtribwidth}}
	\toprule
	$t$
	&
	800
	&
	600
	&
	400
	&
	200
	&
	50
	&
	0 
	&
	Samples
	\\
	\midrule
	\rotatebox{90}{\parbox{\conddirtribwidth}{\centering True theoretical  \\ distribution }}
	&
	\includegraphics[width=\conddirtribwidth]{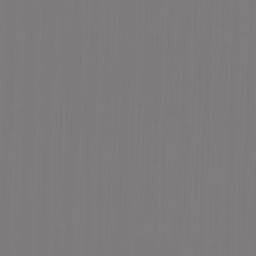}
	&
	\includegraphics[width=\conddirtribwidth]{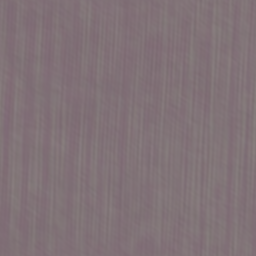}
	&
	\includegraphics[width=\conddirtribwidth]{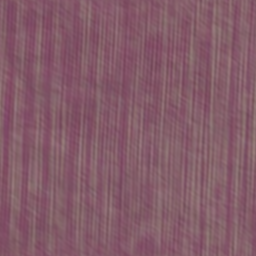}
	&
	\includegraphics[width=\conddirtribwidth]{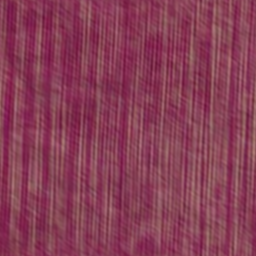}
	&
	\includegraphics[width=\conddirtribwidth]{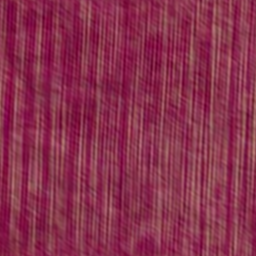}
&
	\includegraphics[width=\conddirtribwidth]{Figures_conditional_Gaussian/ADSN_blur/Results/Fabric/motion_blur1_sig_10/conditional_mean.png}
	&
	\includegraphics[width=\conddirtribwidth]{Figures_conditional_Gaussian/ADSN_blur/Results/Fabric/motion_blur1_sig_10/conditional_sample_1.png}
	\\
	\rotatebox{90}{\parbox{\conddirtribwidth}{\centering CGDM}}
	&
	\includegraphics[width=\conddirtribwidth]{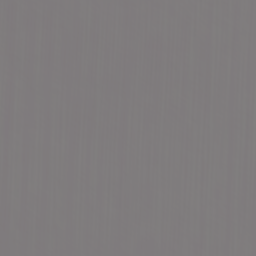}
	&
	\includegraphics[width=\conddirtribwidth]{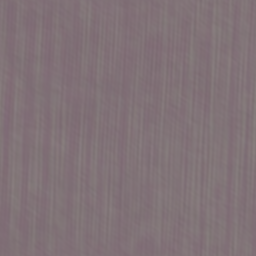}
	&
	\includegraphics[width=\conddirtribwidth]{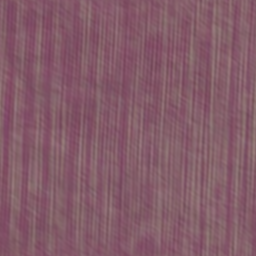}
	&
	\includegraphics[width=\conddirtribwidth]{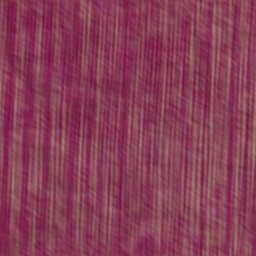}
	&
	\includegraphics[width=\conddirtribwidth]{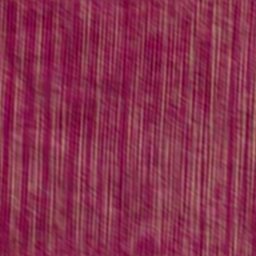}
	&
	\includegraphics[width=\conddirtribwidth]{Figures_conditional_Gaussian/ADSN_blur/Results/Fabric/motion_blur1_sig_10/mean_CGDM.png}
	&
		\includegraphics[width=\conddirtribwidth]{Figures_conditional_Gaussian/ADSN_blur/Results/Fabric/motion_blur1_sig_10/sample_1_CGDM.png}
	\\
	\rotatebox{90}{\parbox{\conddirtribwidth}{\centering $\Pi$GDM}}
	&
	\includegraphics[width=\conddirtribwidth]{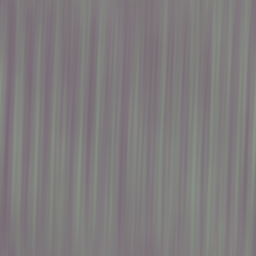}
	&
	\includegraphics[width=\conddirtribwidth]{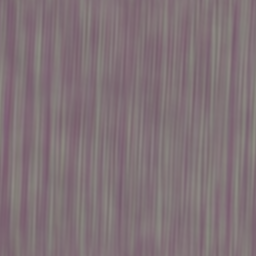}
	&
	\includegraphics[width=\conddirtribwidth]{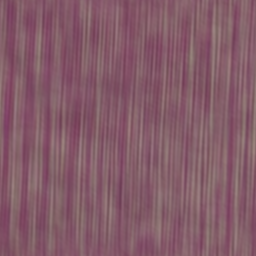}
	&
	\includegraphics[width=\conddirtribwidth]{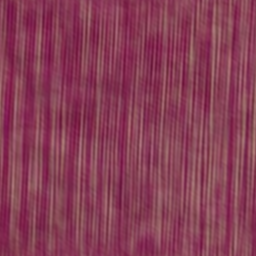}
	&
	\includegraphics[width=\conddirtribwidth]{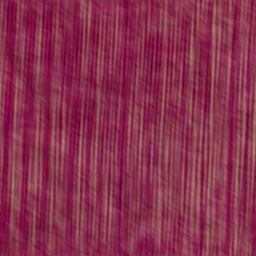}
	&
	\includegraphics[width=\conddirtribwidth]{Figures_conditional_Gaussian/ADSN_blur/Results/Fabric/motion_blur1_sig_10/mean_PiGDM.png}
		&
	\includegraphics[width=\conddirtribwidth]{Figures_conditional_Gaussian/ADSN_blur/Results/Fabric/motion_blur1_sig_10/sample_1_PiGDM.png}
	\\
	\rotatebox{90}{\parbox{\conddirtribwidth}{\centering DPS}}
	&
	\includegraphics[width=\conddirtribwidth]{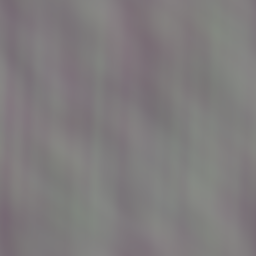}
	&
	\includegraphics[width=\conddirtribwidth]{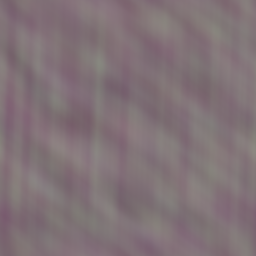}
	&
	\includegraphics[width=\conddirtribwidth]{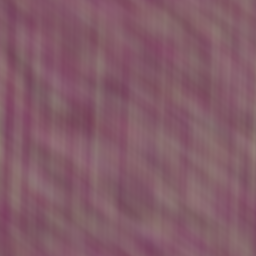}
	&
	\includegraphics[width=\conddirtribwidth]{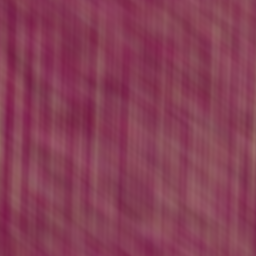}
	&
	\includegraphics[width=\conddirtribwidth]{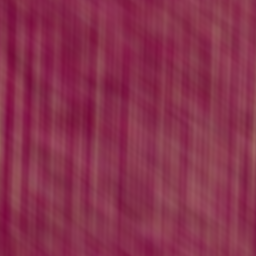}
	&
	\includegraphics[width=\conddirtribwidth]{Figures_conditional_Gaussian/ADSN_blur/Results/Fabric/motion_blur1_sig_10/mean_DPS.png} 
		&
	\includegraphics[width=\conddirtribwidth]{Figures_conditional_Gaussian/ADSN_blur/Results/Fabric/motion_blur1_sig_10/sample_1_DPS.png} \\
	\bottomrule 
	\end{tabular}
	\caption[Means of the algorithms along the time.]{\label{fig:means_Fabric_SR16} \textbf{Means of the algorithms along the time.} It corresponds to the first motion blur kernel, for the fisrt fabric texture in \Cref{fig:W2_SR16}. Note that the DPS algorithms suffers from a relative important bias along times, as observed for Gaussian distributions in small dimension.}
\end{figure*}

\begin{figure*}[t]
\centering
	\begin{tabular}{llll}
	\multicolumn{1}{c}{Image $\U$}
	& \multicolumn{1}{c}{Texton $\bt$}
	& \multicolumn{1}{c}{Blur kernel}
	& \multicolumn{1}{c}{$\V$} \\
	\includegraphics[width=\kernelwidth]{Figures_conditional_Gaussian/ADSN_blur/Textures/Fabric3.png}
	& \includegraphics[width=\kernelwidth]{Figures_conditional_Gaussian/ADSN_blur/Results/Fabric3/motion_blur1_sig_10/t_fft.png} 
	&
	\includegraphics[width=\kernelwidth]{Figures_conditional_Gaussian/ADSN_blur/Results/Fabric/motion_blur1_sig_10/kernel_not_fft.png}
	&
	\includegraphics[width=\kernelwidth]{Figures_conditional_Gaussian/ADSN_blur/Results/Fabric3/motion_blur1_sig_10/v.png}
	\\
	\multicolumn{4}{c}{Algorithms means} \\
	\midrule
	\multicolumn{1}{c}{Conditional distribution} & 
	\multicolumn{1}{c}{CGDM} & 
	\multicolumn{1}{c}{$\Pi$GDM} & 
	\multicolumn{1}{c}{DPS} \\
	\includegraphics[width=\kernelwidth]{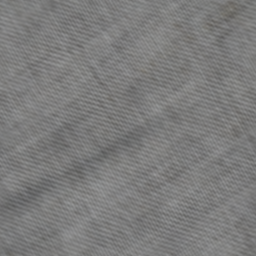}
	& \includegraphics[width=\kernelwidth]{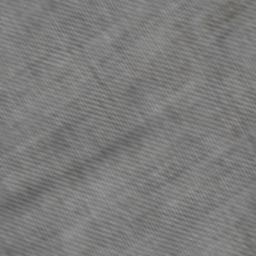} 
	&
	\includegraphics[width=\kernelwidth]{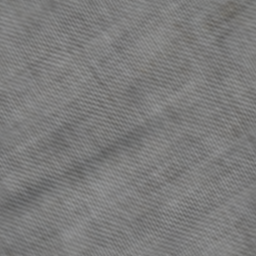} 
	&
	\includegraphics[width=\kernelwidth]{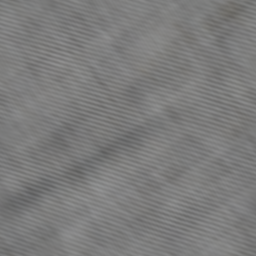}
	\\
	\multicolumn{4}{c}{Algorithms samples} \\
	\midrule
	\multicolumn{1}{c}{Conditional distribution} & 
	\multicolumn{1}{c}{CGDM} & 
	\multicolumn{1}{c}{$\Pi$GDM} & 
	\multicolumn{1}{c}{DPS} \\
	\includegraphics[width=\kernelwidth]{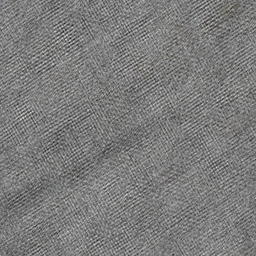}
	& \includegraphics[width=\kernelwidth]{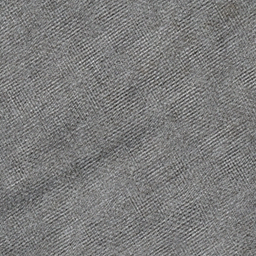} 
	&
	\includegraphics[width=\kernelwidth]{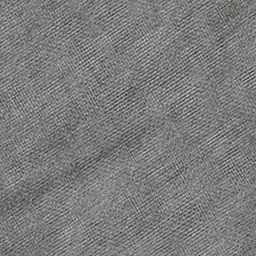}
	&
	\includegraphics[width=\kernelwidth]{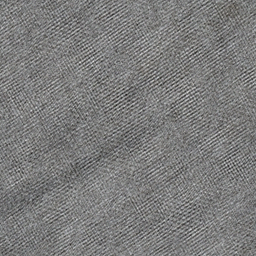}
	\\		
		\includegraphics[width=\kernelwidth]{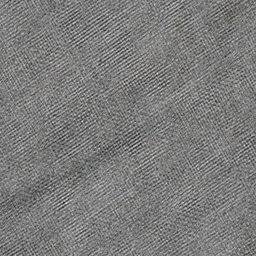}
	& \includegraphics[width=\kernelwidth]{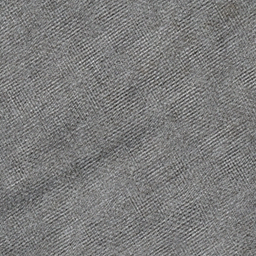} 
	&
	\includegraphics[width=\kernelwidth]{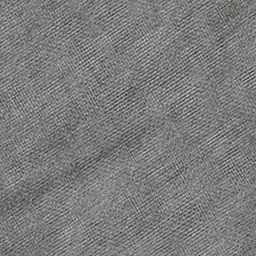}
	&
	\includegraphics[width=\kernelwidth]{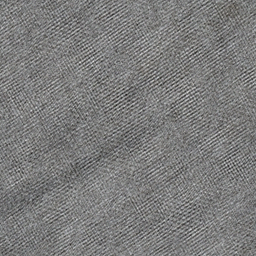}
	\\		
	\end{tabular}
	\caption[Samples and means generated by the different algorithms.]{\label{fig:samples_algorithms_Fabric3_motion_blur1_sig_10} \textbf{Samples and means generated by the different algorithms.} CGDM, $\Pi$GDM, DPS samples are generated from the same noise at each step and are very similar. However, the means are perceptually really different.}
\end{figure*}

\newlength{\cbarheigth}
\setlength{\conddirtribwidth}{0.26\linewidth}
\setlength{\cbarheigth}{\conddirtribwidth}

\begin{figure*}[t]
\centering
	\begin{tabular}{c>{\centering\arraybackslash}p{\conddirtribwidth}c>{\centering\arraybackslash}p{\conddirtribwidth}c>{\centering\arraybackslash}p{\conddirtribwidth}c>{\centering\arraybackslash}p{\conddirtribwidth}c>{\centering\arraybackslash}p{\conddirtribwidth}c>{\centering\arraybackslash}p{\conddirtribwidth}c}
	\toprule
	$t$
	&
	800
	&
	&
	600
	&
	&
	400
	&
	&
	200
	&
	&
	50
	&
	&
	0
	& \\
	\midrule
	\rotatebox{90}{\parbox{\conddirtribwidth}{\centering CGDM }}
	&
	\includegraphics[width=\conddirtribwidth]{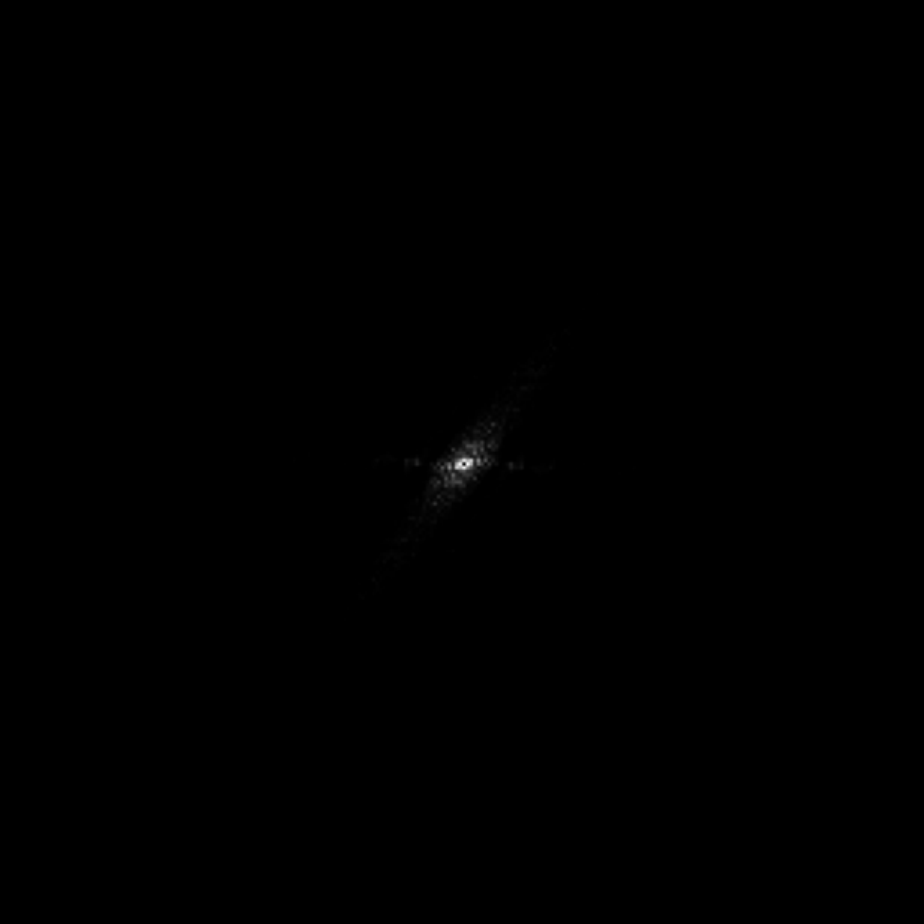}
	&
	\includegraphics[height=\cbarheigth]{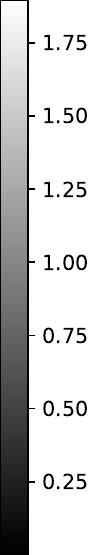}	
	&
	\includegraphics[width=\conddirtribwidth]{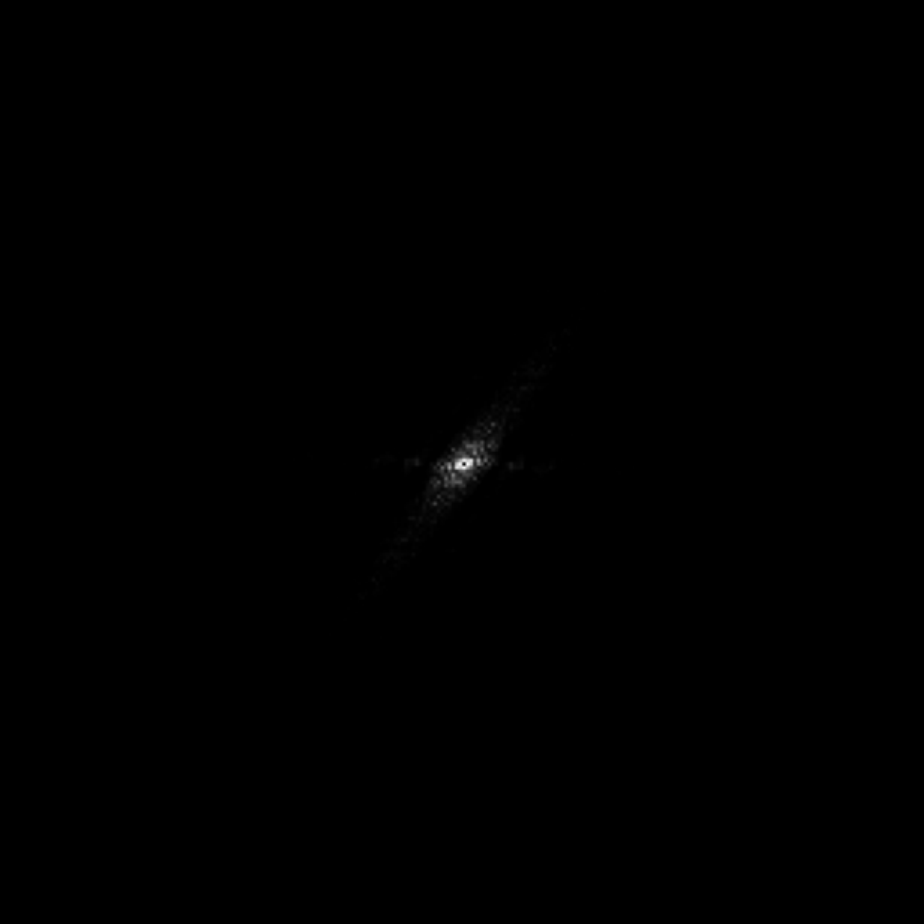}
	&
	\includegraphics[height=\cbarheigth]{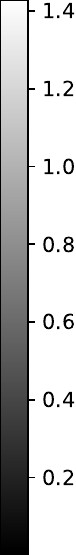}	
	&
	\includegraphics[width=\conddirtribwidth]{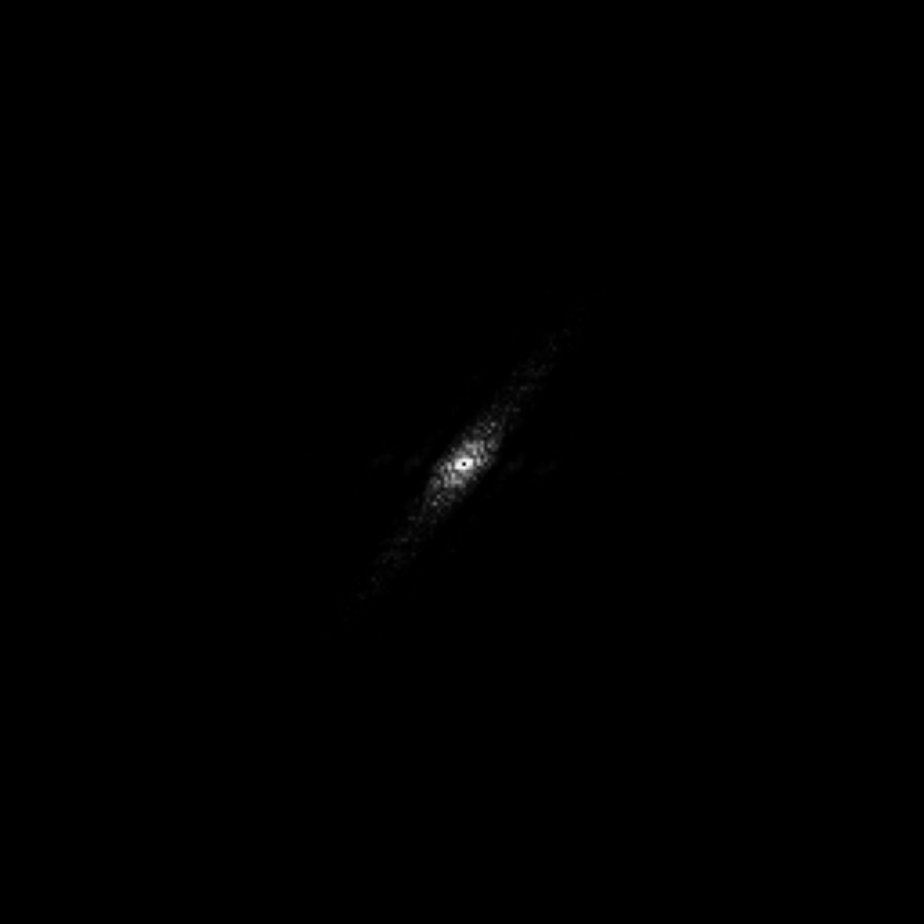}
	&
	\includegraphics[height=\cbarheigth]{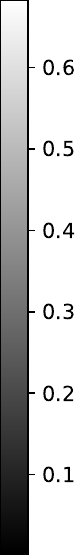}	
	&
	\includegraphics[width=\conddirtribwidth]{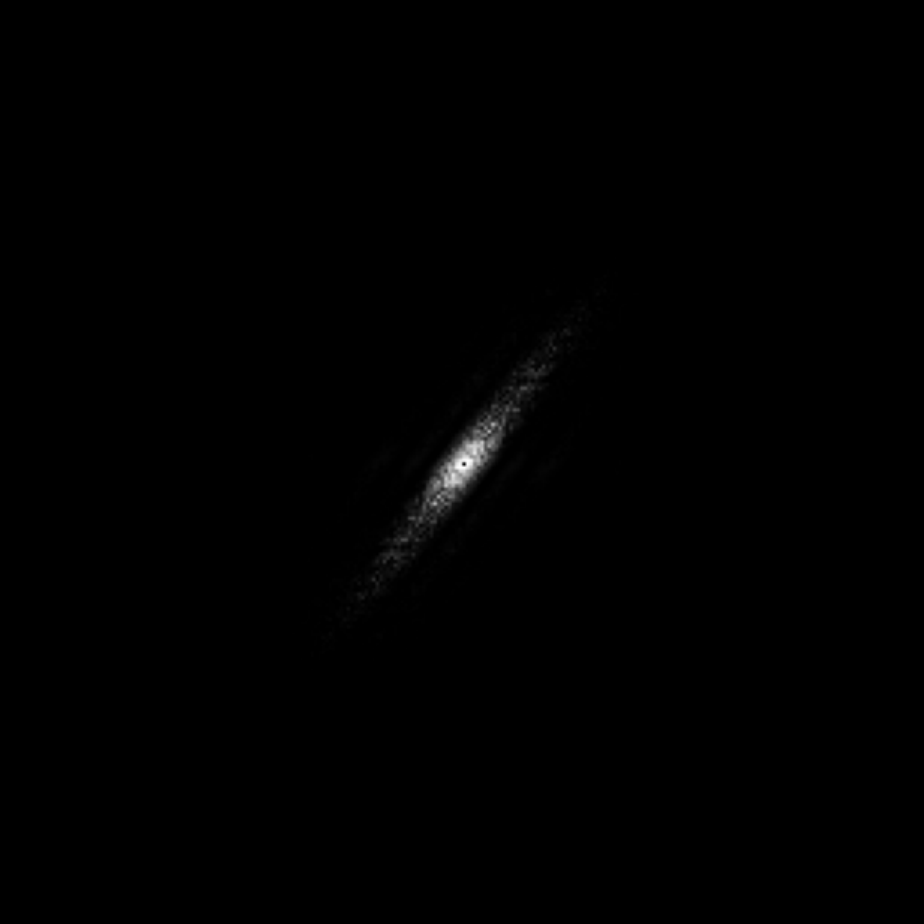}
	&
	\includegraphics[height=\cbarheigth]{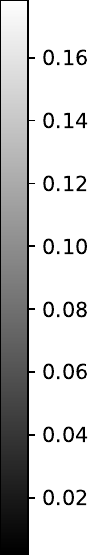}	
	&
	\includegraphics[width=\conddirtribwidth]{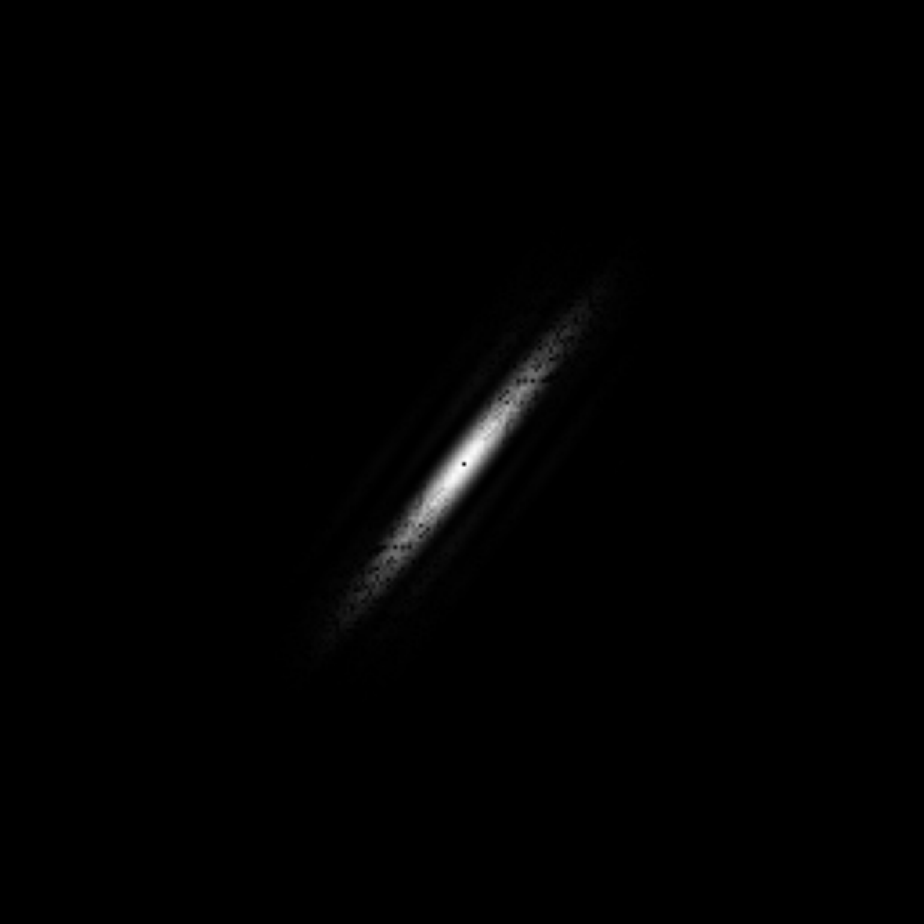}
	&
	\includegraphics[height=\cbarheigth]{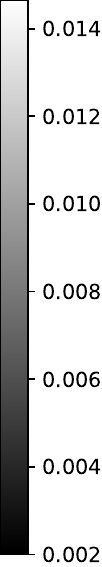}	
	&
	\includegraphics[width=\conddirtribwidth]{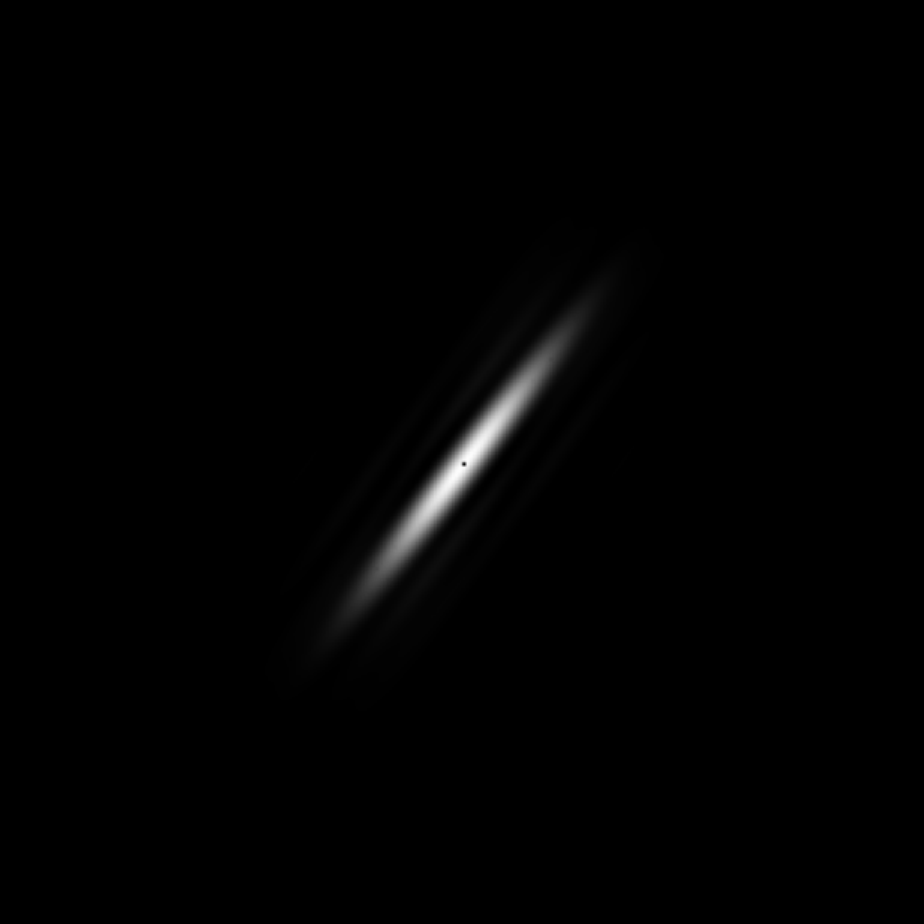}
	&
	\includegraphics[height=\cbarheigth]{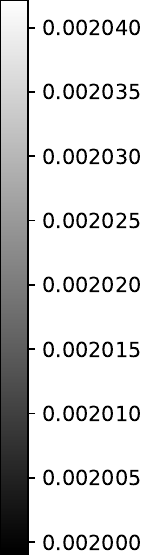}	 \\
	\rotatebox{90}{\parbox{\conddirtribwidth}{\centering $\Pi$GDM }}
	&
	\includegraphics[width=\conddirtribwidth]{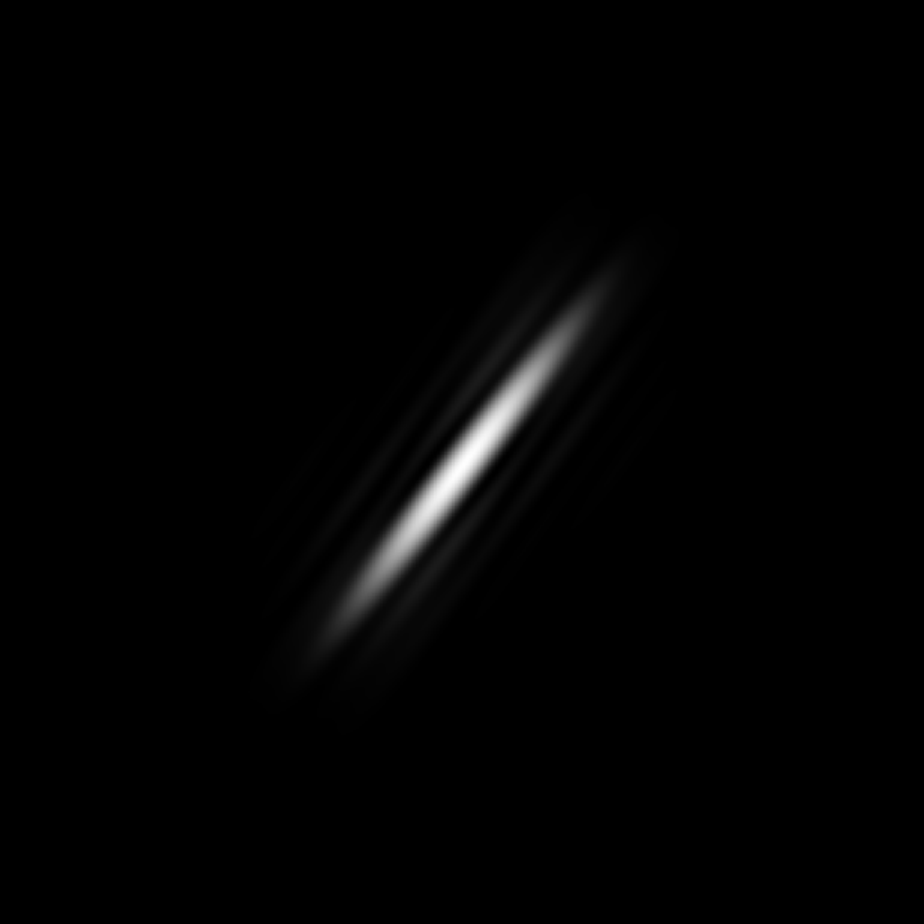}
	&
	\includegraphics[height=\cbarheigth]{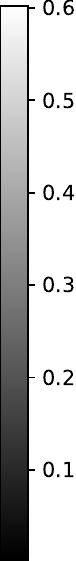}	
	&
	\includegraphics[width=\conddirtribwidth]{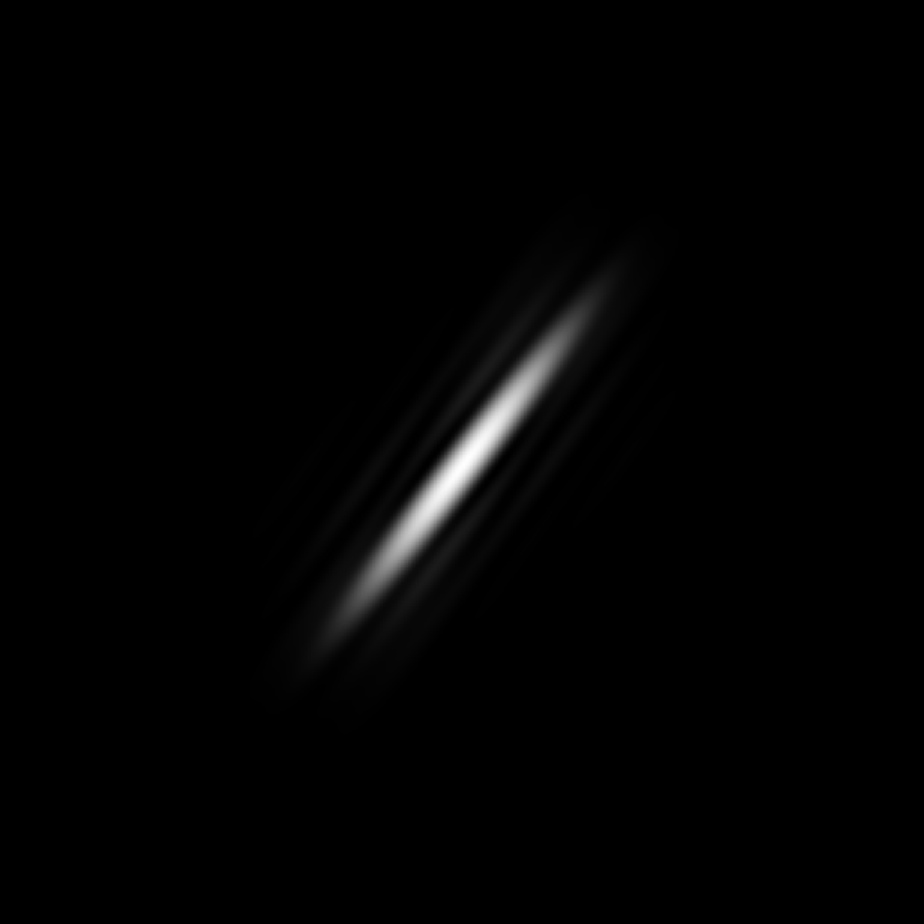}
	&
	\includegraphics[height=\cbarheigth]{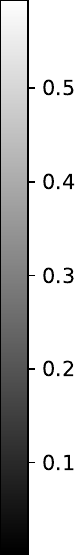}	
	&
	\includegraphics[width=\conddirtribwidth]{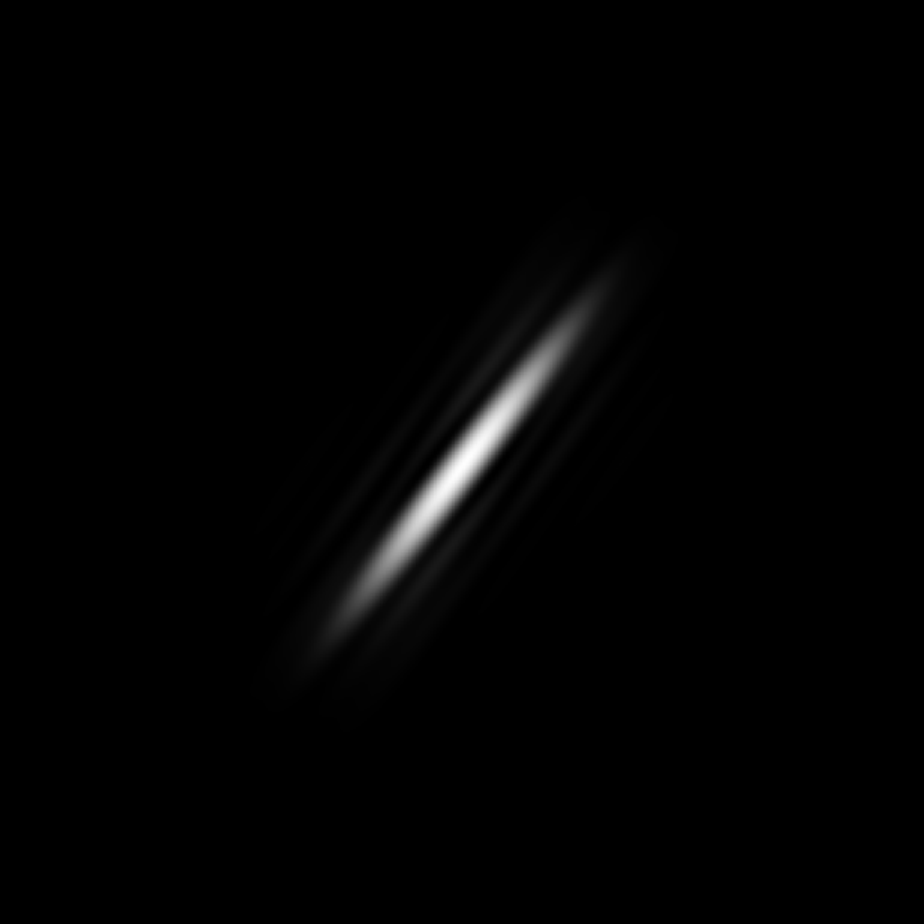}
	&
	\includegraphics[height=\cbarheigth]{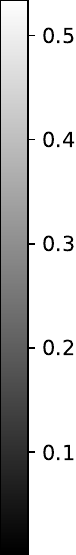}	
	&
	\includegraphics[width=\conddirtribwidth]{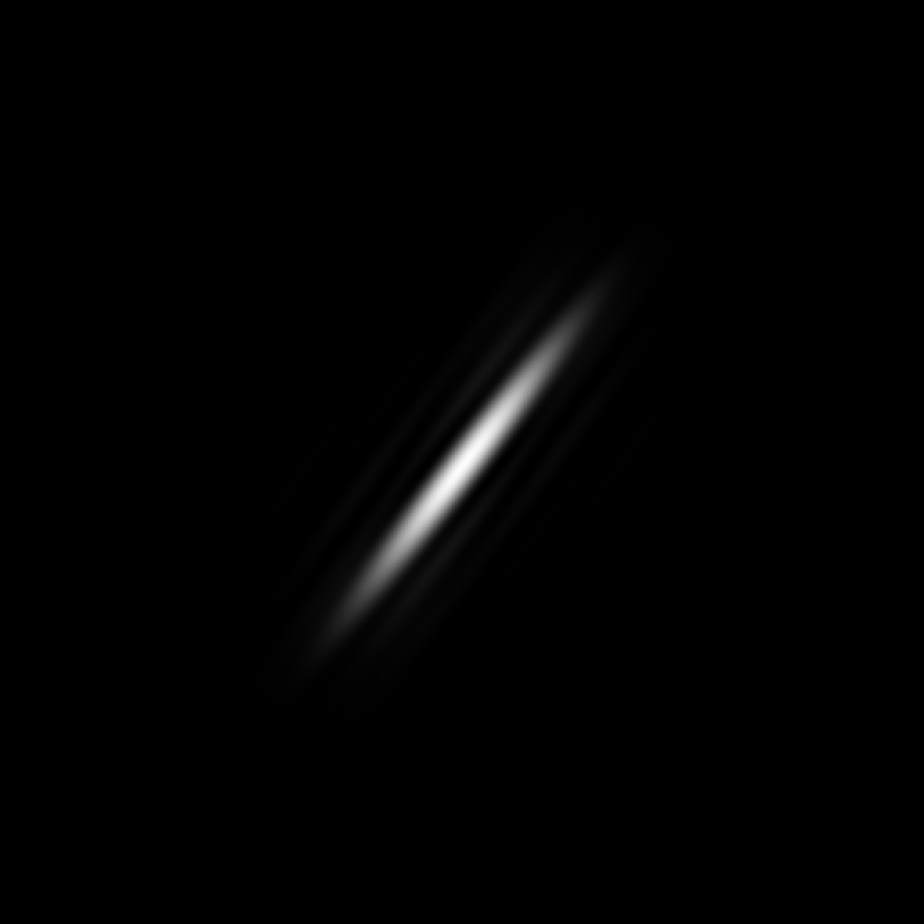}
	&
	\includegraphics[height=\cbarheigth]{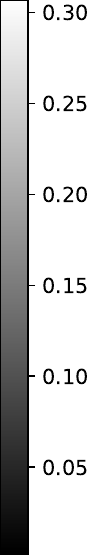}	
	&
	\includegraphics[width=\conddirtribwidth]{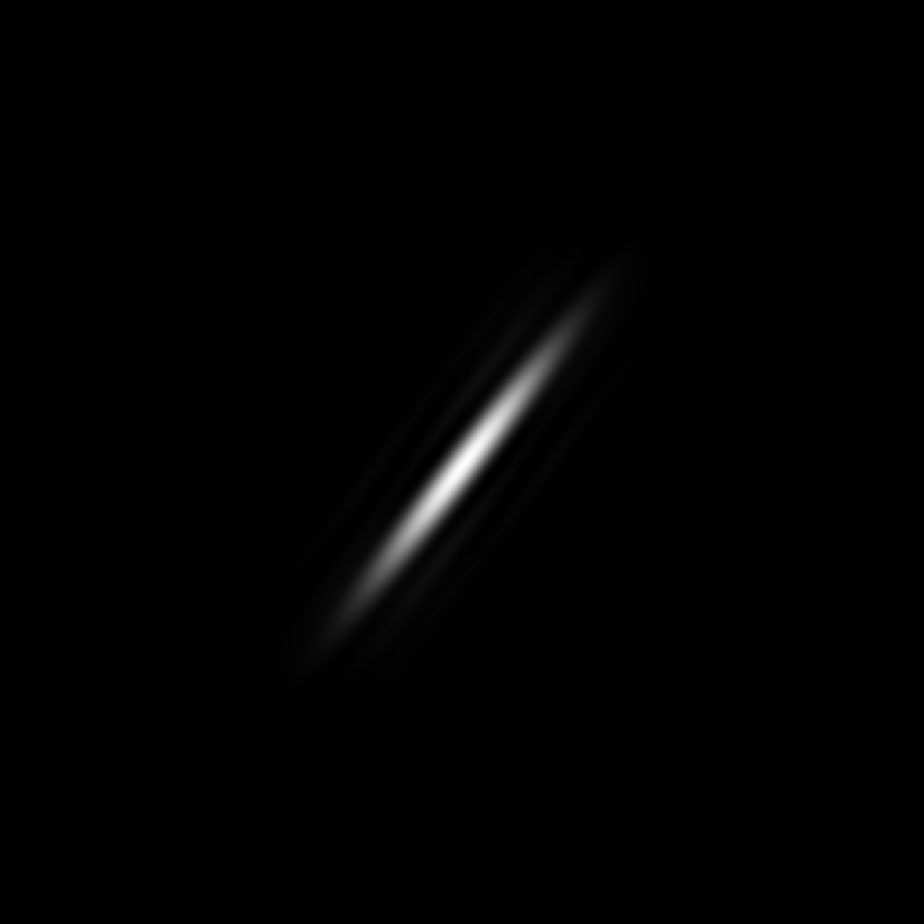}
	&
	\includegraphics[height=\cbarheigth]{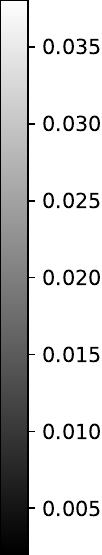}	
	&
	\includegraphics[width=\conddirtribwidth]{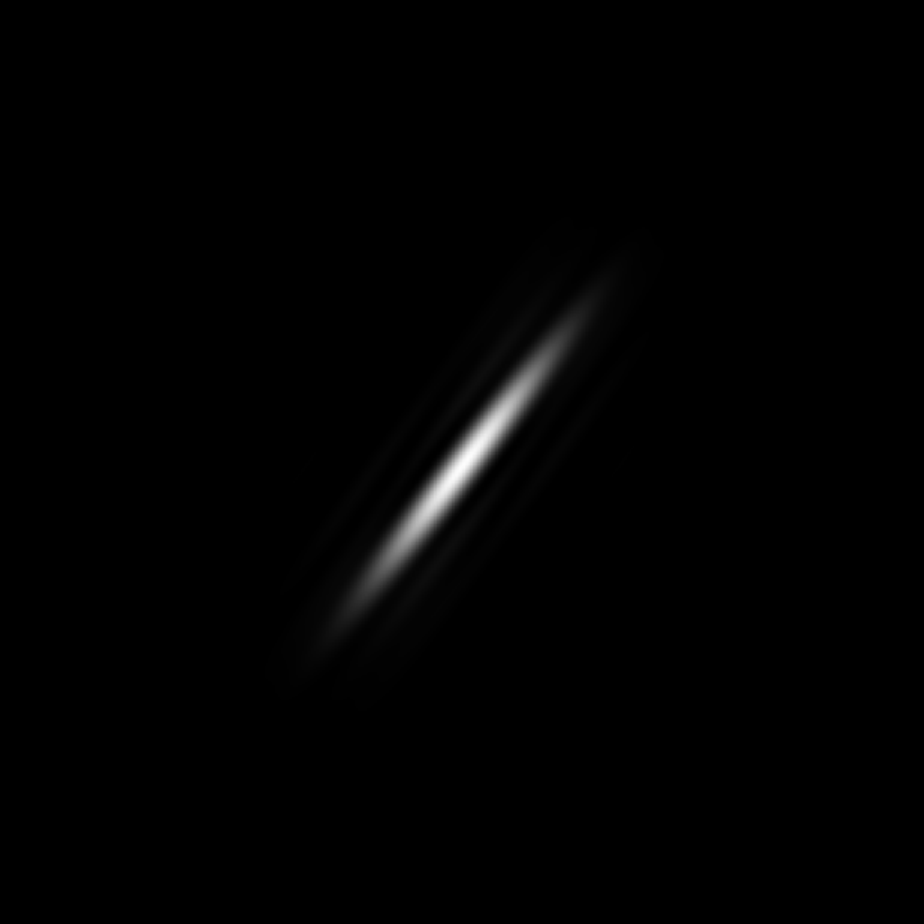}
	&
	\includegraphics[height=\cbarheigth]{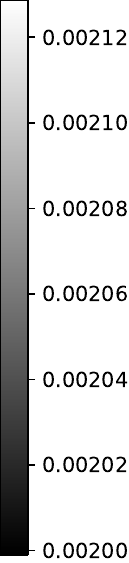}	  \\
	\rotatebox{90}{\parbox{\conddirtribwidth}{\centering DPS }}
	&
	\includegraphics[width=\conddirtribwidth]{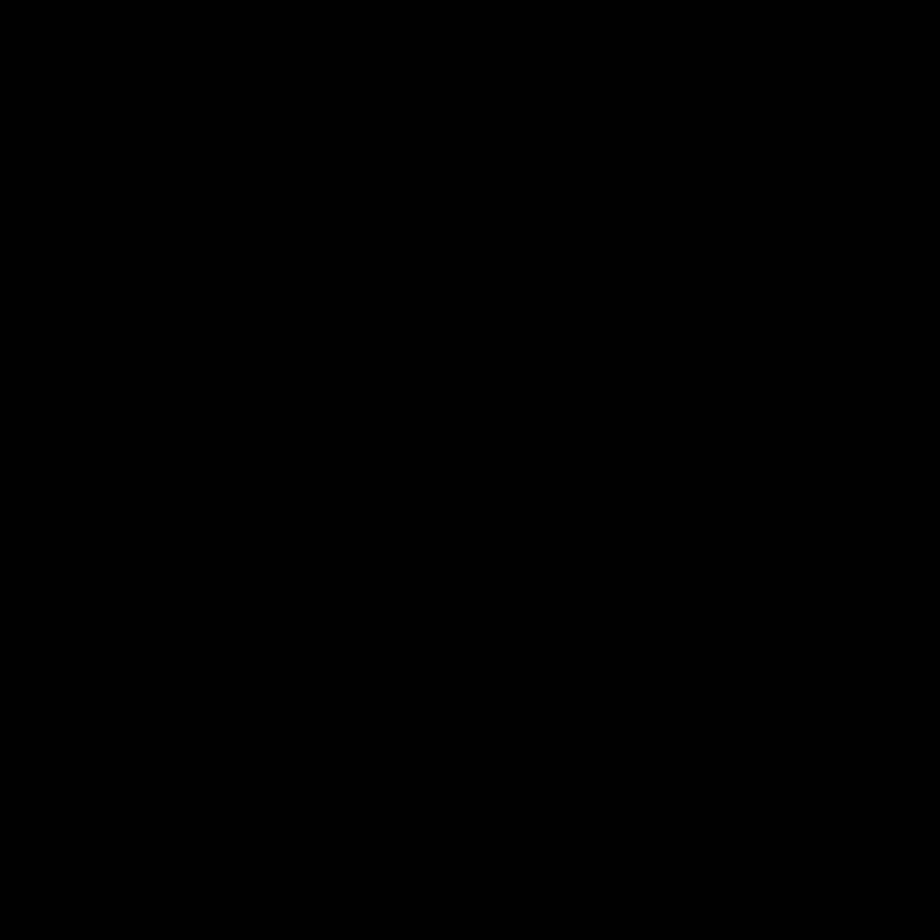}
	&
	\includegraphics[height=\cbarheigth]{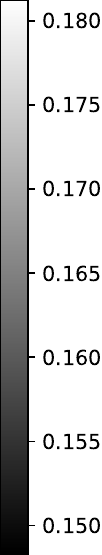}	
	&
	\includegraphics[width=\conddirtribwidth]{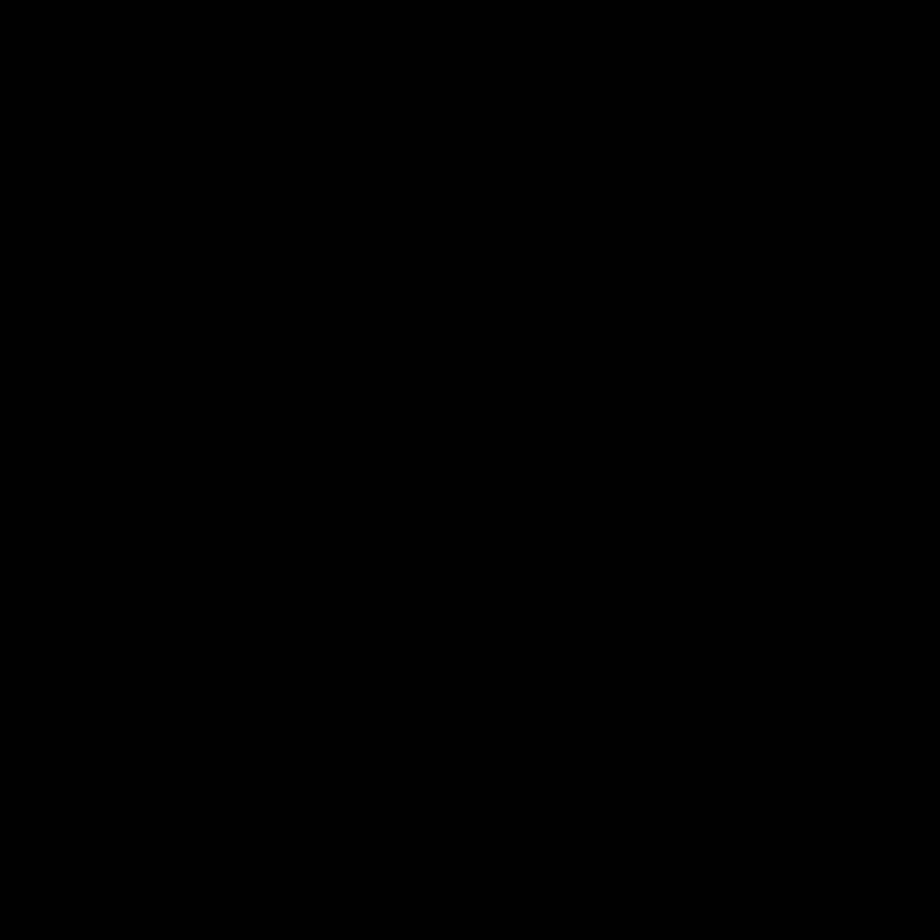}
	&
	\includegraphics[height=\cbarheigth]{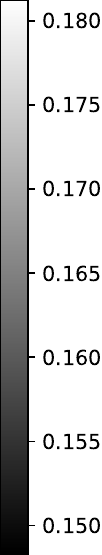}	
	&
	\includegraphics[width=\conddirtribwidth]{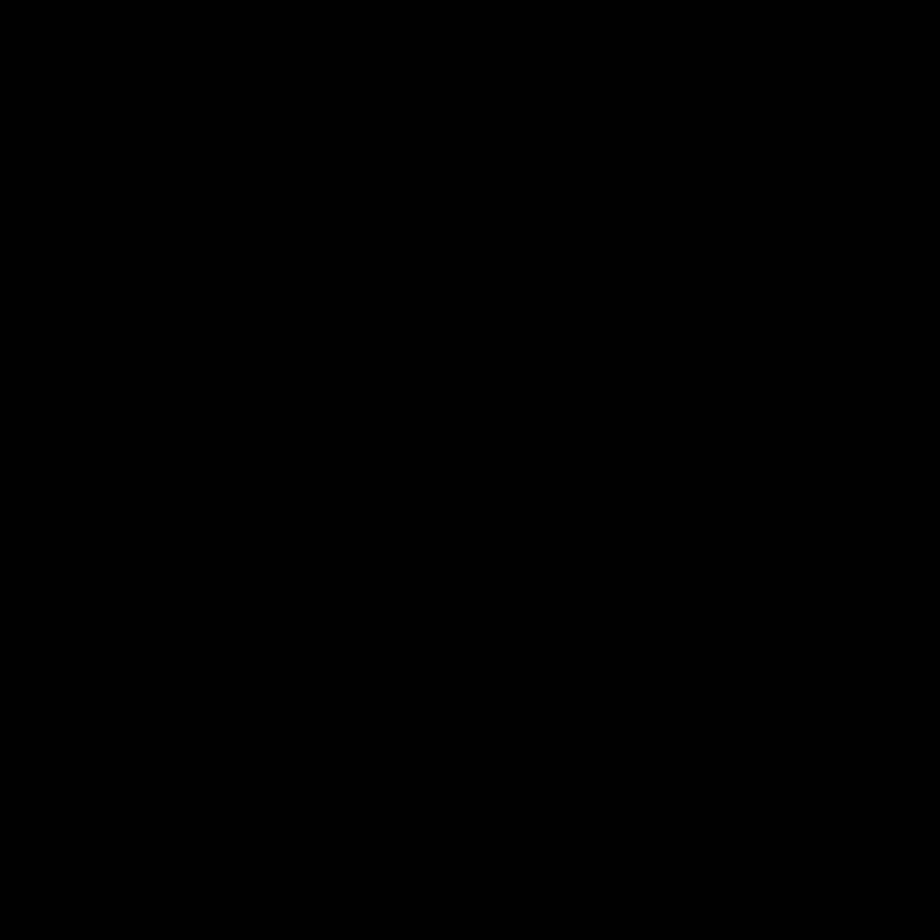}
	&
	\includegraphics[height=\cbarheigth]{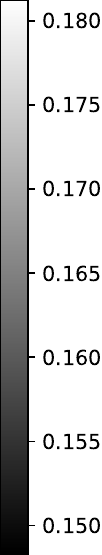}	
	&
	\includegraphics[width=\conddirtribwidth]{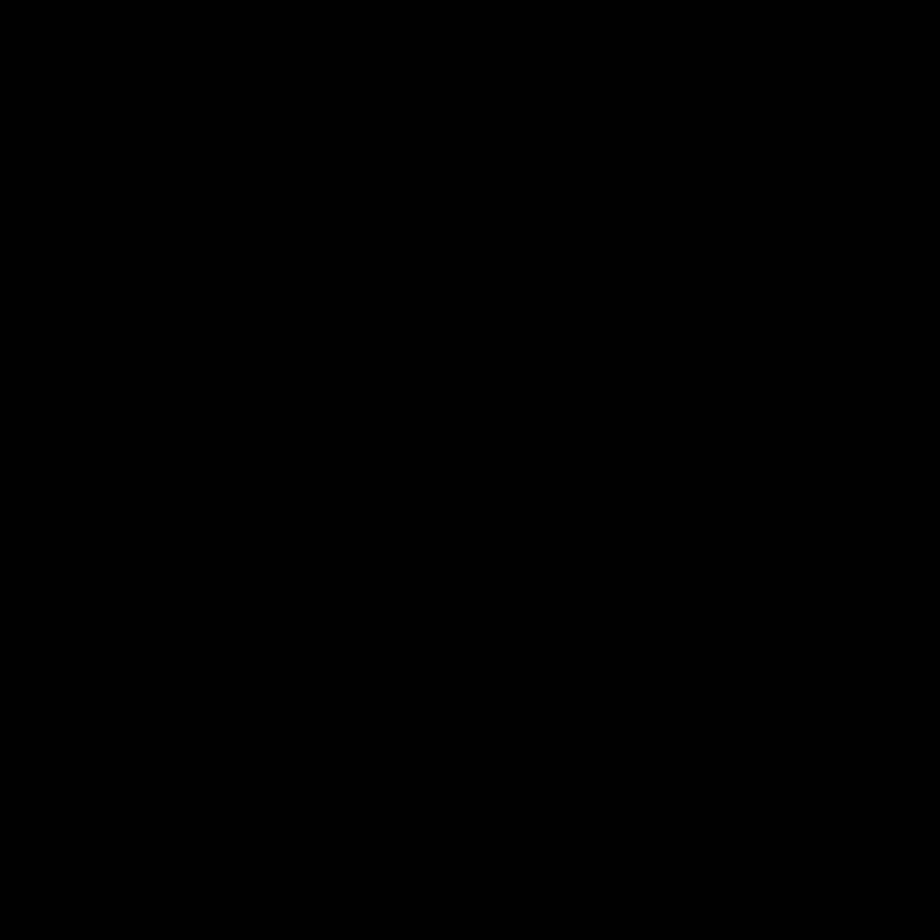}
	&
	\includegraphics[height=\cbarheigth]{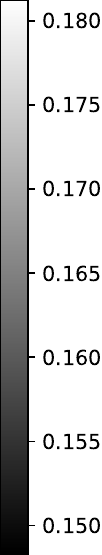}	
	&
	\includegraphics[width=\conddirtribwidth]{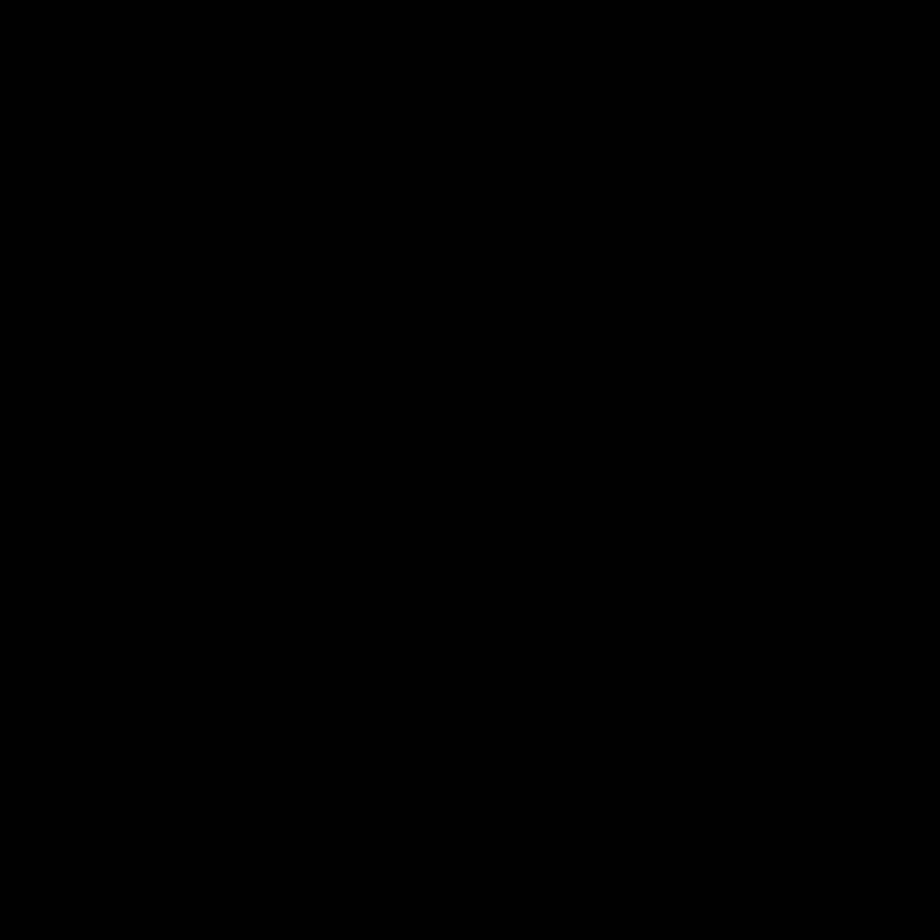}
	&
	\includegraphics[height=\cbarheigth]{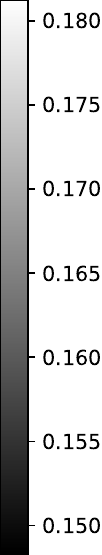}	
	&
	\includegraphics[width=\conddirtribwidth]{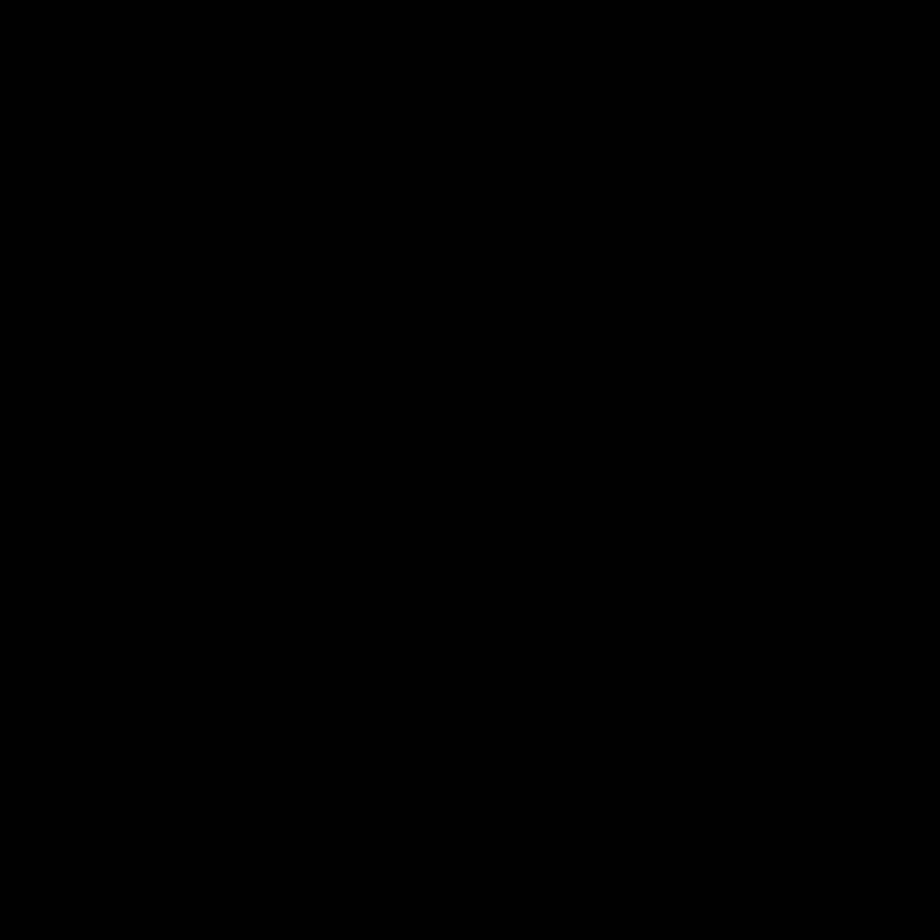}
	&
	\includegraphics[height=\cbarheigth]{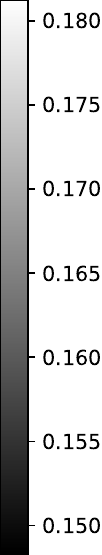}	\\
	\bottomrule
	\end{tabular}
	\caption[Model chosen by the different algorithm for the noisy likelihood $p_t(\V \mid \y_t)$.]{\label{fig:ker_Fabric_motion_blur1} \textbf{Model chosen by the different algorithm for the noisy likelihood $p_t(\V \mid \y_t)$.} DFT of the kernel associated with the distribution $p_t(\V \mid \y_t)$ for the different algorithms at different times for the bicubic kernel, for the fisrt fabric texture in \Cref{fig:W2_SR16}. Note that $\Pi$GDM incorporates the initial motion blur information, whereas the DPS kernel remains constant. CGDM also accounts for the texton information, although the kernel is not perfectly represented. }
\end{figure*}

\setlength{\conddirtribwidth}{0.26\linewidth}

\setlength{\cbarheigth}{\conddirtribwidth}

\begin{figure*}[t]
\centering
	\begin{tabular}{c>{\centering\arraybackslash}p{\conddirtribwidth}c>{\centering\arraybackslash}p{\conddirtribwidth}c>{\centering\arraybackslash}p{\conddirtribwidth}c>{\centering\arraybackslash}p{\conddirtribwidth}c>{\centering\arraybackslash}p{\conddirtribwidth}c>{\centering\arraybackslash}p{\conddirtribwidth}c}
	\toprule
	$t$
	&
	800
	&
	&
	600
	&
	&
	400
	&
	&
	200
	&
	&
	50
	&
	&
	0
	& \\
	\rotatebox{90}{\parbox{\conddirtribwidth}{\centering True theoretical  \\ distribution }}
	&
	\includegraphics[width=\conddirtribwidth]{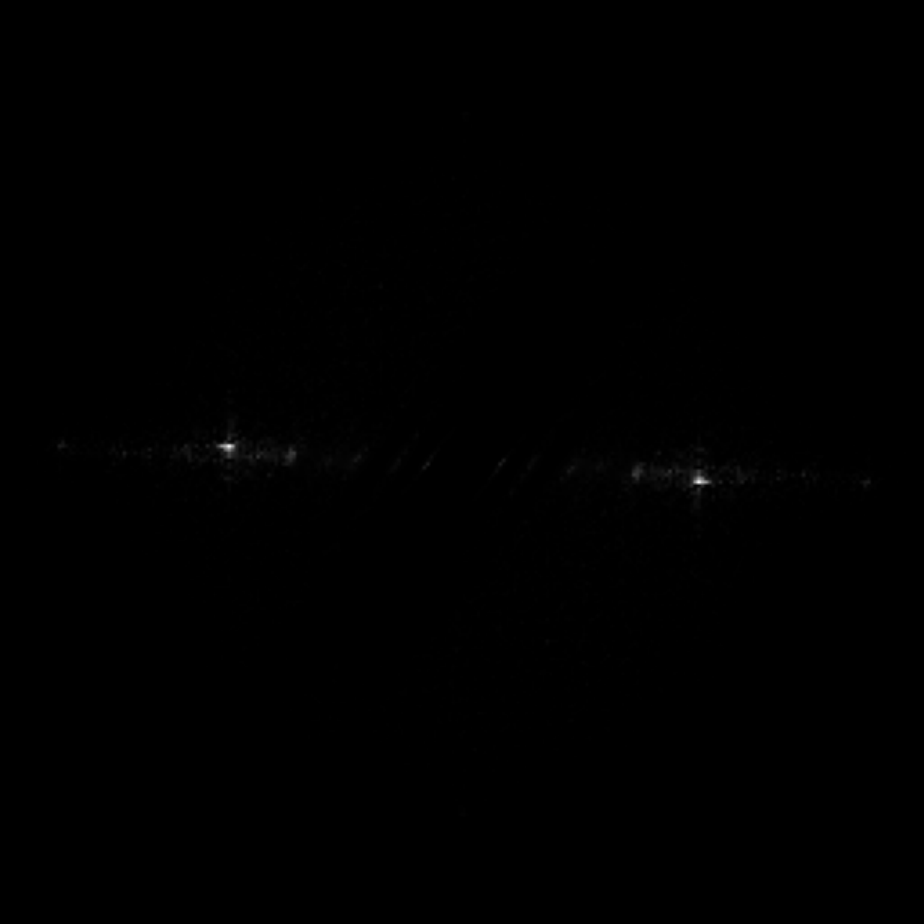}
	&
	\includegraphics[height=\cbarheigth]{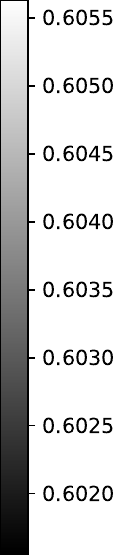} 
	&
	\includegraphics[width=\conddirtribwidth]{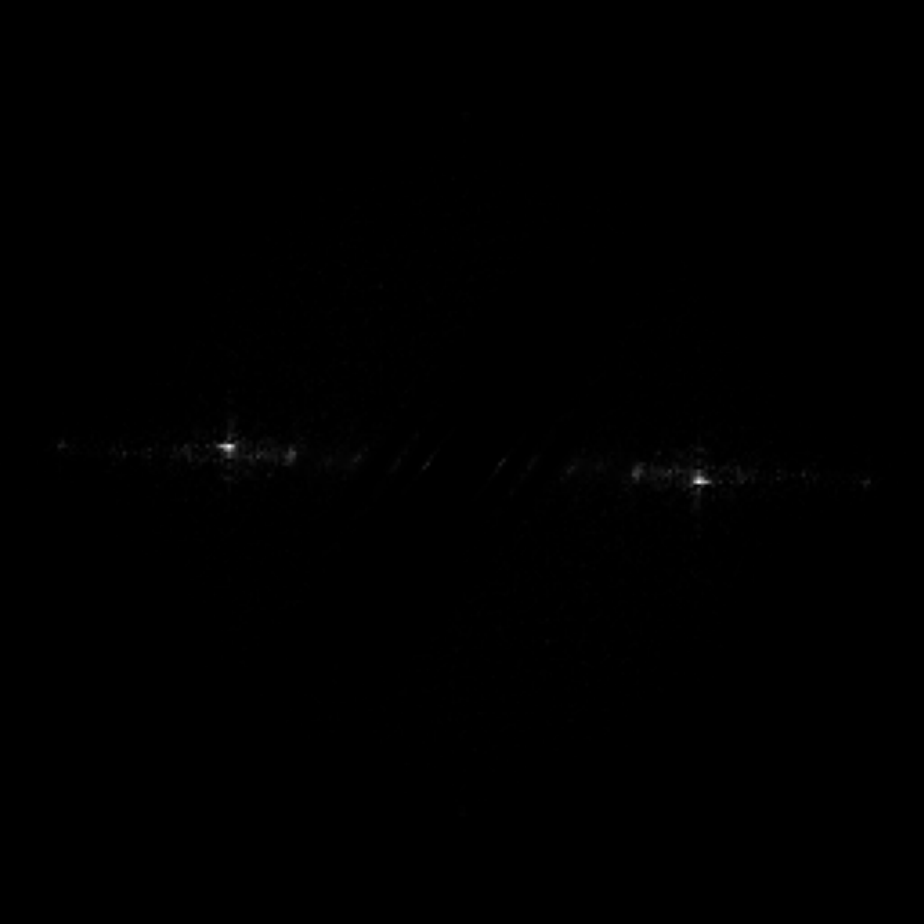}
	&
	\includegraphics[height=\cbarheigth]{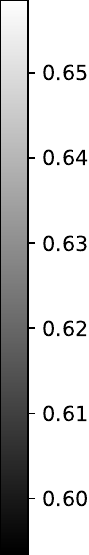}
	&
	\includegraphics[width=\conddirtribwidth]{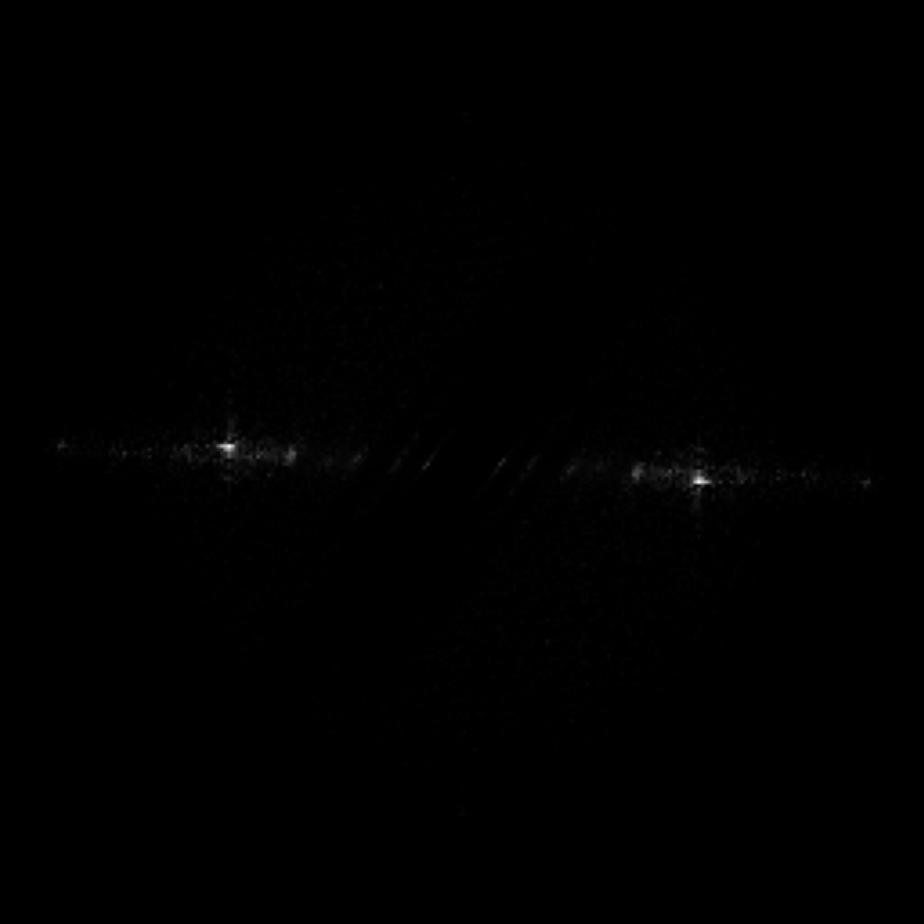}
	&
	\includegraphics[height=\cbarheigth]{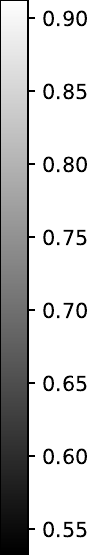}
	&
	\includegraphics[width=\conddirtribwidth]{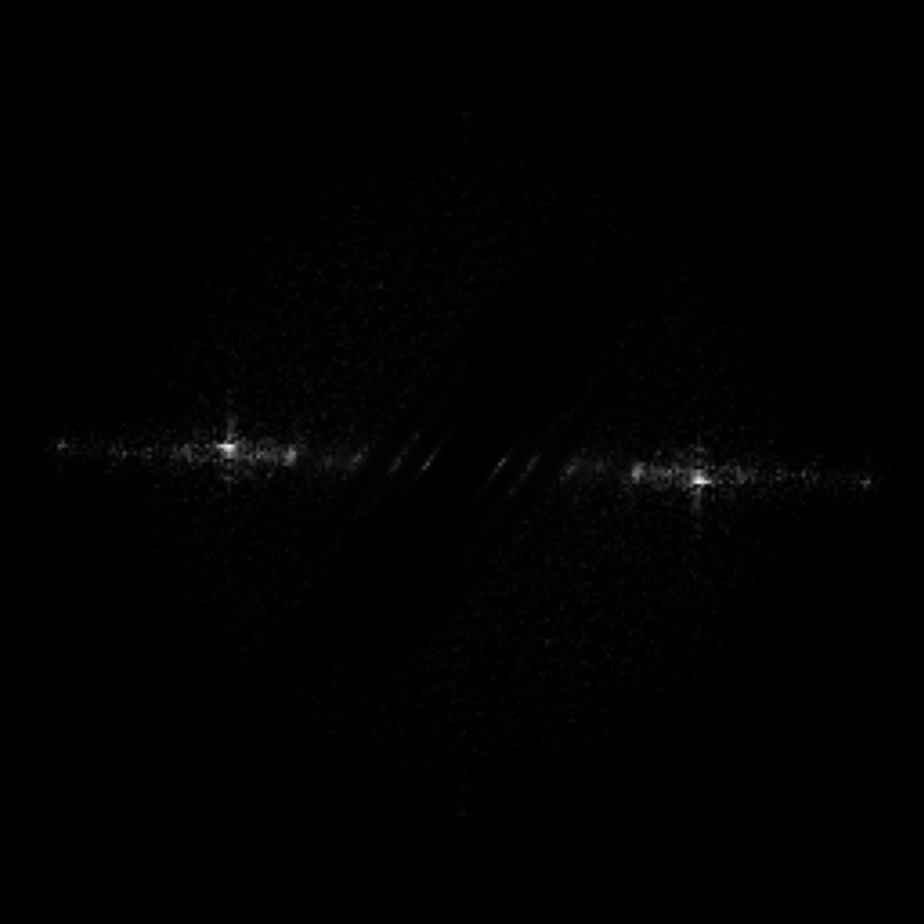}
	&
	\includegraphics[height=\cbarheigth]{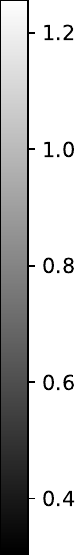}
	&
	\includegraphics[width=\conddirtribwidth]{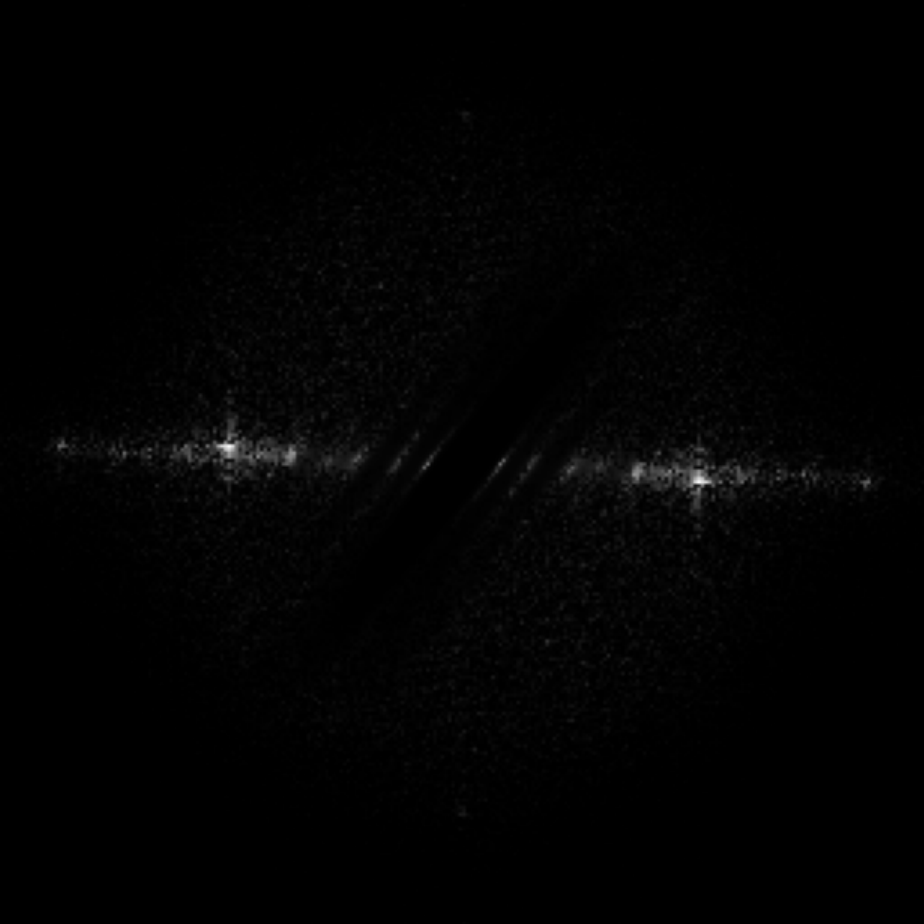}
	&
	\includegraphics[height=\cbarheigth]{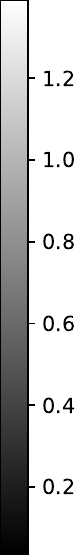}
	&
	\includegraphics[width=\conddirtribwidth]{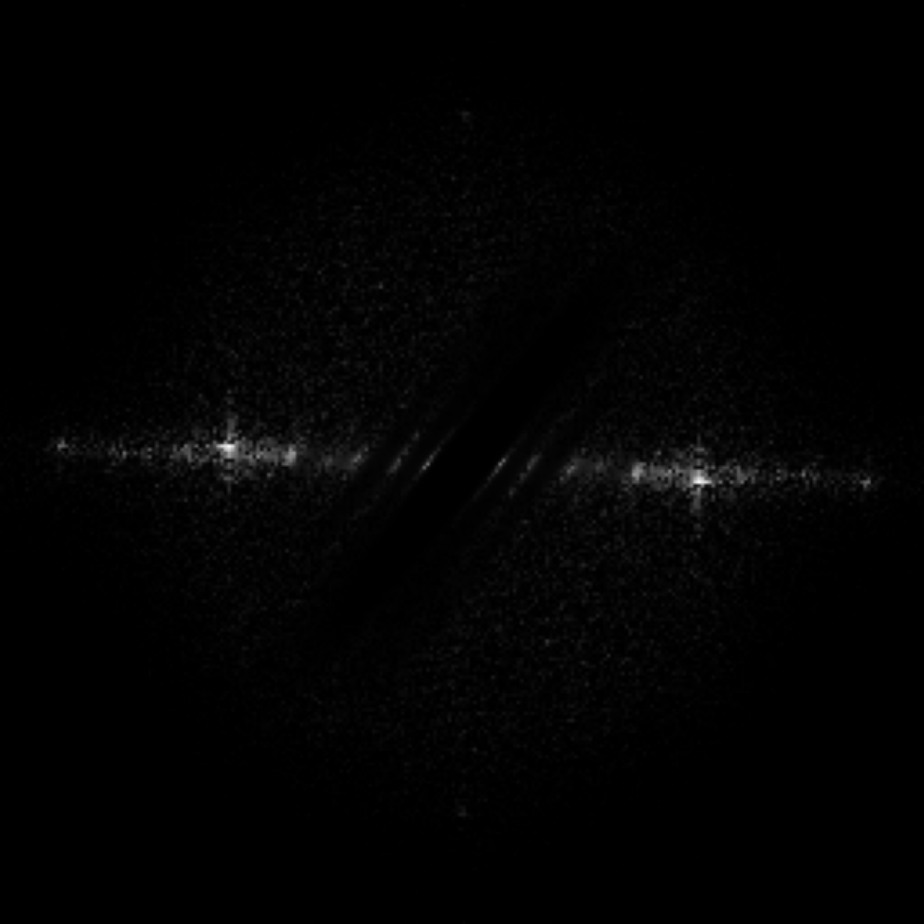}
	&
	\includegraphics[height=\cbarheigth]{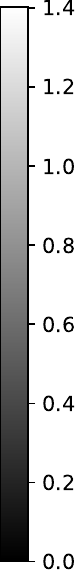} \\
	\rotatebox{90}{\parbox{\conddirtribwidth}{\centering CGDM }}
	&
	\includegraphics[width=\conddirtribwidth]{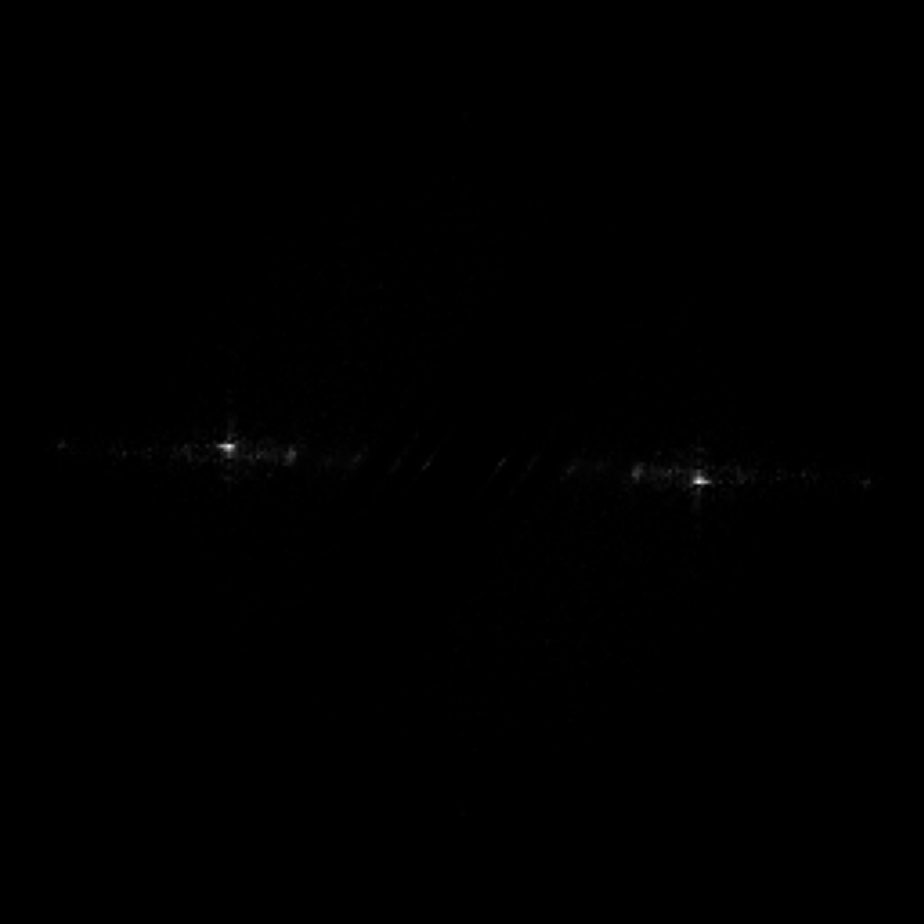}
	&
	\includegraphics[height=\cbarheigth]{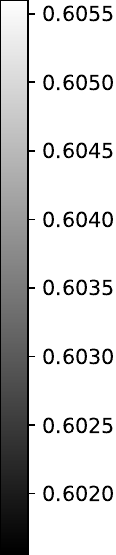}
	&
	\includegraphics[width=\conddirtribwidth]{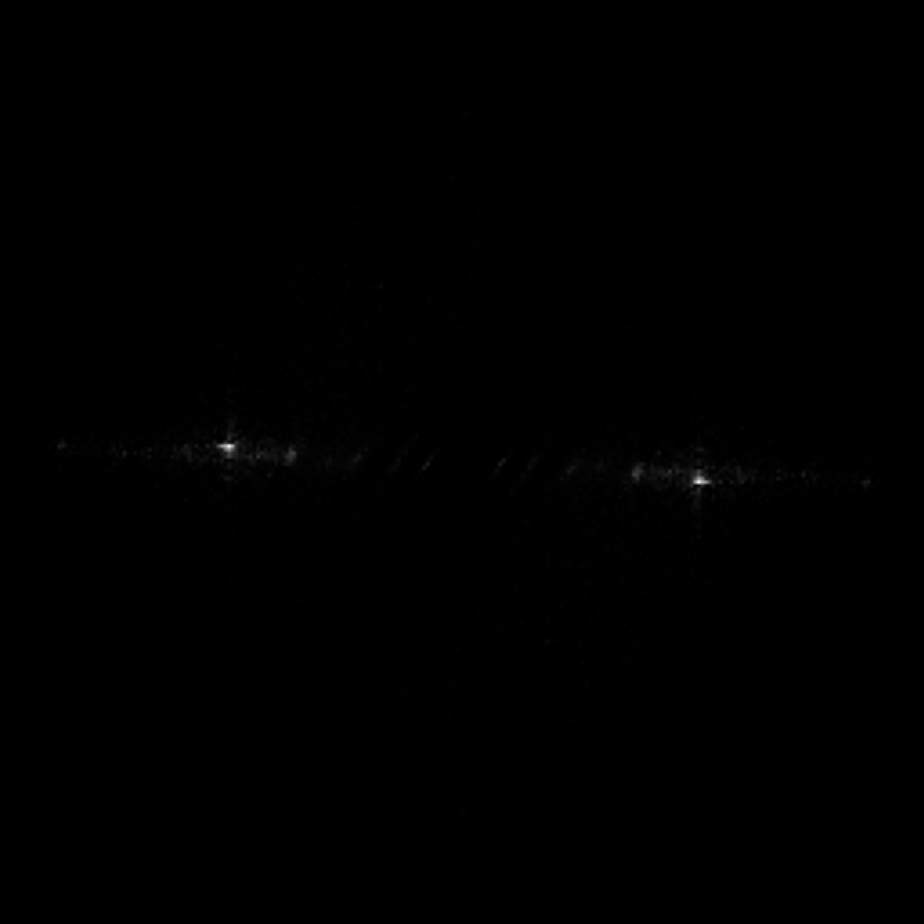}
	&
	\includegraphics[height=\cbarheigth]{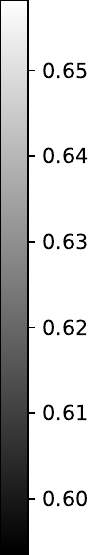}
	&
	\includegraphics[width=\conddirtribwidth]{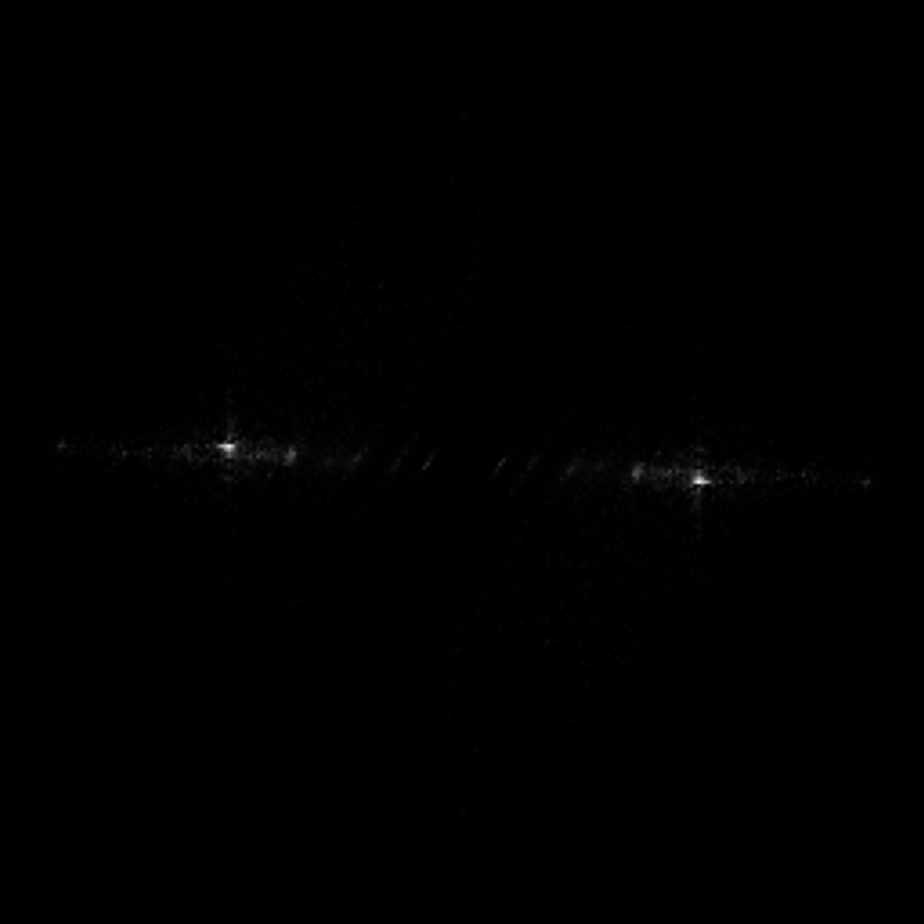}
	&
	\includegraphics[height=\cbarheigth]{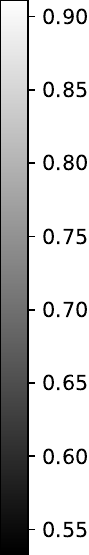}
	&
	\includegraphics[width=\conddirtribwidth]{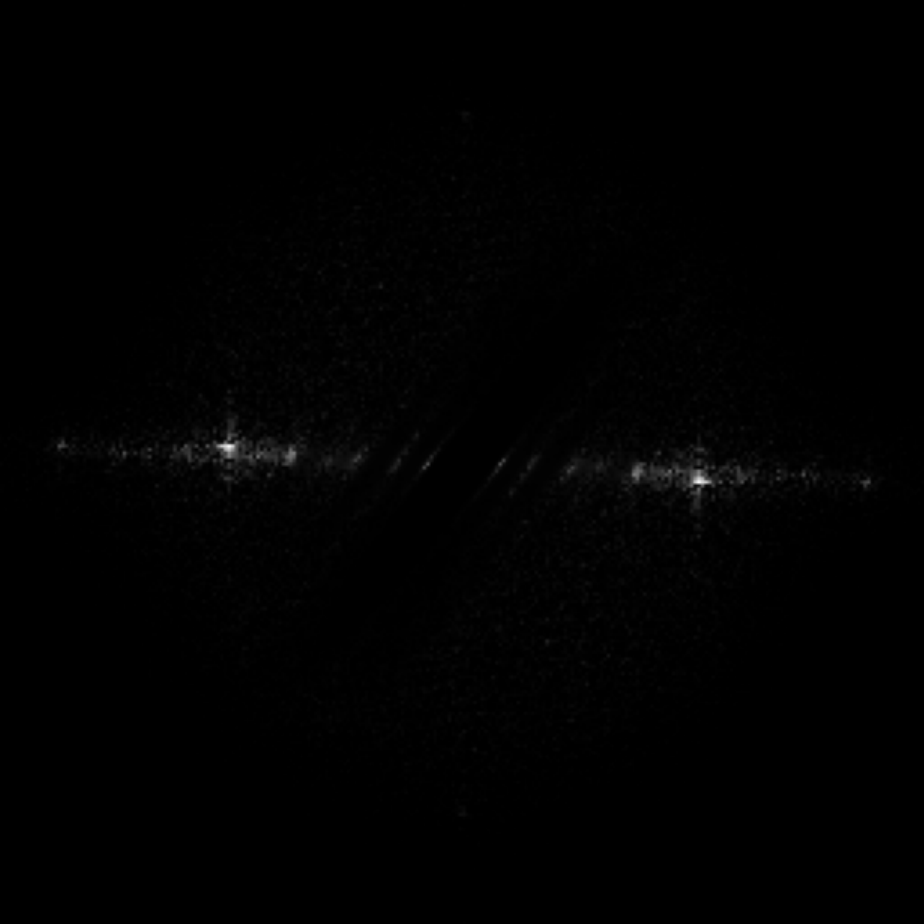}
	&
	\includegraphics[height=\cbarheigth]{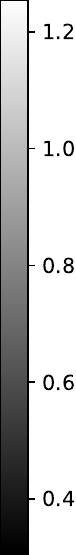}
	&
	\includegraphics[width=\conddirtribwidth]{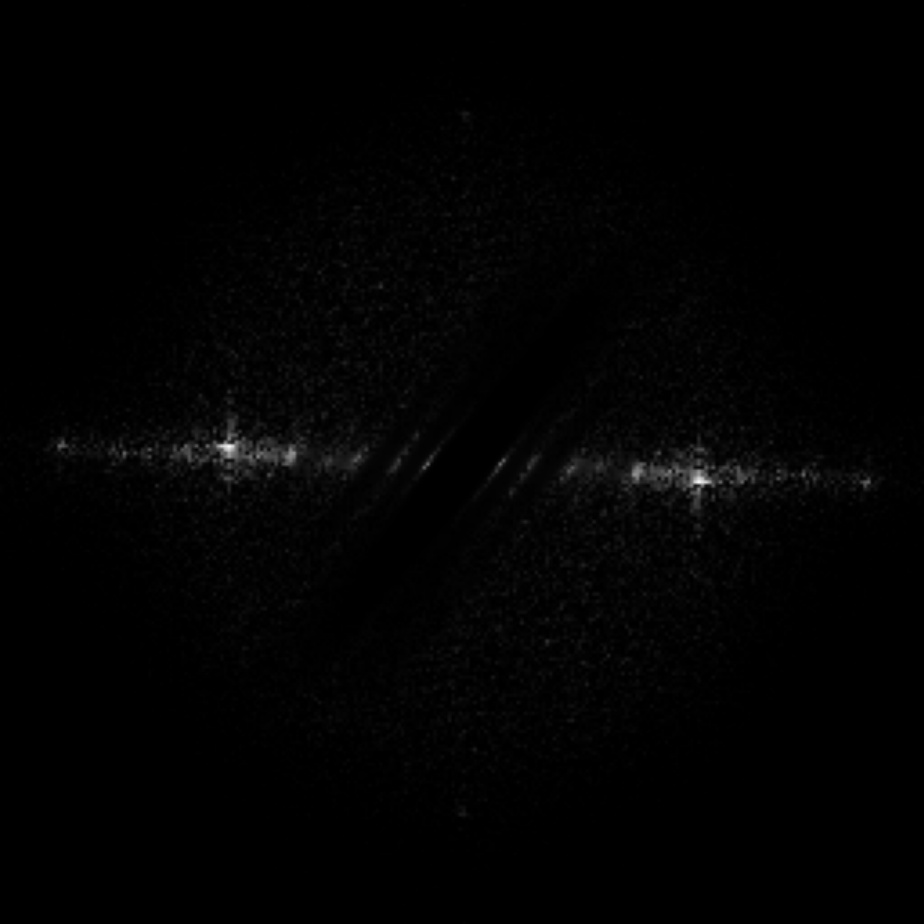}
	&
	\includegraphics[height=\cbarheigth]{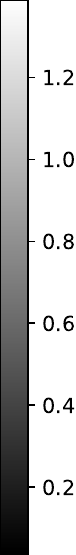}
	&
	\includegraphics[width=\conddirtribwidth]{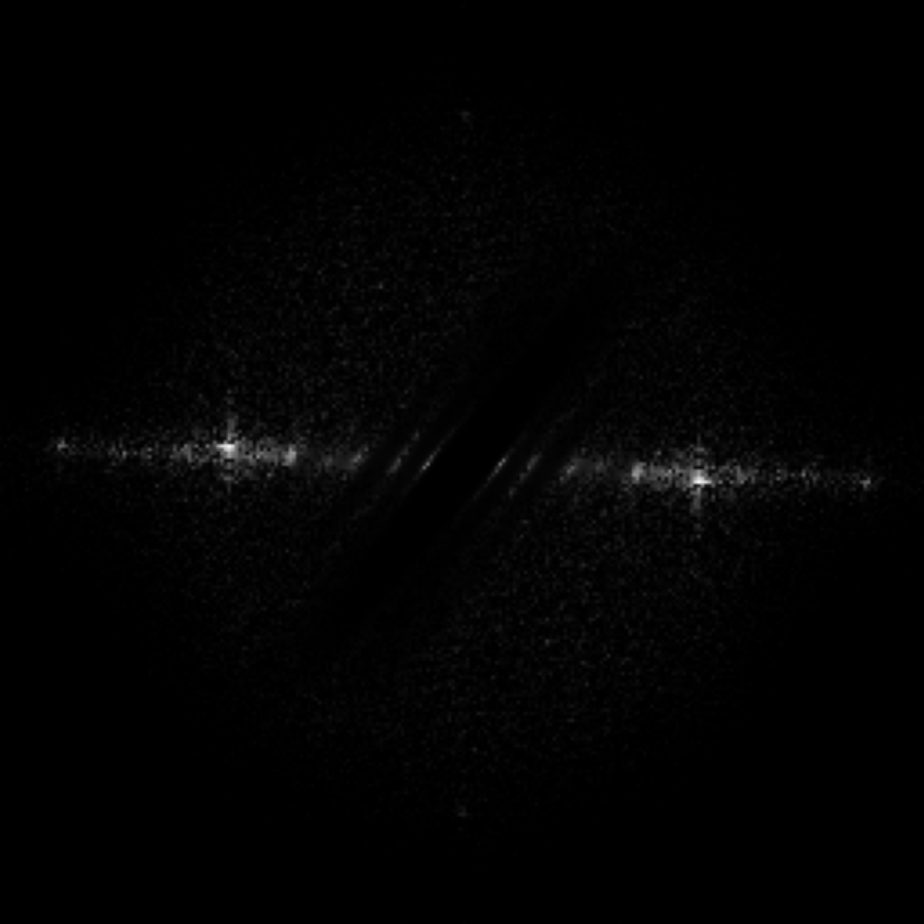}
	&
	\includegraphics[height=\cbarheigth]{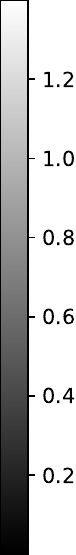}	 \\
	\rotatebox{90}{\parbox{\conddirtribwidth}{\centering $\Pi$GDM }}
		&
	\includegraphics[width=\conddirtribwidth]{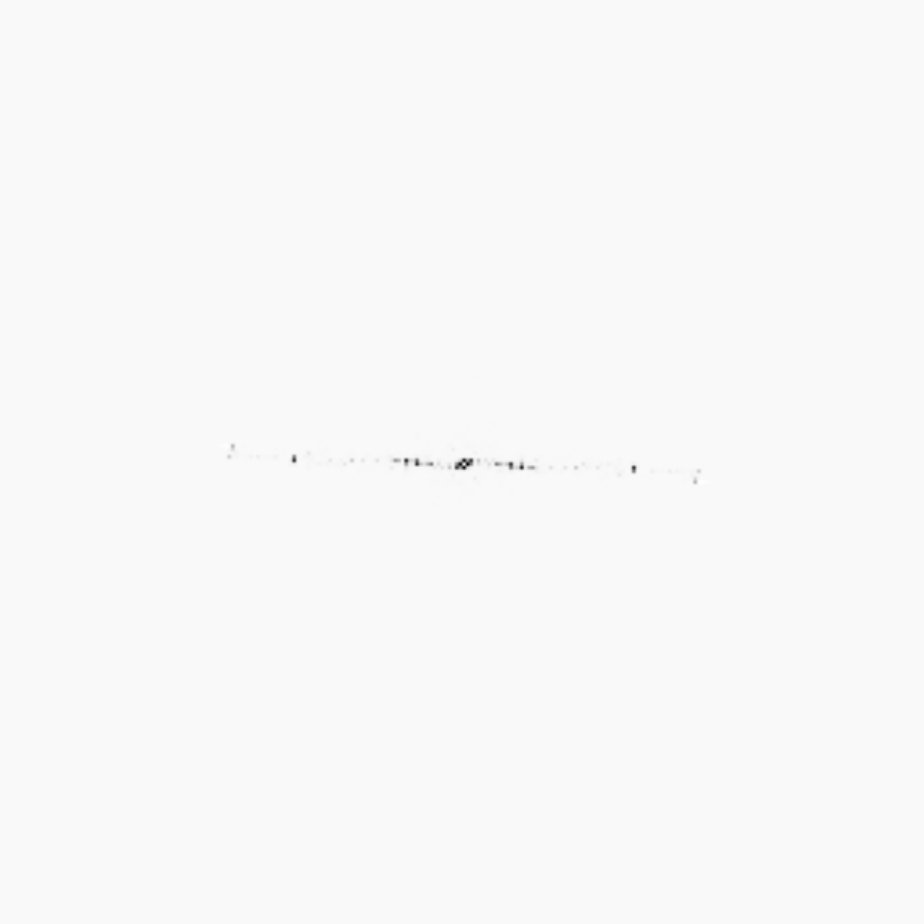}
	&
	\includegraphics[height=\cbarheigth]{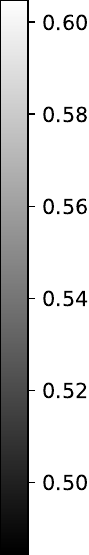}
	&
	\includegraphics[width=\conddirtribwidth]{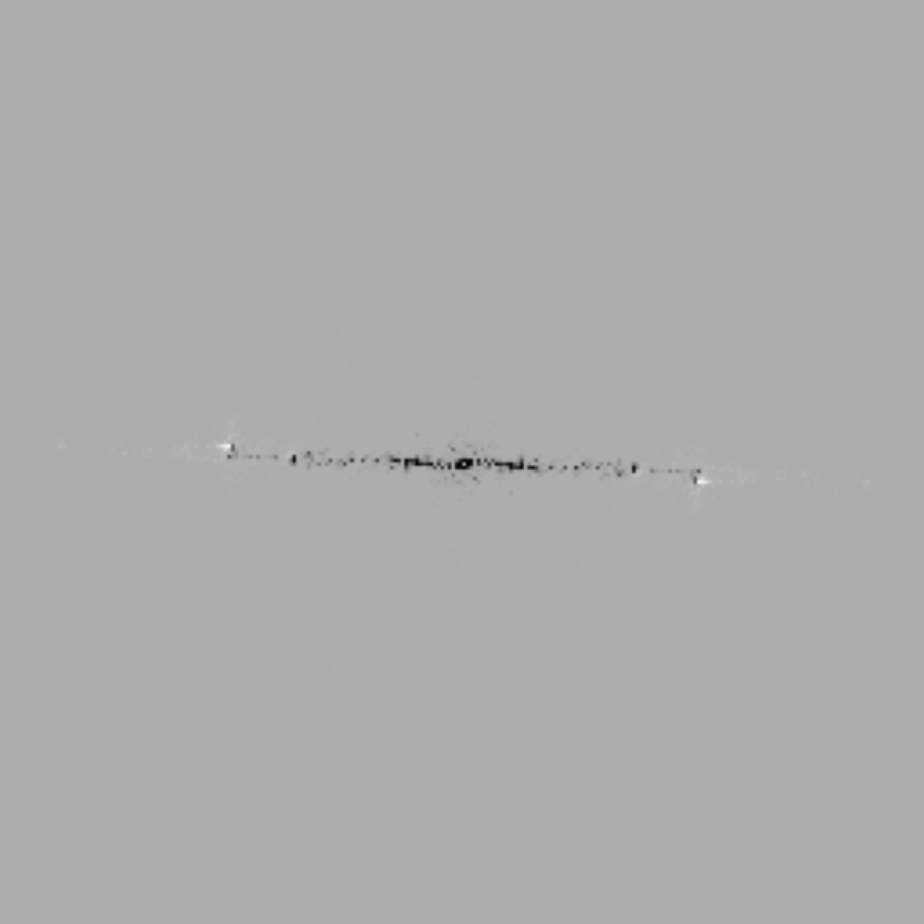}
	&
	\includegraphics[height=\cbarheigth]{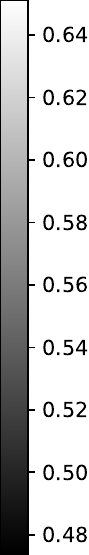}
	&
	\includegraphics[width=\conddirtribwidth]{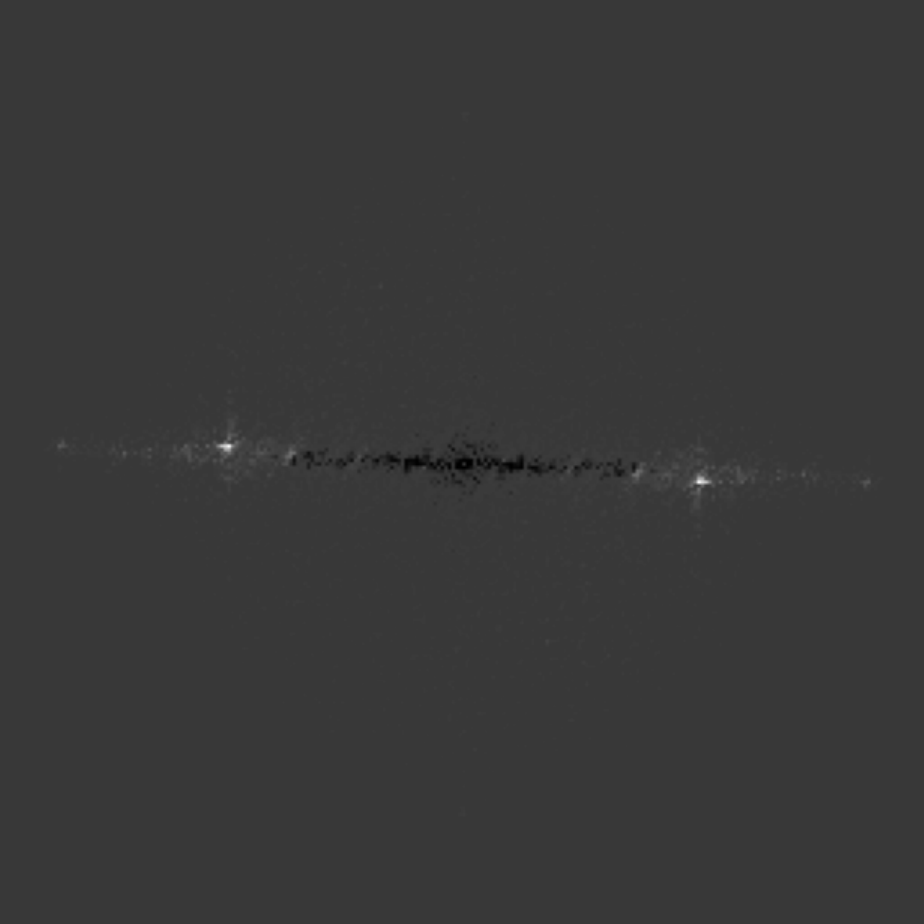}
	&
	\includegraphics[height=\cbarheigth]{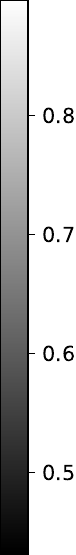}
	&
	\includegraphics[width=\conddirtribwidth]{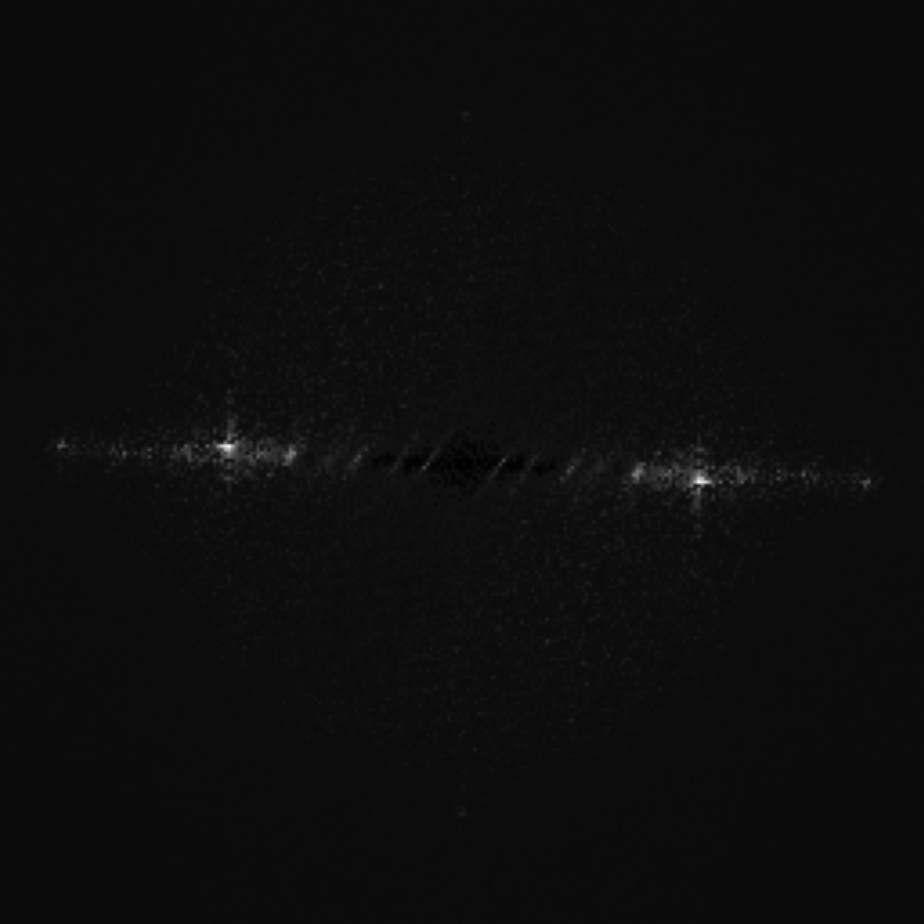}
	&
	\includegraphics[height=\cbarheigth]{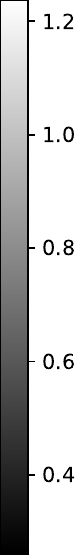}
	&
	\includegraphics[width=\conddirtribwidth]{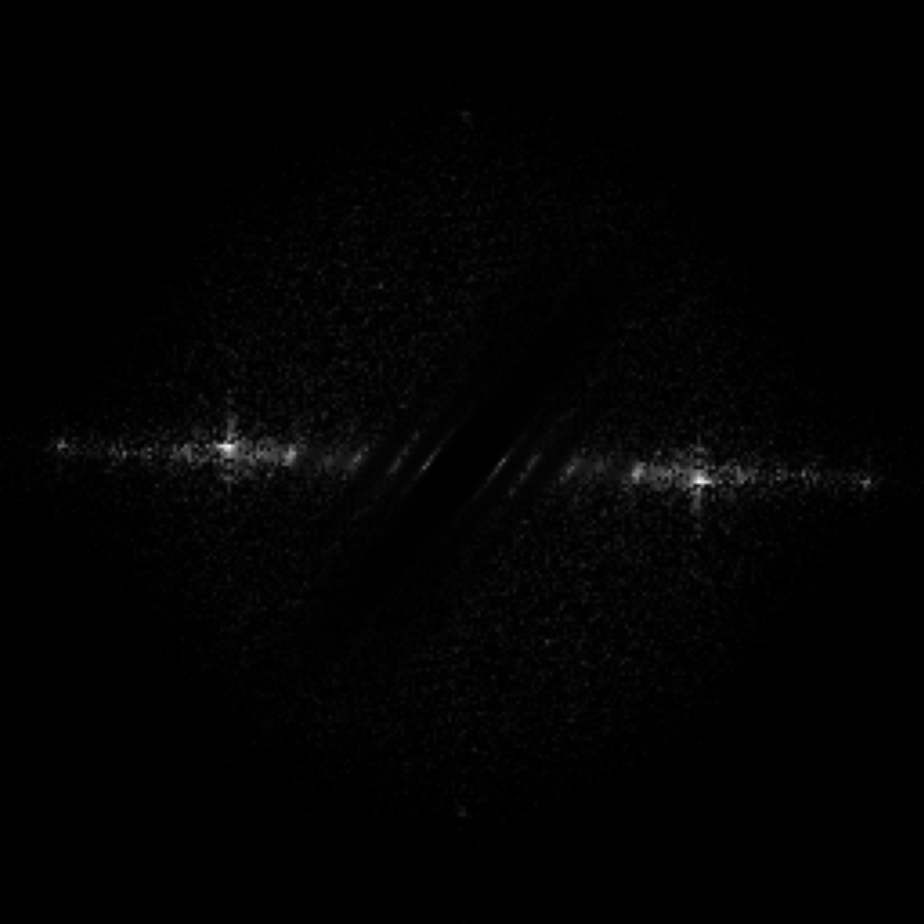}
	&
	\includegraphics[height=\cbarheigth]{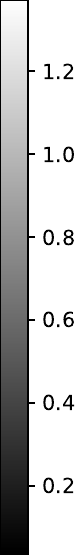}
	&
	\includegraphics[width=\conddirtribwidth]{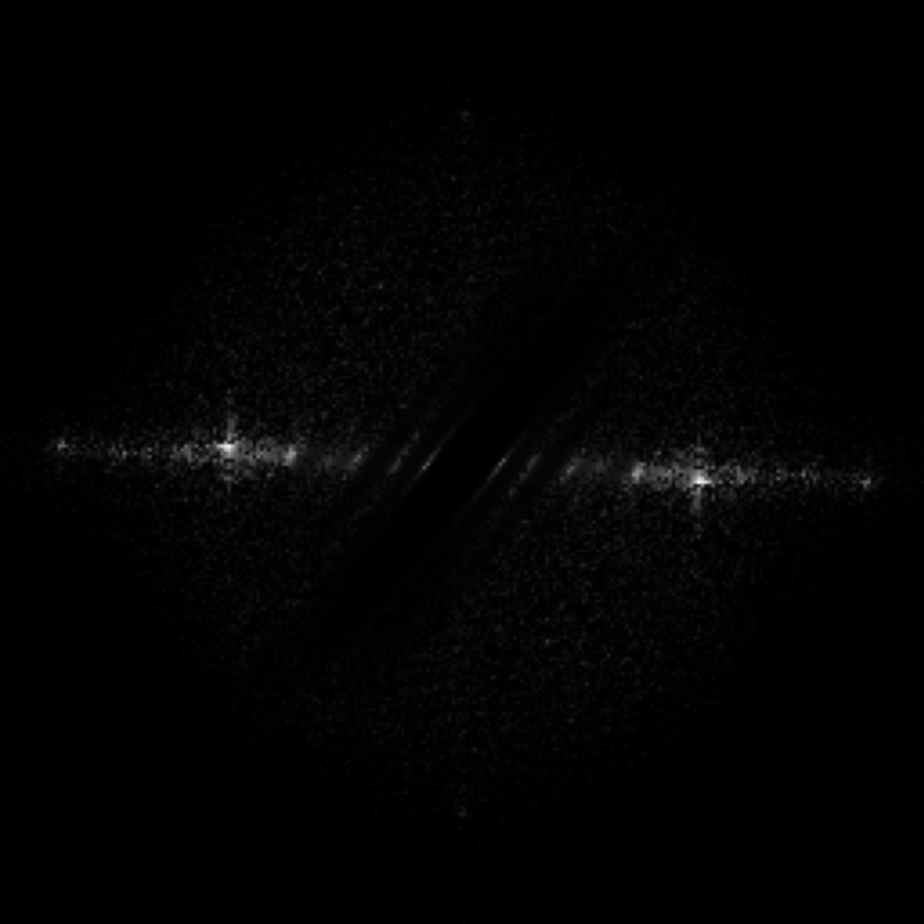}
	&
	\includegraphics[height=\cbarheigth]{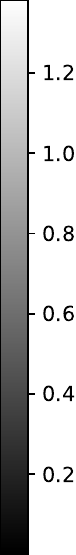}	\\
	\rotatebox{90}{\parbox{\conddirtribwidth}{\centering DPS }}
			&
	\includegraphics[width=\conddirtribwidth]{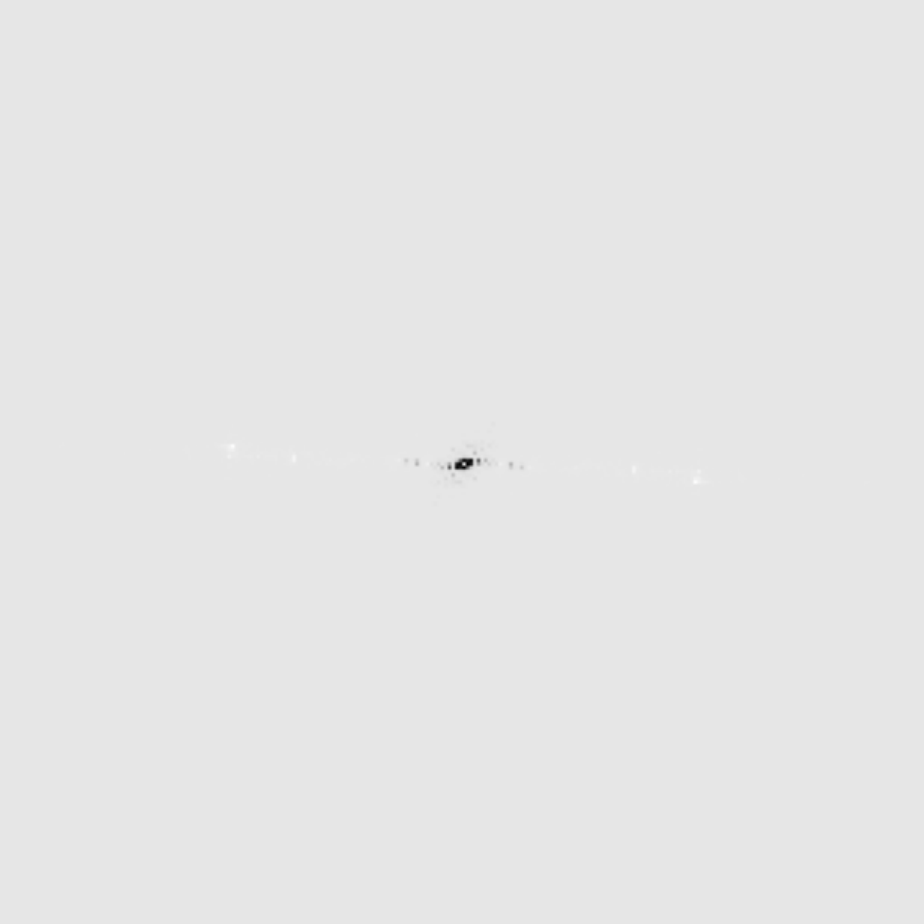}
	&
	\includegraphics[height=\cbarheigth]{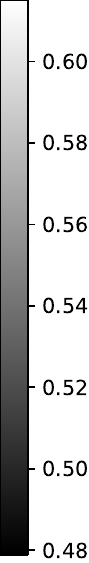}
	&
	\includegraphics[width=\conddirtribwidth]{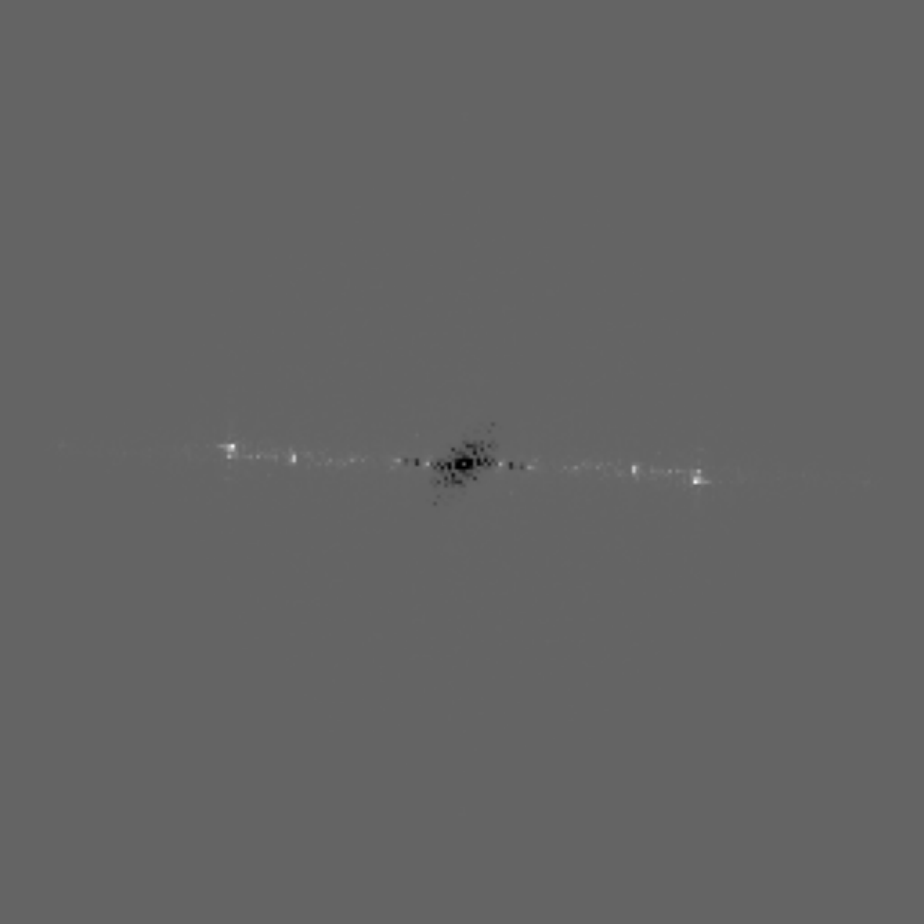}
	&
	\includegraphics[height=\cbarheigth]{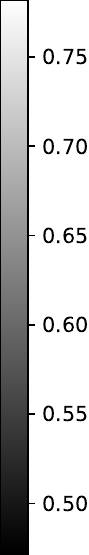}
	&
	\includegraphics[width=\conddirtribwidth]{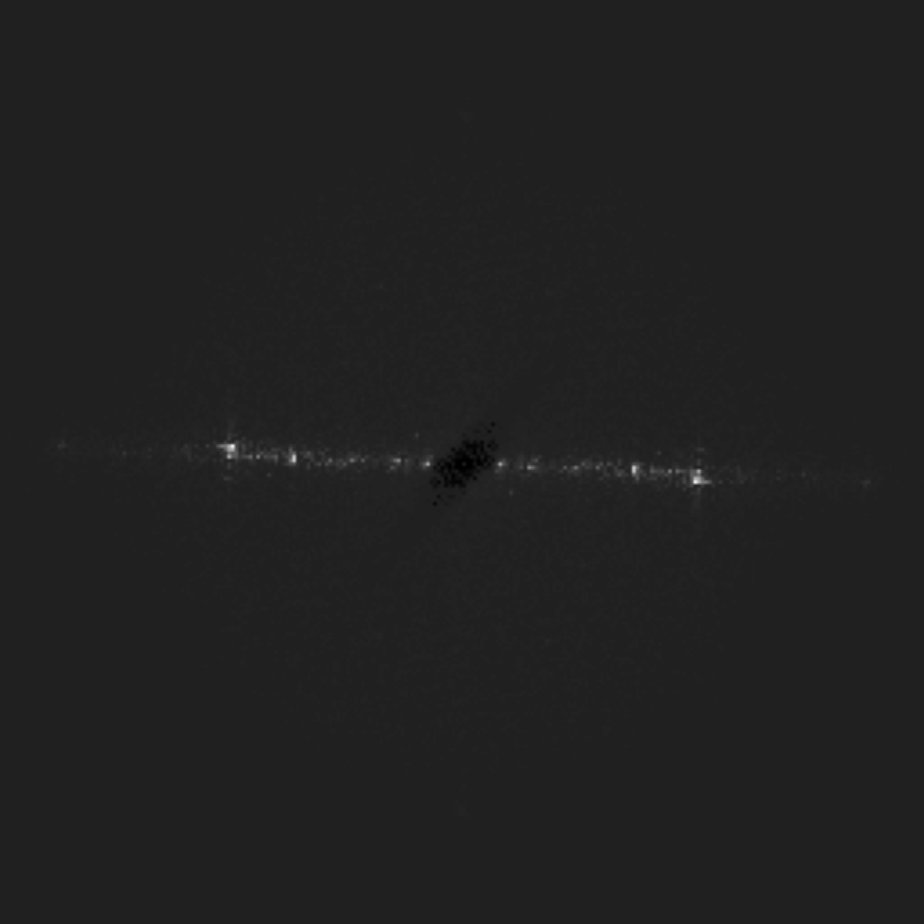}
	&
	\includegraphics[height=\cbarheigth]{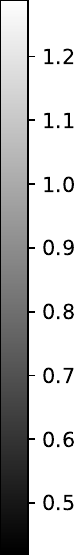}
	&
	\includegraphics[width=\conddirtribwidth]{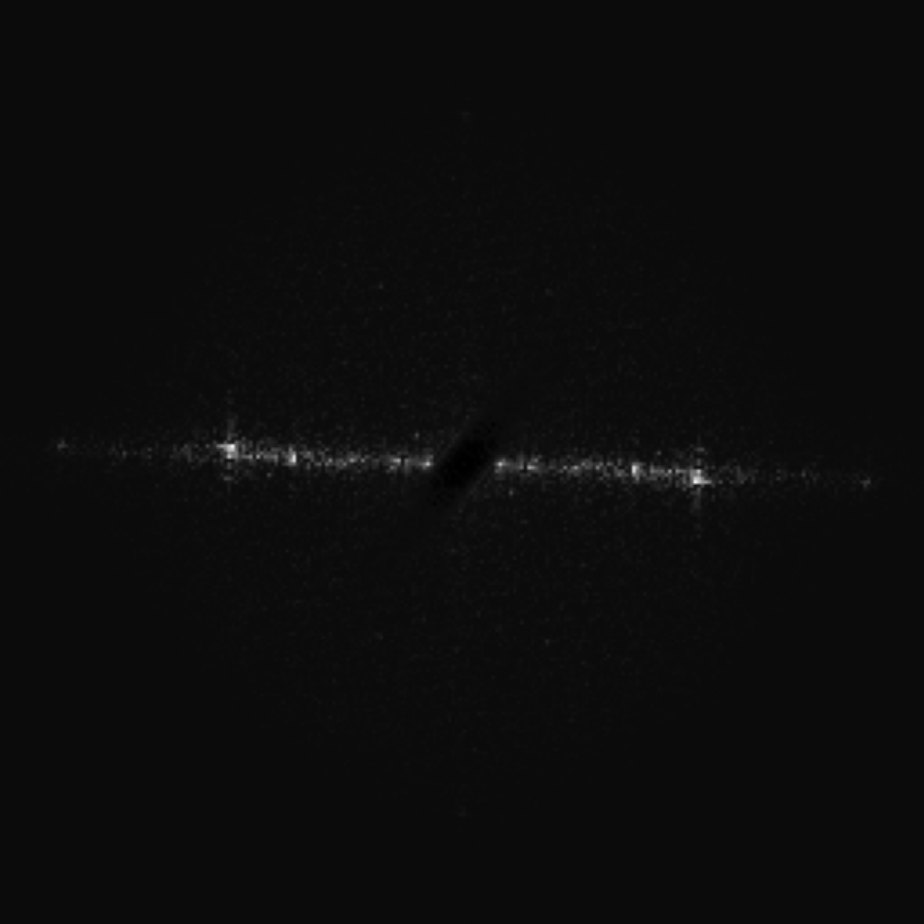}
	&
	\includegraphics[height=\cbarheigth]{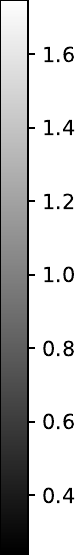}
	&
	\includegraphics[width=\conddirtribwidth]{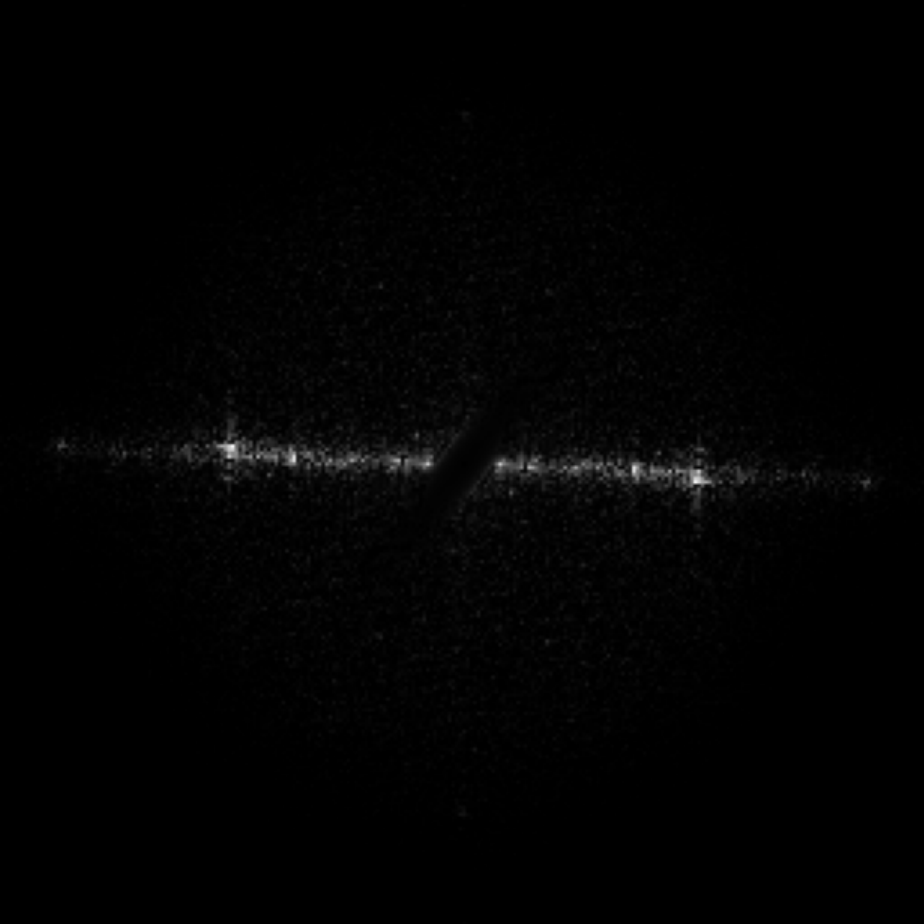}
	&
	\includegraphics[height=\cbarheigth]{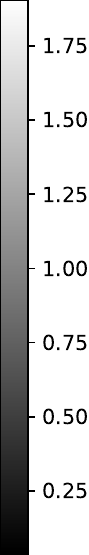}
	&
	\includegraphics[width=\conddirtribwidth]{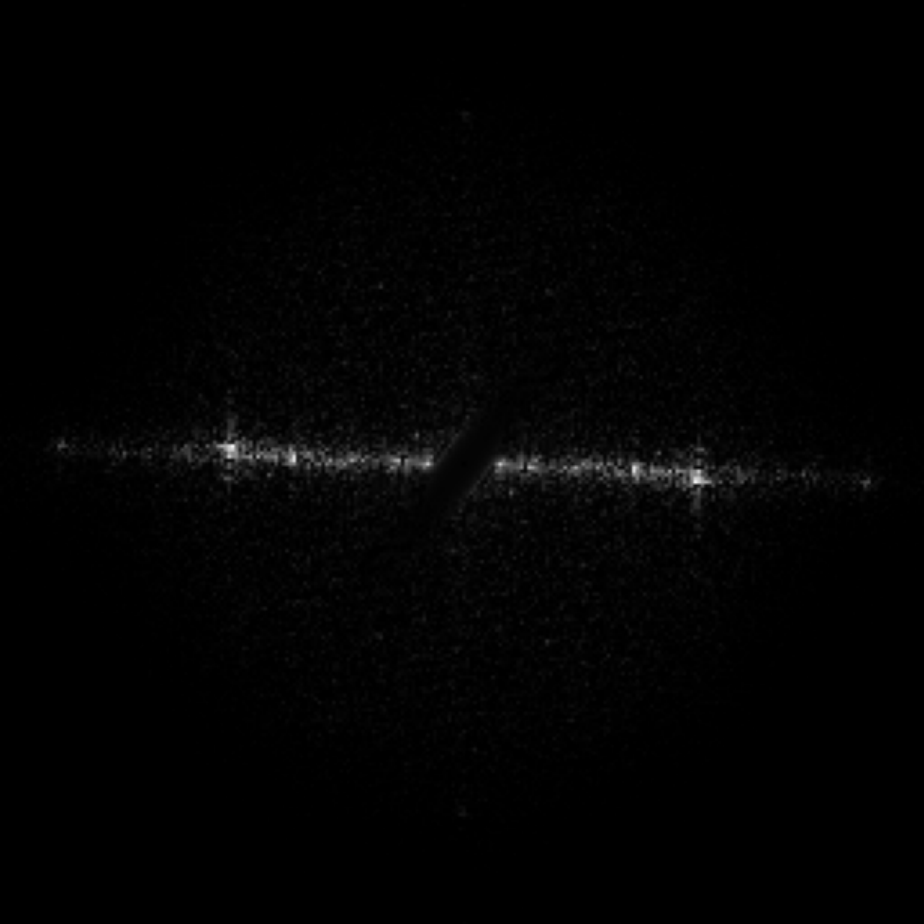}
	&
	\includegraphics[height=\cbarheigth]{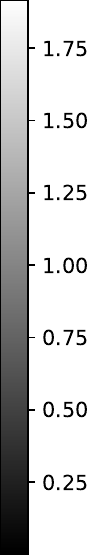}	\\
	\bottomrule
	\end{tabular}
	\caption[Distribution of the backward processes along the time.]{\label{fig:cond_distrib_Fabric_SR16} \textbf{Distribution of the backward processes along the time.} DFT of the kernel associated with the distribution of the backward processes generated by the different algorithms at different times for the first motion blur, for the fisrt fabric texture in \Cref{fig:W2_SR16}. Observe that the distributions of $\Pi$GDM and CGDM are well aligned with the theoretical true conditional distribution near the final times, including $t = 50$.}
\end{figure*}

\section{Discussion}
\label{sec:discussion}
{The bias induced by the two algorithms, DPS and $\Pi$GDM, raises questions about their suitability for uncertainty quantification.}
However, despite its advantages, it is important to note that CGDM is significantly more computationally expensive than $\Pi$GDM. The exact Gaussian computations required by CGDM introduce a higher complexity, which may prevent its deployment in large-scale. 
On the other hand, $\Pi$GDM, while slightly less accurate, provides a much more computationally efficient alternative, making it the preferred method in practical scenarios where the covariance matrix cannot be quickly inverted (see the case of super-resolution below). 
As a result, $\Pi$GDM appears to be the go-to approach for most real-world applications of conditional diffusion models, striking a balance between accuracy and computational feasibility. We discuss below the extension of our study of the CGDM algorithm to more general inverse problems.

\subsection{Extension to the SR inverse problem for the Gaussian microtextures}
Let us discuss the extension of our work to non diagonal inverse problems with a focus on the super-resolution (SR) problem. 
 Let us consider $\A = \Sub \bC$ where $\Sub$ is a subsampling operator with stride $r \in \mathbb{N}$ and $\bC$ is a convolution operator. The conditional sampling of this inverse problem has been considered and solved in \cite{pierret_stochastic_superresolution_gaussian_microtextures_2024} by a kriging reasoning for Gaussian microtextures. We illustrate preliminary results we obtain in this case in Figure~\ref{fig:SR_study}. 
As before, we can compute the backward mean evolution of the different algorithms by applying their backward steps without adding noise.
We can observe the bias evolution of the different algorithms and we observe similar results than in the deblurring case, ranking methods in this order: CGDM, $\Pi$GDM and DPS. We observe that DPS is much more stable, it is possible to take $\alphaDPS = 1$.
However, the extension of the whole previous reasoning on deblurring to this problem is not trivial for the reasons explained below.

\setlength{\conddirtribwidth}{0.295\linewidth}

\begin{figure*}[t]
\centering
	\begin{tabular}{*{7}{>{\centering\arraybackslash}p{\conddirtribwidth}}}
	HR image	& LR image $\V$ & Theoretical mean & CGDM mean & $\Pi$GDM mean & DPS mean & Bias norm evolution \\
	\includegraphics[width=\conddirtribwidth]{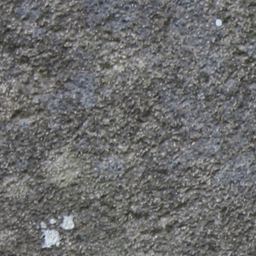}
	& 
	\includegraphics[width=\conddirtribwidth]{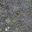}
	&
	\includegraphics[width=\conddirtribwidth]{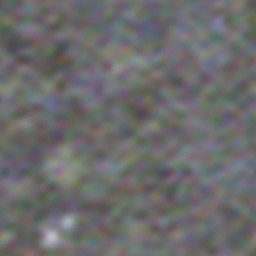}
	&
	\includegraphics[width=\conddirtribwidth]{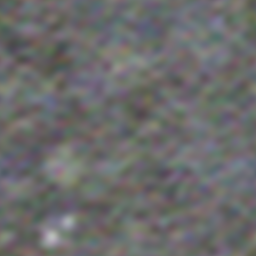}
	&
	\includegraphics[width=\conddirtribwidth]{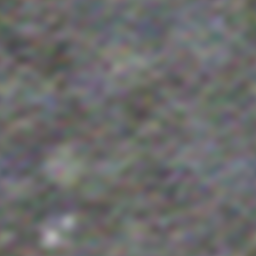}
	&
	\includegraphics[width=\conddirtribwidth]{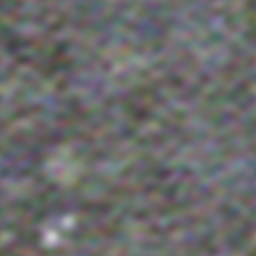}
	&
	\includegraphics[width=\conddirtribwidth]{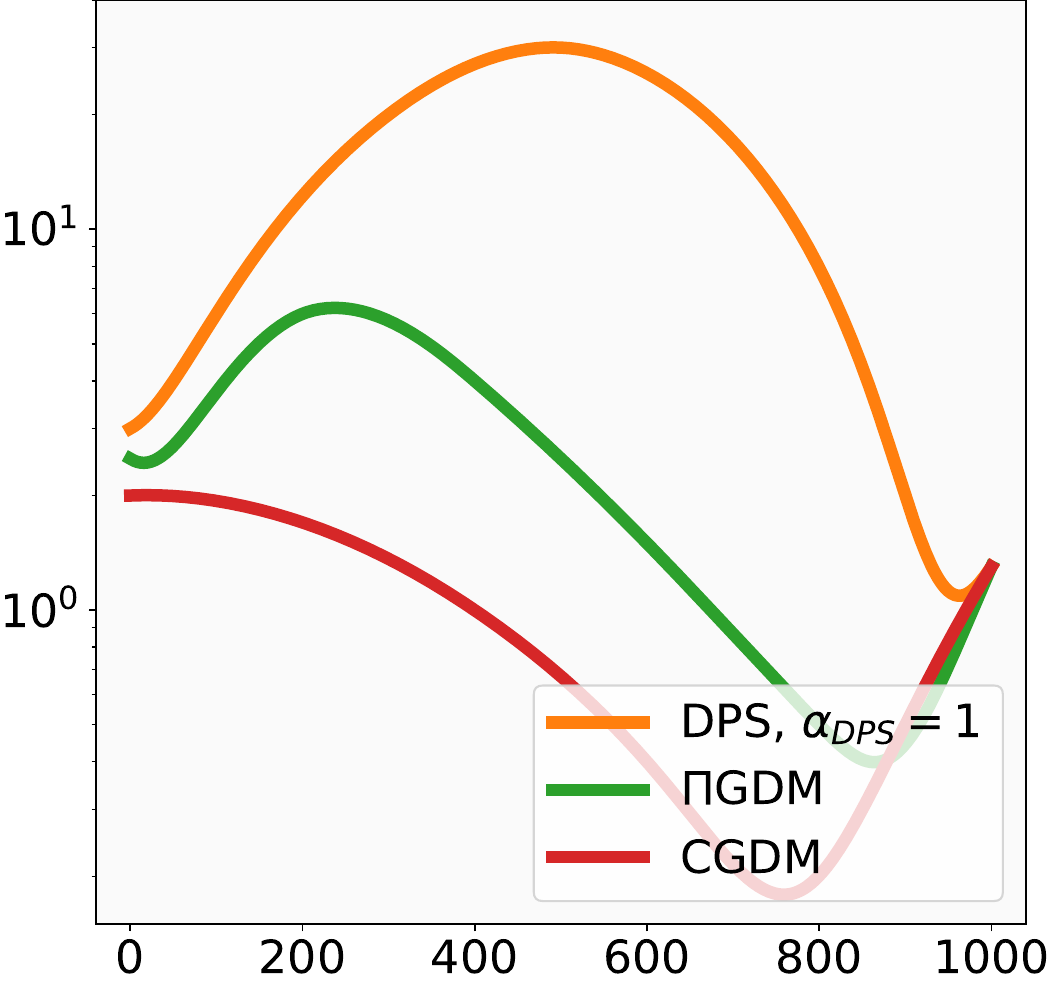}
	\end{tabular}
	\caption[Observation of the bias for the different algorithms for the SR problem.]{\label{fig:SR_study} \textbf{Observation of the bias for the different algorithms for the SR problem.} Illustration of the algorithms' bias for the SR problems with $r = 8$ and $\sigma = 10/255$. The observations are similar to the deblurring case.}
\end{figure*}

\newlength{\cifarwidth}
\setlength{\cifarwidth}{1.04\linewidth}
\begin{figure*}
\begin{tabular}{@{}cc@{}}
Images & Masked images \\
\includegraphics[width=\cifarwidth]{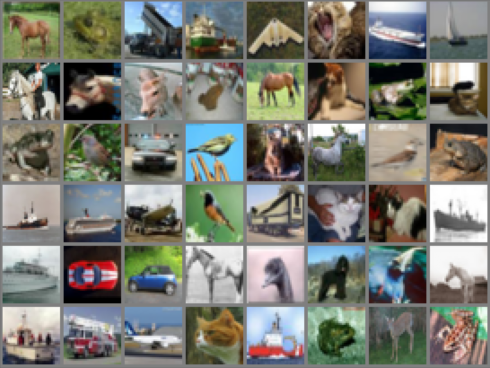}
&
\includegraphics[width=\cifarwidth]{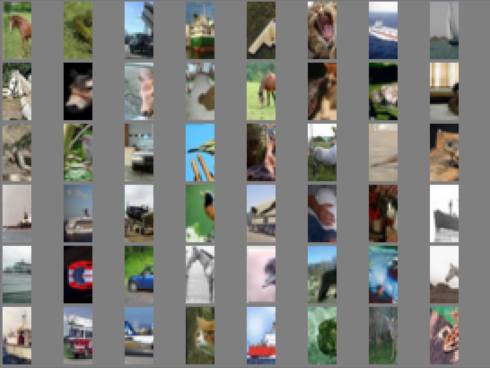} \\
DPS & CGDM \\
\includegraphics[width=\cifarwidth]{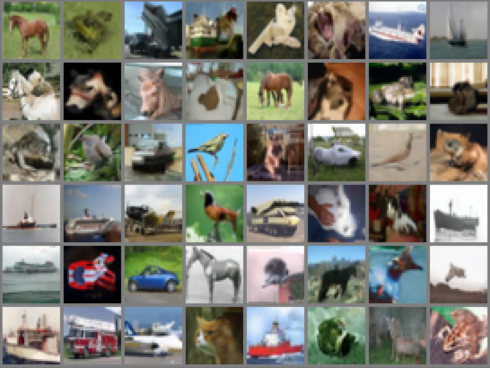}
&
\includegraphics[width=\cifarwidth]{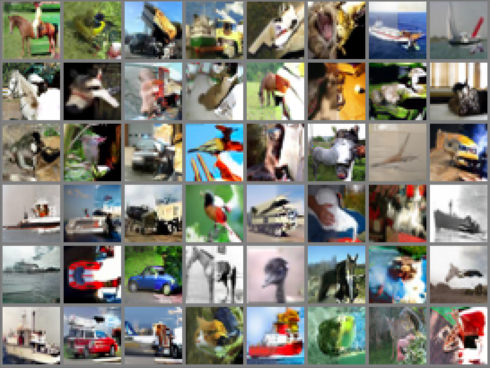} \\
\end{tabular}
	\caption{\label{fig:CGDM_on_CIFAR}
	\textbf{Illustration of the CGDM algorithm on the CIFAR10 dataset.} On a batch of 48 images of CIFAR10, for the inpainting task, we compare the DPS algorithm and the CGDM algorithm, applied with the empirical covariance computed on the training set. 
	}
\end{figure*}

\paragraph{Inability to compute the likelihood $\nabla \log p(\V \mid \y_t)$ term for RGB images.} For CGDM, we need to compute $(\sigma^2\I + \Sub \bC \bSigma \bSigma_t^{-1}\bC^T\Sub^T)^{-1}$. As demonstrated in \cite{pierret_diffusion_models_gaussian_distributions_2024}, computing efficiently and stably this inverse is a hard issue but it can be well-approximated by a diagonal RGB convolution, as done in Figure~\ref{fig:SR_study} and explained in \cite{pierret_diffusion_models_gaussian_distributions_2024}.

\paragraph{Non-simulataneous diagonalizability of the different algorithms along the time.} All the covariance matrices $(\bSigma^{\mathrm{DPS}}_t)_{0 \leq t \leq T}$, $(\bSigma^{{\Pi\mathrm{GDM}}}_t)_{0 \leq t \leq T}$, $(\bSigma^{\mathrm{CGDM}}_t)_{0 \leq t \leq T}$ are not simultaneously diagonalizable. Indeed, the operator $\A^{\mathrm{DPS}}_T$ defined in \Cref{sec:recursive_computation_backward_distributions}, with $\alpha_{\mathrm{DPS}} = 1$, has expression
\begin{equation}
\begin{aligned}
	 \A_T  = &\frac{1}{\sqrt{\alpha_T}}\Big(I -\beta_T \bSigma_t^{-1} \\
	 & -\frac{\beta_T}{\sigma^2} \overline{\alpha}_T\bSigma_T^{-1}\bSigma \bC^T\Sub^T\Sub\bC\bSigma\bSigma_T^{-1}\Big).
\end{aligned}
\end{equation}
\noindent In particular, the operator $\Sub^T\Sub$ is not diagonalizable in the Fourier domain and breaks the Fourier structure of the covariance matrices. {As discussed in Appendix\ref{appendix:proof_proposition_simultaneous_diagonalizable}, this also applies to blur kernels that are not identical across channels.}

\paragraph{Instabilities in high dimension.} The previous issue could be overcome by using the general expression of the 2-Wasserstein metric (\Cref{eq:W2_Gaussian}), which does not rely on simultaneous diagonalization. Nonetheless, the positive symmetric square root matrix of $\bSigma$ size has to be computed. The size of this matrix is $(3MN)^2$ which is too high to be computed in practice. Instabilities increase with the applications of it during 1000 steps.

\subsection{Extension to other inverse problems and non Gaussian datasets}

The inability to compute $\nabla \log p(\V \mid \y_t)$ along time, combined with the lack of simultaneous diagonalizability across different algorithms, also applies to more general inverse problems such as inpainting. In general, there is no reason to expect that the data covariance and the degradation operator share the same eigenvectors \cite{Galerne_Leclaire_gaussian_inpainting_2017_siims}. This presents a challenging research direction for developing metrics that closely approximate the exact 2-Wasserstein distance.

Beyond the Gaussian setting, it is not yet clear whether the CGDM algorithm can be effectively implemented. Nevertheless, preliminary experiments, presented in Figure~\ref{fig:CGDM_on_CIFAR} and conducted on an inpainting task using the CIFAR-10 dataset for visualization purposes, suggest that this approach could be a further investigation as a promising direction for improving the solution of inverse problems with diffusion models.
Some hyperparameters still need to be tuned, as the reconstructions produced by CGDM exhibit noticeable contrast artifacts.

\section{Conclusion}

We presented a rigorous evaluation of conditional diffusion models under Gaussian priors for inverse problems, with exact 2-Wasserstein computations in deblurring tasks. 
Our results show that both DPS and $\Pi$GDM exhibit notable biases and fail to adequately capture the posterior distribution, while the newly proposed CGDM aligns more closely with the true conditional law.
More precisely, we demonstrate, through both 2D toy experiments and real Gaussian microtexture deblurring tasks, that the three algorithms can be ranked in increasing order of performance as follows: DPS, $\Pi$GDM, and CGDM.
This ordering is consistent with the increasing complexity of the models adopted by these algorithms. Indeed, each successive method incorporates progressively richer covariance information about the data. 
Beyond deblurring, our methodology could be extended to a broader range of inverse problems and more complex, non-Gaussian data distributions. Extending this framework to such settings raises challenging open questions for future research.

\section*{Acknowledgements}
The authors are grateful to the anonymous reviewers for their valuable suggestions to improve the manuscript.
The authors acknowledge the support of the project MISTIC (ANR-19-CE40-005).
This project was provided with computing and storage resources by GENCI at IDRIS thanks to the grant 2025-AD011015533R1 on the supercomputer Jean Zay's A100 partition.

\bibliography{sn-bibliography}

\newpage
\clearpage
\begin{appendices}
\onecolumn

\section{Closed-form expressions for Gaussian diffusion models}

{In this appendix, we provide closed-form expressions for Gaussian distributions in the context of diffusion models. We first present a lemma to compute conditional Gaussian distributions and gives a compact expression for the denoised estimate $\widehat{\x}_0(\x_t)$ and use the first lemma to compute the theoretical $p(\x_t \mid \V)$, $p(\V\mid \x_t)$ and the theoretical discrete backward process, which sheds light on the inexactness of CGDM as observed in the Wasserstein error plots. Finally, we provide  the computation of  $p_t(\x_t \mid \V)$ given the noisy likelihood modeled by the different algorithms which is used to study them in a forward form in \Cref{sec:comp_algo_under_Gaussian_assumption}.}

{
\subsection{General computation of conditional Gaussian distributions}
In the following, we derive a lemma to compute conditional Gaussian distributions in our context. A first idea was to use the lemma from Section 2.3.3 in \cite{Bishop_pattern_recognition_and_machine_learning_2006} but it needs invertibility assumptions on the covariance matrices.  For this reason, we propose a more general kriging reasoning providing the following lemma.
\begin{lem}[Conditional Gaussian distrbution computations using a kriging reasoning]
\label{lem:Gaussian_conditonal_kriging}
Given the assumptions 
\begin{align}
    p(\x) & = \mathcal{N}(\bbm,\G) \\ 
    p(\y \mid \x) & = \mathcal{N}(\B\x, \tau^2\I) 
\end{align} 
the conditional distribution of $\x$ given $\y$ is given by
    \begin{align}
        p(\x\mid \y) = \mathcal{N}(\bmu_{\y\mid \x},\bSigma_{\y \mid \x})
    \end{align}
    with
    \begin{align}
        \bmu_{\y\mid \x}& = \bbm + \G \B^T \M^{-1}(\V-\B\bbm)
        \\
        \bSigma_{\y \mid \x} & = \G - \G \B^T\M^{-1}\B\G \\
        \M & = \B\G\B^T + \tau^2\I.
    \end{align}
\end{lem}

\noindent Note that $\M$ is invertible because $\B\G\B^T$ is a positive symmetric matrix.
\begin{proof}
    In the case $\bbm = \zero$, as shown by a kriging reasoning in Appendix E of \cite{pierret_stochastic_superresolution_gaussian_microtextures_2024},
    by denoting $\M = \B\G\B + \tau^2\I$, 
\begin{equation}
    \La^T\y + \tx - \La^T(\B\tx + \tau\boldsymbol{\tilde{n}})
\end{equation}
where $\tx$ is a sample from $p_0$ independent of $\y$ and $\n$ is an independent sample following $\mathcal{N}(\zero,\I)$ follows the posterior distribution $p(\x\mid\y)$ with $\La = \M^{-1}\B\G$ is solution of a kriging equation. Consequently, the posterior covariance matrix is the covariance matrix of this expression with respect to $\x$ which is
\begin{align}
    \bSigma_{\y\mid \x} & = (\I - \La^T\B)\G(\I - \La^T\B)^T + \tau^2\La^T\La \\
    & = \G  - \La^T\B\G + \La^T\B\G \B^T\La^T - \G\B^T\La^T + \tau^2\La^T\La\\
    & = \G  - \La^T\B\G + \La^T(\B\G \B^T + \G\B^T + \tau^2\I)\La - \G\B^T\La^T\\
    & = \G  - \La^T\B\G + \La^T\B\G - \G\B^T\La^T\\
    & = \G  -\G \B^T \M^{-1}\B\G
\end{align}
and the conditional expectation
\begin{align}
    \bmu_{\y\mid\x} 
    & = \G\B^T\M^{-1}\y.
\end{align}

In the general case $\bbm \neq \zero$, the covariance matrix is not affected and the conditional expectation becomes
    \begin{align}
    \bmu_{\y\mid\x} 
    & = \mu+\G\B^T\M^{-1}(\y-\B\bbm).
\end{align}
which is equivalent to rescale the quantity to ensure a null expectation.
\end{proof}

}
\subsection{Computation of \texorpdfstring{$\widehat{\x}_0(\x_t)$}{}}
\label{appendix:x_zero_chap}
{In the Gaussian setting $p_0 = \mathcal{N}(\bmu,\bSigma)$, the expression for $\widehat{\x}_0(\x_t)$ is given by Tweedie’s formula (see \Cref{eq:Tweedie_formula}), which, when combined with the explicit form of the score function (see \Cref{eq:Gaussian_score_DDPM}), yields the closed-form expression}
\begin{align}
	\widehat{\x}_0(\x_t)
	& = \frac{1}{\sqrt{\alphabar_t}}\left(\x_t + (1-\alphabar_t) \nabla_{\x}\log p_t(\x_t)\right) \\
	& = \frac{1}{\sqrt{\alphabar_t}}\left(\x_t - (1-\alphabar_t) \bSigma_t^{-1}(\x_t - \sqrt{\alphabar}_t \bmu)\right) \\
	& = (1-\alphabar_t)\bSigma_t^{-1}\bmu +\frac{1}{\sqrt{\alphabar_t}}\left(\x_t - (1-\alphabar_t) \bSigma_t^{-1}\x_t \right) \\
	& = (1-\alphabar_t)\bSigma_t^{-1}\bmu +\frac{\bSigma_t^{-1}}{\sqrt{\alphabar_t}}\left(\bSigma_t \x_t - (1-\alphabar_t) \x_t \right) \\
	& = (1-\alphabar_t)\bSigma_t^{-1}\bmu +\frac{\bSigma_t^{-1}}{\sqrt{\alphabar_t}}\left(\alphabar_t\bSigma\x_t\right) \\
	& = (1-\alphabar_t)\bSigma_t^{-1}\bmu +\sqrt{\alphabar_t}\bSigma_t^{-1}\bSigma\x_t.
\end{align}

\noindent Then, one can also derive
\begin{align}
(1-\alphabar_t)\bSigma_t^{-1}\bmu +\sqrt{\alphabar_t}\bSigma_t^{-1}\bSigma\x_t
& = \bSigma_t^{-1}((1-\alphabar_t)\I)\bmu+\sqrt{\alphabar_t}\bSigma_t^{-1}\bSigma\x_t \\
& = \bSigma_t^{-1}(\bSigma_t-\alphabar_t\bSigma)\bmu+\sqrt{\alphabar_t}\bSigma_t^{-1}\bSigma\x_t \\
& =  \bmu + \sqrt{\alphabar_t}\bSigma \bSigma_t^{-1}(\x_t - \sqrt{\alphabar}_t \bmu).
\end{align}

\subsection{Computation of \texorpdfstring{$p(\x_t \mid \V)$}{}}

\label{appendix:x_t_mid_v}
In this section, we compute $p(\x_0 \mid \V)$ and $p(\x_t \mid \V)$ in the Gaussian setting $p_0 = \mathcal{N}(\bmu,\bSigma)$.
We apply \Cref{lem:Gaussian_conditonal_kriging} with
\begin{align}
    \x & = \x_0 \\
    \y & = \V \\
    \bbm & = \bmu \\
    \G & = \bSigma \\
    \B & = \A \\
    \tau & = \sigma.
\end{align}
 By denoting $\M = \A\bSigma\A + \sigma^2\I$, it ensures that 
 \begin{equation}
     p(\x_0\mid \V) = \mathcal{N}(\bmu_{0\mid \V},\bSigma_{0\mid \V})
 \end{equation}
 with
\begin{align}
    \bSigma_{\x_0 \mid \V}
    & = \bSigma  -\bSigma \A^T \M^{-1}\A\bSigma \\
    \bmu_{0\mid \V} & = \bmu +\bSigma\A^T\M^{-1}(\V-\A\bmu) \\
    \M & = \A\bSigma\A^T+\sigma^2\I.
\end{align}

Then, because $p(\x_t\mid \x_0) = \mathcal{N}(\sqrt{\alphabar}_t\x_0, (1-\alphabar_t)\I)$,

\begin{equation}
    p(\x_t\mid\V) = \mathcal{N}(\sqrt{\alphabar}_t\E[\x_0\mid \V], \alphabar_t\bSigma_{\x_0 \mid \V}+(1-\alphabar_t)\I )
\end{equation}

\subsection{Computation of \texorpdfstring{$p(\V \mid \x_t)$}{}}
\label{appendix:v_mid_x_t}
Let us first compute $p(\x_0\mid \x_t)$. 
We apply \Cref{lem:Gaussian_conditonal_kriging} with
\begin{align}
    \x & = \x_0 \\
    \y & = \x_t \\
    \bbm & = \bmu \\
    \G & = \bSigma \\
    \B & = \sqrt{\alphabar}_t\I \\
    \tau & = (1-\alphabar_t).
\end{align}
We obtain the covariance matrix is
\begin{align}
    \bSigma_{0\mid t}
    & =
    \bSigma  - \alphabar_t\bSigma (\alphabar_t\bSigma +(1-\alphabar_t) \I))^{-1}\bSigma\\
    & = \bSigma  - \alphabar_t\bSigma^2 \bSigma_t^{-1} \\
    & = \bSigma\bSigma_t^{-1}(\bSigma_t - \alphabar_t\bSigma) \\
    & = (1-\alphabar_t)\bSigma\bSigma_t^{-1},
\end{align}
and the expectation is $\widehat{\x}_0(\x_t)$.
Then, $\V = \A\x_0 + \sigma^2\I$ and consequently,
\begin{equation}
    p(\V \mid \x_t) = \mathcal{N}(\A\widehat{\x}_0(\x_t),(1-\alphabar_t)\A\bSigma\bSigma_t^{-1} \A^T + \sigma^2\I).
\end{equation}

\subsection{Computation of the theoretical backward transitions \texorpdfstring{$p(\x_{t-1} \mid \x_{t})$}{}}
\label{appendix:exact_Gaussian_backward}
{In this section, we prove that the theoretical backward process associated with diffusion models applied to Gaussian distributions does not, in general, have a diagonal covariance matrix.
We apply \Cref{lem:Gaussian_conditonal_kriging} with
\begin{align}
    \x & = \x_{t-1} \\
    \y & = \x_t \\
    \bbm & = \sqrt{\alphabar}_{t-1}\bmu \\
    \G & = \bSigma_{t-1} \\
    \B & = \sqrt{\alpha_t}\I \\
    \tau & = \sqrt{\beta_t}.
\end{align}
With 
\begin{align}
    \M 
    & = \alpha_t \bSigma_{t-1}+\beta_t\I \\
    & = \alpha_t(\alphabar_{t-1}\bSigma + (1-\alphabar_{t-1})\I)+(1-\alpha_t)\I \\
    & = \alphabar_{t}\bSigma + (1-\alphabar_t)\I \\
    & = \Sigma_t
\end{align}
The covariance matrix of the distribution \( p(\x_{t-1} \mid \x_{t}) \) is given by 
\begin{align}
    \bSigma_{t-1\mid t}
    & = \bSigma_{t-1} - \alpha_t\bSigma_{t-1}^2\bSigma_t^{-1}
	  \\
    & = \alpha_t\bSigma_{t-1}^2\bSigma_t^{-1}(\bSigma_t-\alpha_t \bSigma_{t-1})\\
	& = \bSigma_{t-1}\bSigma^{-1}_t\left(\alphabar_t\bSigma+(1-\alphabar_t)\I - \alphabar_t\bSigma-(\alpha_t-\alphabar_t)\I\right) \\
	& = \beta_t\bSigma_{t-1}\bSigma^{-1}_t
\end{align}
\noindent and the conditional expectation is
\begin{align}
	\E(\x_{t-1}\mid \x_t)
	& = \sqrt{\alphabar}_{t-1}\bmu+\sqrt{\alpha}_t\bSigma_{t}^{-1}\bSigma_{t-1}(\x_t-\sqrt{\alphabar}_t\bmu).
\end{align}
\noindent Using the identity
\begin{equation}
	\bSigma_{t} = \alphabar_t \bSigma + (1-\alphabar_t)\I = \alpha_t \bSigma_{t-1} + \beta_t \I \iff \bSigma_{t-1} = \frac{1}{\alpha_t}(\bSigma_{t}-\beta_t\I),
\end{equation}
\noindent we obtain
\begin{align}
	\E(\x_{t-1}\mid \x_t)
& = \frac{1}{\sqrt{\alpha}_t}\left((\x_t-\sqrt{\alphabar}_t\bmu)-\beta_t\bSigma_{t}^{-1}(\x_t-\sqrt{\alphabar}_t\bmu)\right)+\sqrt{\alphabar}_{t-1}\bmu \\
& = \frac{1}{\sqrt{\alpha}_t}\left(\x_t-\beta_t\bSigma_{t}^{-1}(\x_t-\sqrt{\alphabar}_t\bmu)\right) \\
& = \frac{1}{\sqrt{\alpha_t}}(\x_t + \beta_t \nabla \log p_t(\x_t))
\end{align}
which is the correct expression for the conditional expectation given by the backward process (see \Cref{eq:backward_DDPM}).

}

\subsection{Computation of \texorpdfstring{$\tilde{p}(\x_t \mid \V)$}{} for the different algorithms}

\label{appendix:proof_expression_C_v_t}
{In the Gaussian setting where $p_0 = \mathcal{N}(\bmu,\bSigma)$, the forward process defined in \Cref{eq:forward_DDPM} yields}  $p_t(\x_t) = \mathcal{N}(\sqrt{\alphabar}_t \bmu,\bSigma_t)$ with $\bSigma_t = \alphabar_t\bSigma + (1-\alphabar_t)\I$ and $p(\V \mid \x_t) = \mathcal{N}(\A\widehat{\x}_0(\x_t),\bC_{\V \mid t})$, as developed in \Cref{sec:conditional_DDPM}. By Bayes' rule,
\begin{equation}
	\nabla_{\x} \log \tilde{p}_t(\x_t \mid \V) 
	= \nabla_{\x} \log p_t( \x_t) +\nabla_{\x} \log p_t( \V\mid \x_t) 
\end{equation}
\noindent Consequently,
\begin{equation}
	\nabla_{\x} \log \tilde{p}_t(\x_t \mid \V) 
	= -\bSigma_t^{-1}(\x_t-\sqrt{\alphabar}_t\bmu)-\sqrt{\alphabar_t}\bSigma \bSigma_t^{-1}\A^T\bC_{\V \mid t}^{-1}(\A\widehat{\x}_0(\x_t)-\V)  
\end{equation}
\noindent By denoting $\tilde{p}_t(\x_t\mid \V) = \mathcal{N}(\tilde{\bmu}_{t \mid \V},\tilde{\bC}_{t\mid \V})$, $\nabla_{\x} \log \tilde{p}_t(\x_t\mid \V) = -\tilde{\bC}_{\V \mid t}^{-1}(\x_t - \tilde{\bmu}_{\V \mid \x_t})$ and by identifying the terms in $\x_t$,
\begin{align}
	\tilde{\bC}_{t \mid \V}^{-1}
	& = \bSigma_t^{-1}+\alphabar_t\bSigma \bSigma_t^{-1}\A^T\bC_{\V \mid t}^{-1}\A\bSigma\bSigma_t^{-1}.
\end{align}
Furthermore, by Woodburry matrix identity \cite{Higham_accuracy_stability_Woodburry_2002},
\begin{equation}
	\left(B + UDV \right)^{-1} = B^{-1} - B^{-1}U \left(D^{-1} + VB^{-1}U \right)^{-1} VB^{-1}.
\end{equation}
\noindent Consequently, with $B = \bSigma_t, U = -\sqrt{\alphabar}_t\bSigma\A^T, V = \sqrt{\alphabar}_t\A\bSigma,$ and $D$ such that $D^{-1} + VB^{-1}U = \bC_{\V \mid t}$ ie $D = \left(\bC_{\V \mid t}+\alphabar_t \A\bSigma^2\bSigma_t^{-1}\A^T\right)^{-1}$,
\begin{align}
	\tilde{\bC}_{t \mid \V}
	& = \bSigma_t-\alphabar_t\bSigma\A^T\left(\bC_{\V \mid t}-\alphabar_t \A\bSigma^2\bSigma_t^{-1}\A^T\right)^{-1}\A\bSigma
\end{align}
and finally,
{\small
\begin{align}
	\tilde{\bC}^{\mathrm{DPS}}_{t \mid \V}
	& = \bSigma_t-\alphabar_t\bSigma\A^T\left(\sigma^2\I+\alphabar_t \A\bSigma^2\bSigma_t^{-1}\A^T\right)^{-1}\A\bSigma \\
	\tilde{\bC}^{{\Pi\mathrm{GDM}}}_{t \mid \V}
	& = \bSigma_t-\alphabar_t\bSigma\A^T\left(\sigma^2\I+(1-\alphabar_t)\A\A^T+\alphabar_t \A\bSigma^2\bSigma_t^{-1}\A^T\right)^{-1}\A\bSigma \\
	\tilde{\bC}^{\mathrm{CGDM}}_{t \mid \V}
	& = \bSigma_t-\alphabar_t\bSigma\A^T\left(\sigma^2\I+(1-\alphabar_t)\A\bSigma\bSigma_t^{-1}\A^T+\alphabar_t \A\bSigma^2\bSigma_t^{-1}\A^T\right)^{-1}\A\bSigma \\
	& = \bSigma_t-\alphabar_t\bSigma\A^T\left(\sigma^2\I+\A\bSigma\bSigma_t^{-1}((1-\alphabar_t)\I+\alphabar_t \bSigma)\A^T\right)^{-1}\A\bSigma \\
	& = \bSigma_t-\alphabar_t\bSigma\A^T\left(\sigma^2\I+\A\bSigma\A^T\right)^{-1}\A\bSigma
\end{align}}
By identifying the other terms,
\begin{align}
	\tilde{\bC}_{t \mid \V}^{-1}\tilde{\bmu}_{t \mid \V}
	& =  \sqrt{\alphabar}_t\bSigma_t^{-1}\bmu+\sqrt{\alphabar_t}\bSigma \bSigma_t^{-1}\A^T\bC_{\V \mid t}^{-1}\left(\V +\alphabar_t \A\bSigma\bSigma_t^{-1}\bmu-\A\bmu \right) \\
	& = \sqrt{\alphabar}_t\left(\bSigma_t^{-1}+\alphabar_t\bSigma\bSigma_t^{-1}\A^T\bC_{\V \mid t}^{-1}\A\bSigma\bSigma_t^{-1}\right)\bmu +\sqrt{\alphabar_t}\bSigma \bSigma_t^{-1}\A^T\bC_{\V \mid t}^{-1}\left(\V -\A\bmu \right)\\
	& = \sqrt{\alphabar}_t\tilde{\bC}_{t \mid \V}^{-1}\bmu+\sqrt{\alphabar_t}\bSigma \bSigma_t^{-1}\A^T\bC_{\V \mid t}^{-1}\left(\V -\A\bmu \right),
\end{align}
\noindent and
\begin{equation}
	\bmu_{t \mid \V}
	=  \sqrt{\alphabar}_t\bmu+\sqrt{\alphabar_t}\tilde{\bC}_{t \mid \V}\bSigma \bSigma_t^{-1}\A^T\bC_{\V \mid t}^{-1}\left(\V -\A\bmu \right).
\end{equation}

\section{Proof of Proposition~\ref{prop:compute_recursively_cov_mus_algos}}
\label{appendix:proof:prop:compute_recursively_cov_mus_algos}
Each algorithm defines a backward process, as described in \Cref{algo:backward_conditional_DDPM}. The objective is to characterize these processes at each time step.
Under the Gaussian assumption, the relation between $\y_t$ and $\widehat{\x}_0(\y_t)$ is linear, as given in \Cref{eq:Gaussian_x_0_chap}, and
\begin{equation}
	\frac{1}{2}\nabla_{\y_t} \|\A \widehat{\x}_0(\y_t)-\V\|^2_{\bC_{\V \mid t}^{-1}}
	= \sqrt{\alphabar_t}\bSigma \bSigma_t^{-1}\A^T\bC_{\V \mid t}^{-1}(\A\widehat{\x}_0(\y_t)-\V).
\end{equation}
As a consequence, the backward process associated with a given algorithm can be written as
\begin{equation}
		\y_T \sim \N, \quad 
		\y_{t-1} = \A_t^{\mathrm{algo}}\y_{t} + \bb_t^{\mathrm{algo}}+\sqrt{\beta_t} \z_t,\\
		\quad 1 \leq t \leq T, \z_t \sim \N.
\end{equation}
with
\begin{equation}
\label{eq:A_t}
	 \A_t^{\mathrm{algo}}
	 = \frac{1}{\sqrt{\alpha_t}}
	\Big(
	\I
	-\beta_t \bSigma_t^{-1}
	 -\beta_t \overline{\alpha}_t
	\bSigma_t^{-1}\bSigma \A^T
	\left(\bC^\mathrm{algo}_{\V \mid t}\right)^{-1}
	\A\bSigma\bSigma_t^{-1}
	\Big)
\end{equation}
and
\begin{equation}
\label{eq:b_t}
	 \bb_t^{\mathrm{algo}}
	=
	\beta_t\sqrt{\overline{\alpha}}_{t-1}
	\bSigma_t^{-1}\bSigma \A^T
	\left( \bC^\mathrm{algo}_{\V \mid t}\right)^{-1} 
	(\V-\A\bmu+\alphabar_t\A\bSigma\bSigma_t^{-1}\bmu)
	+ \frac{\beta_t}{\sqrt{\alpha_t}}\bSigma_t^{-1}\bmu.
\end{equation}
This formulation shows that the resulting backward processes remain Gaussian, since all operations involved are linear. Their characterization therefore reduces to computing the sequences of means $(\bmu_t^\mathrm{algo})_{0 \leq t \leq T}$ and covariance matrices $(\bSigma_t^\mathrm{algo})_{0 \leq t \leq T}$ at each time step.
Because the score operations are linear, the sequence of means $(\bmu_t^\mathrm{algo})_{0 \leq t \leq T}$ can be obtained by running \Cref{algo:backward_conditional_DDPM} without injecting noise at each iteration.

\section{Analytical derivations for diffusion models on Gaussian microtextures}

\label{appendix:ADSN_computations}

\subsection{Application of the exact score function}

\label{appendix:ADSN_score_application}

The ADSN model \cite{Galerne_Gousseau_Morel_random_phase_textures_2011_IEEE} allows for the exact computation of the score function associated with the iterations of diffusion models.
The inverse of the matrix $\bSigma_t = \alphabar_t \bSigma + (1 - \alphabar_t)\I$ appears in the score function, as given in Equation~\eqref{eq:Gaussian_score_DDPM}.
Let us now describe why this inversion is feasible.

\noindent First, we recall \Cref{eq:score_matrix_3d}: for $\xi \in \OMN$,
\begin{equation}
\widehat{\bSigma}_t(\xi) = \alphabar_t \widehat{\bt}(\xi)\left[\widehat{\bt}(\xi)\right]^T + (1 - \alphabar_t)\I_3. \tag{{\color{black}\ref{eq:score_matrix_3d}}}
\end{equation}
In a certain sense, the action of $\bSigma_t$ is separable across all frequencies.
We can invert it using the following lemma.
\begin{lem}
\label{lem:inverse_1_rank_3d}
	Let $\y \in \C^3$, $a,b \in \R^+$,
	\begin{equation}
		\left(a\y\overline{\y}^T + b\I_3\right)^{-1} = \frac{1}{b}I_3 -  \frac{a\|y\|^2}{b(a\|y\|^2+b)}\y\y^T.
	\end{equation}
\end{lem}
\begin{proof}
    It is well-known that $\frac{\y\overline{\y}^T}{\|y\|^2}$ is the orthogonal projection on $\Span(\y)$ (see for example \cite{golub_matrix_computations_2013}).
Consequently, by completing $\frac{\y}{\|y\|^2}$ in an orthogonal basis and considering its matrix $\bP$,
	\begin{equation}
		\y\overline{\y}^T 
		= \bP^T\begin{pmatrix}
			\|y\|^2 & 0 & 0 \\
			0 & 0 & 0 \\
			0 & 0 & 0
		\end{pmatrix}\bP.
	\end{equation}
	Then,
	\begin{equation}
		\bP (a\y\overline{\y}^T+bI_3) 
		= \bP^T\begin{pmatrix}
			a\|y\|^2+b & 0 & 0 \\
			0 &  b& 0 \\
			0 & 0 & b
		\end{pmatrix}\bP
	\end{equation}
	and
	\begin{align}
		 (a\y\overline{\y}^T+bI_3)^{-1}
		& = \bP^T\begin{pmatrix}
			\frac{1}{a\|y\|^2+b} & 0 & 0 \\
			0 &  \frac{1}{b}& 0 \\
			0 & 0 & \frac{1}{b}
		\end{pmatrix}\bP  \\
		& = \frac{1}{b}I_3 + \left(\frac{1}{a\|y\|^2+b}-\frac{1}{b}\right)\y\overline{\y}^T \\
		& = \frac{1}{b}I_3-  \frac{a\|y\|^2}{b(a\|y\|^2+b)}.
	\end{align}
\end{proof}
\noindent This lemma follows from the fact that the symmetric rank-one matrix $\frac{\y \overline{\y}^T}{|\y|^2}$ is the orthogonal projection onto $\Span(\y)$.
It can be applied to the matrix $\bSigma_t$ separately at each frequency, following \Cref{eq:score_matrix_3d}. To be complete, for $\x \in \R^{3\OMN}$ and $ \xi \in \OMN$,

\begin{equation}
    \widehat{\bSigma_t^{-1}\x}(\xi)
    = \frac{1}{1-\alphabar_t}\widehat{\x}(\xi) - \frac{\alphabar_t\|\widehat{\bt}(\xi)\|^2}{(1-\alphabar_t)(\alphabar_t\|\widehat{\bt}(\xi)\|^2 + \alphabar_t) } \widehat{\bSigma}(\xi)\widehat{\x}(\xi)
\end{equation}
where $\widehat{\x}(\xi) = (\begin{smallmatrix}
    \widehat{\x}_1(\xi) & \widehat{\x}_2(\xi) & \widehat{\x}_3(\xi)
\end{smallmatrix})^T \in \R^3$

\subsection{Proof of \Cref{prop:simult_diago_cov_backwards}}

\label{appendix:proof_proposition_simultaneous_diagonalizable}

As proved in Appendix I.2 of \cite{pierret_diffusion_models_gaussian_distributions_2024}, it is sufficient to study the 3D structure of the covariance matrix of ADSN to determine its eigenvectors and associated eigenvalues. In a certain sense, the matrix is block-diagonalizable with respect to 3D blocks. We will show that its eigenvectors are preserved over time. For a given frequency $\xi \in \OMN$, we denote 
\begin{equation}
    \widehat{\V}_1(\xi) = \widehat{\bt}(\xi) =  \begin{pmatrix}
        \widehat{\bt}_1(\xi) \\
        \widehat{\bt}_2(\xi) \\
        \widehat{\bt}_3(\xi) \\
    \end{pmatrix}, \widehat{\V}_2(\xi)  =  \begin{pmatrix}
        -\overline{\widehat{\bt}}_3(\xi) \\
        0 \\
        \overline{\widehat{\bt}}_1(\xi) \\
    \end{pmatrix}, \widehat{\V}_2(\xi)  =  \begin{pmatrix}
       0 \\
         -\overline{\widehat{\bt}}_3(\xi)\\
        \overline{\widehat{\bt}}_2(\xi) \\
    \end{pmatrix}.
\end{equation}
It is an orthogonal basis of eigenvectors of $\widehat{\bSigma}(\xi)$ because
\begin{align}
    \overline{\widehat{\V}}_k(\xi)^T\widehat{\V}_\ell(\xi) & = 0 \text{ for $1\leq k< \ell \leq 3$} \\
    \widehat{\bSigma}(\xi)\widehat{\V}_1(\xi) 
    & = [\overline{\widehat{\bt}}(\xi)^T\widehat{\bt}(\xi)]\widehat{\V}_1(\xi) \\
    \widehat{\bSigma}(\xi)\widehat{\V}_2(\xi) 
    & =\zero \\
    \widehat{\bSigma}(\xi)\widehat{\V}_3(\xi) 
    & = \zero .
\end{align}
We denote $\lambda_k(\xi)$ the eigenvalues associated with $\widehat{\V}_k(\xi)$ (respectively $[\overline{\widehat{\bt}}(\xi)^T\widehat{\bt}(\xi)]$, $0$ and $0$). 

Recall that in the deblurring context, $\A = \bC$ an RGB convolution that applies the kernel $\bc\in\R^{\OMN}$ to each channel.
Denoting $\widehat{\bC}(\xi)\in\R^{3\times 3}$ the action of the operator in the Fourier domain, $\widehat{\bC}(\xi)$ conserves the same eigenvectors because
\begin{equation}
\label{eq:c_preserves_structure}
     \widehat{\bC}(\xi)\widehat{\V}_k(\xi) 
     = \hat{\bc}(\xi)\widehat{\V}_k(\xi) \text{ for $1\leq k \leq 3$}.
\end{equation}
Besides, $\widehat{\bSigma}_t^{-1}$ share the same eigenvectors as $\bSigma$.
This ensures that the matrix $\A_t^\mathrm{algo}$ given in \Cref{eq:A_t} and recalled here
\begin{equation}
	\A_t^{\mathrm{algo}}
	 = \frac{1}{\sqrt{\alpha_t}}\left(\I -\beta_t \bSigma_t^{-1} -\beta_t \overline{\alpha}_t\bSigma_t^{-1}\bSigma \A^T\left(\bC^\mathrm{algo}_{\V \mid t}\right)^{-1}\A\bSigma\bSigma_t^{-1}\right)
\end{equation}
\noindent preserves the eigenvectors of $\bSigma$. Finally, the application of \Cref{prop:compute_recursively_cov_mus_algos} ensures that the covariance matrices associated with the algorithms' backward processes preserve the eigenvector basis. 
Similarly, the conditional forward covariance ${\bC}_{t\mid \V}$ is a combination of operators that preserve the eigenvectors of $\bSigma$.

\begin{remark}
    \Cref{eq:c_preserves_structure} shows that our method cannot be extended to cases where different kernels are applied to different image channels, highlighting the crucial role of the specific structure of the deblurring problem degradation operator.
\end{remark}

\section{Proof of Proposition~\ref{prop:cut_Wass_in_two_parts}}
\label{appendix:proof:prop:cut_Wass_in_two_parts}



Let us consider the orthogonal decomposition $\R^d = \ker \bSigma \oplus \left(\ker \bSigma\right)^\perp$ and denote by $\PkerS$ and $\PkerSperp = \I - \PkerS$ the associated orthogonal projectors.
By Proposition~\ref{prop:simult_diago_cov_backwards}, the Gaussian distributions $p_t^\mathrm{algo}(\y_t \mid \V) = \mathcal{N}(\bmu_t^{\mathrm{algo}},\bSigma_t^{\mathrm{algo}})$ and $p_t(\x_t \mid \V) = \mathcal{N}({\bmu}_{t\mid \V},{\bC}_{t\mid \V})$ share the same eigenvalues as $\bSigma$. Hence, one can decompose the Wassertein error by projecting the two processes, that is,
\begin{equation}
\begin{aligned}
& \Wass^2\left(p_t^\mathrm{algo}(\y_t \mid \V) ,	p_t(\x_t \mid \V)\right)
\\
& =
\Wass^2\left(
	\mathcal{N}(\bmu_t^{\mathrm{algo}},\bSigma_t^{\mathrm{algo}}),
	\mathcal{N}({\bmu}_{t\mid \V},{\bC}_{t\mid \V})
	\right)
	\\
& = \Wass^2\left(
	\mathcal{N}(\PkerS \bmu_t^{\mathrm{algo}},\PkerS\bSigma_t^{\mathrm{algo}}\PkerS),
	\mathcal{N}(\PkerS{\bmu}_{t\mid \V},\PkerS{\bC}_{t\mid \V}\PkerS)
	\right)
\\
& \quad + \Wass^2\left(
	\mathcal{N}(\PkerSperp \bmu_t^{\mathrm{algo}},\PkerSperp\bSigma_t^{\mathrm{algo}}\PkerSperp),
	\mathcal{N}(\PkerSperp{\bmu}_{t\mid \V},\PkerSperp{\bC}_{t\mid \V}\PkerSperp)
	\right)
\\
&=: \left(\WasskerS^\mathrm{algo}\right)^2+\left(\WassorthokerS^\mathrm{algo}\right)^2.
\end{aligned}
\end{equation}

Let us discuss why this metric is of particular interest by showing that $\WasskerS^\mathrm{algo}$ is identical across all algorithms. Indeed, on $\ker \bSigma$, we can precisely describe the iterations of the different algorithms. All algorithms behave identically on the zero eigenvalues of $\bSigma$, acting like an unconditional DDPM within this kernel subspace. This can be seen by considering the operators $\A_t^{\mathrm{algo}}$ and vectors $\bb_t^{\mathrm{algo}}$ defined in \Cref{prop:compute_recursively_cov_mus_algos} with expression given in \Cref{eq:A_t} and \Cref{eq:b_t}. Indeed,
\begin{equation}
\begin{aligned}
	 \PkerS\A_t^{\mathrm{algo}} 
	& = \frac{1}{\sqrt{\alpha_t}}\left(\PkerS -\beta_t \PkerS\bSigma_t^{-1}\right) = \frac{1}{\sqrt{\alpha_t}}\PkerS\left(\I -\beta_t \frac{1}{1-\alphabar_t}\I\right), \\
	\PkerS \bb_t^{\mathrm{algo}}
	& = \PkerS\frac{\beta_t}{\sqrt{\alpha_t}}\frac{1}{1-\alphabar_t}\bmu.
\end{aligned} 
\end{equation}
These expressions are independent of $\bC_{\V \mid t}^{\mathrm{algo}}$, the covariance matrix that distinguishes the different algorithms.

\begin{remark}
One can further interpret the common value $\WasskerS^2$.
Recall from Equation~\eqref{eq:conditional_forward_DDPM} that 
\begin{equation}
{\bmu}_{t\mid \V} = \sqrt{\alphabar_t}\bmu+\sqrt{\alphabar_t}\bSigma \A^T \M^{-1}\left(\V-\A\bmu\right)
\quad \text{and}\quad 
{\bC}_{t\mid \V} = \bSigma_t - \alphabar_t\bSigma \A^T \M^{-1} \A\bSigma.
\end{equation}
with $\bSigma_t = \alphabar_t \bSigma + (1-\alphabar_t)\I$.	
Hence,
\begin{equation}
\PkerS{\bmu}_{t\mid \V} = \sqrt{\alphabar_t}\PkerS\bmu
\quad \text{and}\quad 
\PkerS{\bC}_{t\mid \V}\PkerS = (1-\alphabar_t)\PkerS.
\end{equation}
Hence $\WasskerS^\mathrm{algo}$ corresponds to the error committed by the DDPM algorithm to approximate the constant distribution $\delta_{\PkerS\bmu}$.
This error is due to the fact that the DDPM scheme is unable to retrieve the rank of $\bSigma$. Note that this error can be managed by modifying the noise schedule, as explored in \cite{Strasman_analysis_noise_schedule_SGM_2025_TMLR}.
\end{remark}

\end{appendices}

\end{document}

%% file: Notations.tex
\DeclareMathOperator{\Cov}{Cov}

\DeclareMathOperator{\ADSN}{ADSN}

\newcommand{\Wass}{\mathbf{W}_2}

\DeclareMathOperator{\Span}{span}

\newcommand{\R}{\mathbb{R}}

\newcommand{\C}{\mathbb{C}}
\newcommand{\N}{\mathcal{N}(\zero,\I)}

\newcommand{\OMN}{{\Omega_{M,N}}}

\newcommand{\E}{\mathbb{E}}

\newcommand{\I}{\boldsymbol{I}}
\newcommand{\A}{\boldsymbol{A}}
\newcommand{\B}{\boldsymbol{B}}
\newcommand{\M}{\boldsymbol{M}}
\newcommand{\bbm}{\boldsymbol{m}}
\newcommand{\bC}{\boldsymbol{C}}

\newcommand{\Sub}{\boldsymbol{S}}
\newcommand{\G}{\boldsymbol{\Gamma}}
\newcommand{\La}{\boldsymbol{\Lambda}}

\newcommand{\bSigma}{\boldsymbol{\Sigma}}

\newcommand{\bP}{\boldsymbol{P}}

\newcommand{\PkerS}{\boldsymbol{P}_{\ker\bSigma}}
\newcommand{\PkerSperp}{\boldsymbol{P}_{(\ker\bSigma)^\perp}}

\newcommand{\WasskerS}{\mathbf{W}_{2,\ker\bSigma,t}}
\newcommand{\WassorthokerS}{\mathbf{W}_{2,(\ker\bSigma)^\perp,t}}

\newcommand{\x}{\boldsymbol{x}}
\newcommand{\y}{\boldsymbol{y}}
\newcommand{\z}{\boldsymbol{z}}
\newcommand{\tx}{\tilde{\boldsymbol{x}}}
\newcommand{\ty}{\tilde{\boldsymbol{y}}}
\newcommand{\tz}{\tilde{\boldsymbol{z}}}

\newcommand{\w}{\boldsymbol{w}}
\newcommand{\V}{\boldsymbol{v}}

\newcommand{\bmu}{\boldsymbol{\mu}}
\newcommand{\bxi}{\boldsymbol{\xi}}

\newcommand{\n}{\boldsymbol{n}}

\newcommand{\zero}{\boldsymbol{0}}

\newcommand{\U}{\boldsymbol{u}}

\newcommand{\bt}{\boldsymbol{t}}
\newcommand{\bb}{\boldsymbol{b}}
\newcommand{\bc}{\boldsymbol{c}}

\newcommand{\tp}{\tilde{p}}

\newcommand{\alphabar}{\overline{\alpha}}

\newcommand{\alphaDPS}{\alpha_{\text{\tiny DPS}}}

\newtheorem{assumption}{Assumption}
\newtheorem{lem}{Lemma}
\newtheorem{prop}{Proposition}

\newcommand{\algo}{{\mathrm{algo}}}